\documentclass[sigconf]{acmart}
 \renewcommand\footnotetextcopyrightpermission[1]{} 
\usepackage{booktabs} 

\usepackage[ruled]{algorithm2e} 

\SetAlFnt{\small}
\SetAlCapFnt{\small}
\SetAlCapNameFnt{\small}
\SetAlCapHSkip{0pt}

\acmJournal{TOG}
\usepackage{amsmath}
\usepackage{booktabs}
\usepackage{graphicx}
\usepackage{xcolor}
\usepackage{overpic}
\usepackage{enumitem} 
\usepackage{overpic} 
\usepackage{color}
\usepackage{colortbl}
\usepackage{multirow}
\usepackage{algorithm2e}
\RestyleAlgo{ruled}
\usepackage{algpseudocode}
\usepackage{subcaption}
\usepackage{comment}
\usepackage{wrapfig}

\definecolor{turquoise}{cmyk}{0.65,0,0.1,0.3}
\definecolor{purple}{rgb}{0.65,0,0.65}
\definecolor{dark_green}{rgb}{0, 0.5, 0}
\definecolor{orange}{rgb}{0.8, 0.6, 0.2}
\definecolor{red}{rgb}{0.8, 0.2, 0.2}
\definecolor{darkred}{rgb}{0.6, 0.1, 0.05}
\definecolor{blueish}{rgb}{0.0, 0.3, .6}
\definecolor{light_gray}{rgb}{0.7, 0.7, .7}
\definecolor{pink}{rgb}{1, 0, 1}
\definecolor{greyblue}{rgb}{0.25, 0.25, 1}
\usepackage{graphicx}
\usepackage[export]{adjustbox}

\newcommand{\method}{~GEM3D~} 

\renewcommand{\paragraph}[1]{\vspace{1em}\noindent\textbf{#1}.}

\newcommand{\bc}{\mathbf{c}}
\newcommand{\bd}{\mathbf{d}}

\newcommand{\bff}{\mathbf{f}}

\newcommand{\bh}{\mathbf{h}}

\newcommand{\bm}{\mathbf{m}}
\newcommand{\bn}{\mathbf{n}}

\newcommand{\bp}{\mathbf{p}}
\newcommand{\bq}{\mathbf{q}}

\newcommand{\bs}{\mathbf{s}}

\newcommand{\bx}{\mathbf{x}}

\newcommand{\bz}{\mathbf{z}}

\newcommand{\bI}{\mathbf{I}}

\newcommand{\bphi}{\mbox{\boldmath$\phi$}}

\newcommand{\btheta}{\mbox{\boldmath$\theta$}}
\newcommand{\bomega}{\mbox{\boldmath$\omega$}}
\newcommand{\bzero}{\mbox{\boldmath$0$}}

\newcommand{\mD}{\mathcal{D}}

\newcommand{\mN}{\mathcal{N}}

\usepackage{blindtext}

\newcommand{\revision}[1]{\textbf{\textcolor{purple}{#1}}}

\definecolor{blueish}{rgb}{0.0, 0.3, .6}

\definecolor{purple}{rgb}{0.6, 0.0, .6}

\begin{document}

\title{GEM3D: GEnerative Medial Abstractions for 3D Shape Synthesis}


\author{Dmitry Petrov}
\affiliation{
  \institution{UMass Amherst}
  \country{}
}
\email{dmpetrov@umass.edu}

\author{Pradyumn Goyal}
\affiliation{
  \institution{UMass Amherst}
 \country{}
}
\email{pradyumngoya@umass.edu}

\author{Vikas Thamizharasan}
\affiliation{
  \institution{UMass Amherst}
  \country{}
}
\email{vthamizharas@umass.edu }

\author{Vladimir G. Kim}
\affiliation{
  \institution{Adobe Research}
  \country{}
}
\email{vokim@adobe.com}

\author{Matheus Gadelha}
\affiliation{
  \institution{Adobe Research}
 \country{}
}
\email{matheusabrantesgadelha@gmail.com}

\author{Melinos Averkiou}
\affiliation{
  \institution{CYENS CoE and University of Cyprus}
  \country{}
}
\email{m.averkiou@cyens.org.cy}

\author{Siddhartha Chaudhuri}
\affiliation{
  \institution{Adobe Research}
 \country{}
}
\email{sidch@adobe.com}

\author{Evangelos Kalogerakis}
\affiliation{
  \institution{UMass Amherst and CYENS CoE}
 \country{}
}
\email{kalo@cs.umass.edu}

\authorsaddresses{}

%
%


%
%


\begin{abstract}
We introduce GEM3D 
\footnote{Project page (with code): \href{https://lodurality.github.io/GEM3D/}{lodurality.github.io/GEM3D}}
-- a new deep, topology-aware generative model of 3D shapes. The key ingredient of our method is a neural skeleton-based representation encoding information on both shape topology and geometry. Through a denoising diffusion probabilistic model, our method first generates skeleton-based representations following the Medial Axis Transform (MAT), then generates surfaces through a skeleton-driven neural implicit formulation. The neural implicit takes into account the topological and geometric information stored in the generated skeleton representations to yield surfaces that are more topologically and geometrically accurate compared to previous neural field formulations. We discuss applications of our method in shape synthesis and point cloud reconstruction tasks, and evaluate our method both qualitatively and quantitatively. We demonstrate significantly more faithful surface reconstruction and diverse shape generation results compared to the state-of-the-art, also involving challenging scenarios of reconstructing and synthesizing structurally complex, high-genus shape surfaces from Thingi10K and ShapeNet.
\end{abstract}    

\begin{teaserfigure}
\vspace{-1mm}
\begin{center}
 \includegraphics[width=1\textwidth]{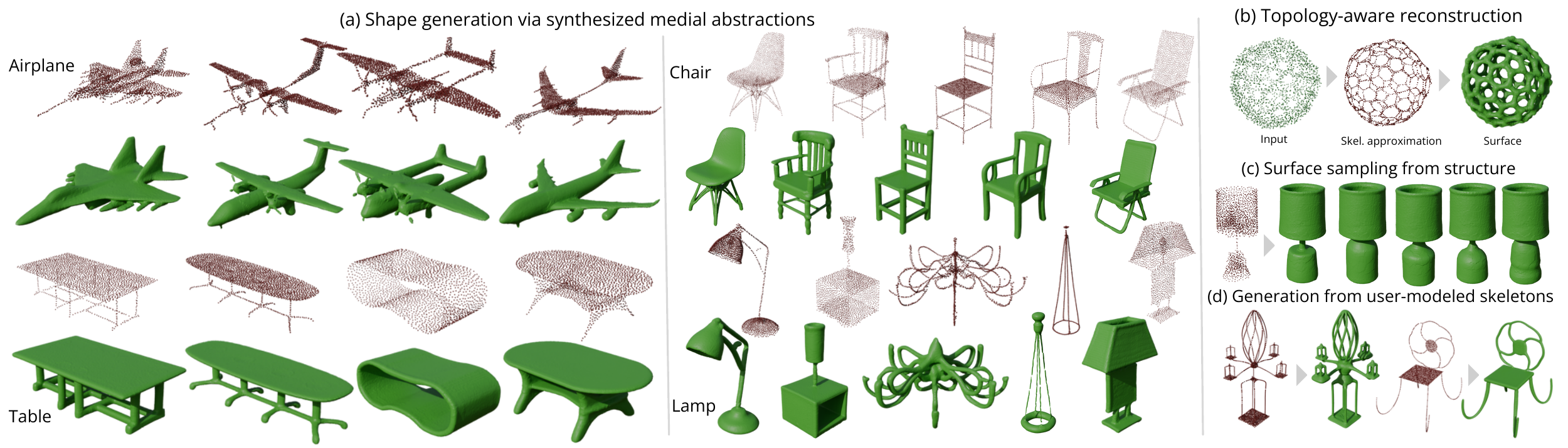}
\end{center} 
 \vspace{-3mm}
 \caption{We present GEM3D, a neural generative model able to (a) synthesize skeleton-based representations, or in other words, medial abstractions (in red), and surfaces (in green) that follow the structure and topology information encoded in the skeletons. GEM3D can also be used in various applications, such as (b) reconstructing topologically complex surfaces, (c) generating diverse shape samples following a given skeleton, and (d) generating surfaces from novel, user-specified skeletons.
 \label{fig:teaser} 
 }  
\end{teaserfigure}

\maketitle
\pagestyle{plain}

\section{Introduction}
\label{sec:intro}

Automated or semi-automated 3D shape synthesis is a significant and challenging problem in geometric modeling, with wide-ranging applications to computer-aided design, fabrication, architecture, art, and entertainment. While early work in this space primarily focused on handwritten models~\cite{lsystems,urban_rec,architecture}, subsequent work employed statistical learning to infer generative design principles from data~\cite{prob_models, prob_scenes}. In recent years, a variety of approaches have developed deep neural network-based architectures for 3D synthesis~\cite{xu2023survey,shi2023survey,generative_models,data_driven_scenes}. While these methods can capture diverse macro-level appearances, they rarely model shape structure or topology explicitly, relying instead on the representational power of the network to generate plausible-looking voxel grids~\cite{gen_voxels}, point clouds~\cite{gen_points}, meshes~\cite{gen_meshes}, or implicit fields~\cite{gen_implicits}. Since 3D networks 
are hampered by the additional resource overheads incurred by the extra dimension compared to 2D image generation networks, they often struggle to model fine detail and connectivity. Some approaches model part layouts~\cite{li2017grass}, but are limited in the complexity of the structures they can generate.

At the same time, these prior 3D synthesis methods rarely give artists flexible, precise control. They act more as black boxes for unconditional generation, or reconstruction from images or 3D scans. Recent methods introduce synthesis based on text prompts~\cite{poole2022dreamfusion,lin2023magic3d}, with remarkable results but only high-level, global control via prompt engineering. 3D character artists have long been accustomed to posing skeleton rigs for accurate character configuration. However, such direct local control and interpretability through intuitive abstractions has had limited success in general 3D shape synthesis. Approaches without explicit structural modeling lack the ability to specify a particular desired topology, e.g. a~chair with a particular slat configuration in the back. On the other hand, approaches that do model part-level structure are restricted to simple topologies defined by a few coarse primitives and cannot model complex fretwork or ornamentation.

We are interested in realistic 3D shape generation that accurately models complex topological and geometrical details, and supports more interpretable control of shape structure and geometry. To achieve this, we build upon three key insights: (1) topological detail can often be captured in a ``skeletal abstraction'', like that obtained by a medial axis transform~\cite{TagliasacchiSOTAReport}, 
which serves as a simplified structural proxy for the shape, even in the absence of meaningful part decompositions; (2) these abstractions can be synthesized with generative methods~\cite{karras2022elucidating}, predicted from sparse point clouds~\cite{yin2018p2pnet,mesoskelpcomp}, or created manually by an artist, and need not be perfect since they are simply intermediate representations; and (3) each abstraction can be decoded into realistic surfaces by another trained model.

Our approach implements the surface generation step by inferring and assembling a collection of locally-supported neural implicit functions, conditioned on the skeletal abstraction. We draw inspiration from recent work in this area which associates a latent code with each 3D point in a sparse set, and generates a local implicit from a latent grid~\cite{Zhang_3dilg}. The mixture of these implicits defines the overall synthesized shape, and allows for better generation of fine geometric details than a single large implicit. However, the sparse set of point supports in prior work tends to be arbitrary and not very interpretable. Follow-up work based on 3D neural fields and cross-attention~\cite{3DShape2VecSet} drops explicit spatial grounding on the latent grids altogether. In contrast, our skeleton-based latent grids are more structure-aware, providing interpretable supports for latent codes in 3D space, while remaining capable of representing complex, fine-grained topological structure.

We summarize our contributions as follows:
\vspace{-1mm}
\begin{itemize}
 \item We propose a generative model based on diffusion to automatically synthesize skeleton-based shape representations along with their supporting latents encoding shape information. The model's training procedure is performed without any form of user input or manual tuning. 
  \item We also devise a neural implicit representation that can be used to accurately regress the shape surface from a skeletal representation and associated latent field.
  \item According to our experiments, our method produces significantly more faithful surface reconstruction and diverse shape generation results compared to the state-of-the-art. Our method handles challenging scenarios of reconstructing and synthesizing structurally complex, high-genus shape surfaces from Thingi10K and ShapeNet (Figure \ref{fig:teaser}), including  synthesizing surfaces from user-specified skeletons largely different from ones observed during training.    
\end{itemize}
\vspace{-1mm}

\section{Related work}

\label{sec:related}

\paragraph{3D Skeletonization} The computation of medial skeletons from a surface representation, is a well-studied problem in geometry processing. 1D curves or 2D sheets or a combination of them have been traditionally used as skeletal representations of 3D shapes.  
Most skeletonization methods rely on analytic approximations to skeletons, such as Voronoi diagrams  \cite{Amenta98Surface,yan2018voxel} and Power diagrams \cite{wang2022computing}, morphological operations, such as surface contraction \cite{taglia_sgp12}, topological graphs, such as Reeb graphs \cite{Hilaga01Topology}, or optimization based on various geometric criteria (e.g., symmetry preservation \cite{Tag2009}, normal preservation \cite{Wu2015DeepPoints},  set coverage \cite{dou2022coverage}).
For a full discussion of geometric methods for skeletonization, we refer readers to the survey by Tagliasacchi et al.~\shortcite{TagliasacchiSOTAReport}.
More recently, neural network approaches have been proposed for extracting skeleton-based representations from point clouds ~\cite{yin2018p2pnet,ge2023point2mm},  implicit shape representations \cite{rebain2021deep,Clemot2023},  volumetric representations \cite{AnimSkelVolNet}, meshes \cite{RigNet}, point cloud sequences \cite{xu2022morig},
or images \cite{hu2023s3ds,hu2022immat,xu2022apes}.
Our work is focused not on computing those representations, but on creating efficient
\emph{shape generation} models grounded on medial abstractions. While we also explore how to estimate such abstractions from point clouds, we primarily demonstrate how these intermediate representations can be used to guide the generation process, leading to more interpretable and topologically faithful shapes.

\paragraph{Skeleton-Guided Surface Representation} 3D shapes can be approximated as unions of spheres. For instance, a surface can be reconstructed by piecewise linear interpolation of medial spheres centered on skeleton points \cite{Li2005}.
Shape modeling tools like ZBrush~\cite{ZBrush} allow artists to manually create such a skeleton of spheres via a feature called \emph{ZSpheres}.
\emph{Sphere Meshes} \cite{Thiery2013SphereMeshes} used this representation for automatic shape approximation and extended it to non-tubular geometry.
\emph{Convolution Surfaces} \cite{Bloomenthal1991ConvolutionSurfaces} are another class of skeleton-based implicit surfaces where the surface is defined as the level set of a function obtained by integrating a kernel function along the skeleton.
To avoid blending artifacts, Zanni et al.~\shortcite{Zanni2013SCALIS} proposed scale-invariant blending. Our work investigates using such abstractions to guide the generative process for the best of two worlds -- efficient topological representation and interpretability from medial abstractions, and detailed surface generation from data-driven neural fields.
 More recently, neural networks have been developed to translate complete or partial surface point clouds \cite{yin2018p2pnet, mesoskelpcomp, yang2020p2mat} or RGB images \cite{tang2019skeleton, tang2021skeletonnet,hu2022immat} to skeletal representations, and back to either surface points or 3D meshes. None of these approaches are generative i.e., are capable of generating multiple valid skeletal representations given their inputs. We also develop a novel neural surface representation based on skeleton-supported local implicits that directly yields a continuous surface, and is designed to be efficient, detailed, and artifact-free.



\paragraph{Neural Fields}
Scalar fields parameterized by neural networks have been frequently used as a shape representation in many deep generative models~\cite{mescheder2019occupancy,park2019deepsdf,gen_implicits}.
When compared to other representations like voxels or meshes, those \emph{neural fields} are capable
of representing shapes with varying topology with reasonable amount of detail without requiring
prohibitive memory footprint.
More recent approaches encode spatially-varying features that locally modulate the neural fields to
improve the accuracy of the generated shapes~\cite{Zhang_3dilg,3DShape2VecSet,peng2020convolutional, hui2022wavelet,cheng2023sdfusion,shue2023triplane,zheng2023locally,zhou2023ga_sketching}.
These approaches can generate topologically incorrect shapes.
Our approach addresses these issues by representing shapes as neural fields that encode a ``thickening'' operation on a medial skeletal representation, that naturally leads to more interpretable and topologically-faithful shapes.

\paragraph{Structure-Aware 3D Generation}
A full review of 3D generative models is beyond our scope: we refer the reader to excellent surveys~\cite{xu2023survey,shi2023survey}. Methods that directly generate ``raw'' representations such as voxels, meshes, or point clouds frequently yield topological artifacts, omissions, and other inaccuracies in regions with thin/fine details. Further, they lack interpretable intermediates to help users control the generative process. Structure-aware representations~\cite{generative_models}, such as part-based models (e.g.,~\cite{Huang:2015:deeplearningsurfaces,li2017grass,dubrovina2019composite,wu2020pq,anise_tvcg2023}) potentially have better topological fidelity and interpretability. However, they are suitable for shapes with small numbers of parts and not for more topologically complex structures with ambiguous part decompositions. 
We use medial abstractions as structure-aware intermediate representation to avoid these limitations.

\section{Method}

\begin{figure*}[t!]
    \centering
    \includegraphics[width=1.0\textwidth]{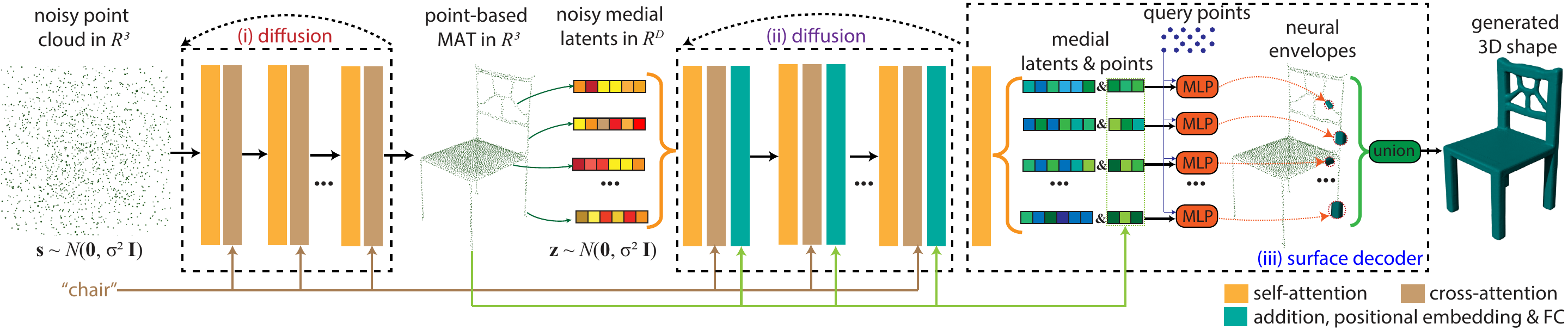}
    \caption{\method generative architecture: starting with Gaussian noise in $\mathbb R^3$, our \textcolor{red}{first diffusion stage} generates a point-based medial (skeletal) shape representation conditioned on a shape category embedding. Conditioned on this representation, our \textcolor{purple}{second diffusion stage} generates latent codes capturing shape information around the medial points. In the last stage, our \textcolor{blue}{surface decoder} decodes the medial latent codes and points to local neural implicit surface representations, which are then aggregated to create an output 3D shape.}     \label{fig:architecture}    
\end{figure*}

At the heart of our method lies a neural generative model (Figure \ref{fig:architecture}) that first generates a skeletal representation of a shape dedicated to compactly capturing its topology. Then, conditioned on this skeletal representation, our generative model synthesizes the surface in the form of an implicit function locally modulated by the skeletal data. The skeletal representation approximates the medial axis of a 3D shape, which has widely been used to capture shape topology.
In the following section, we briefly overview preliminaries, then we discuss our neural implicit function and generative approach.

\subsection{Preliminaries}
\label{sec:preliminaries}

\paragraph{Medial Axis Transformation}

\begin{wrapfigure}{R}{0.5\columnwidth}
\centering
\includegraphics[width=1\linewidth]{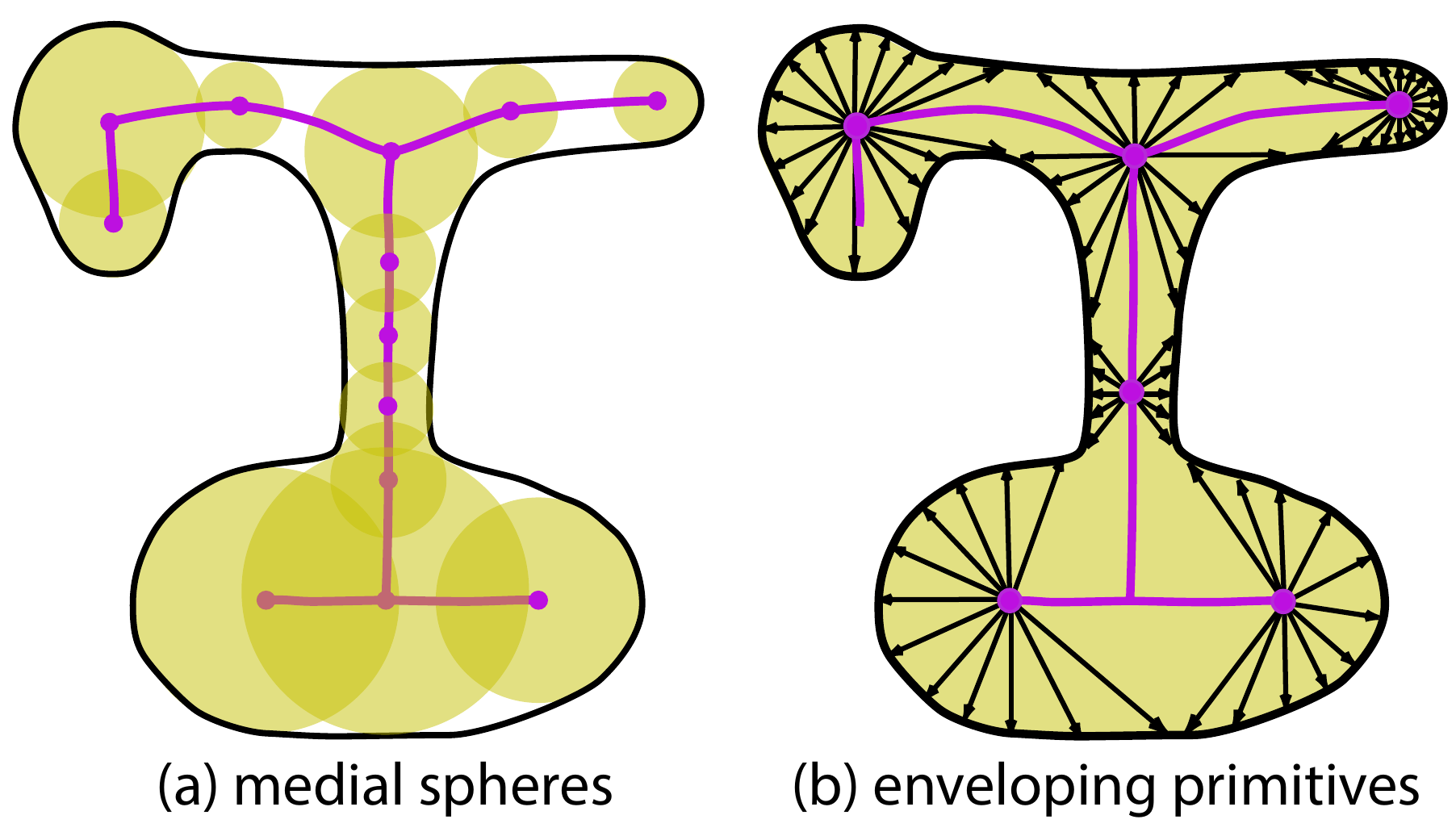}
  \vspace{-8mm}
  \caption{A shape and its medial axis  (purple) (a) medial balls (yellow) vs (b) enveloping primitives.
  \label{fig:medial_axis_vs_medial_enveloping} }
\vspace{-3mm}       
\end{wrapfigure}

Consider a closed, oriented, and bounded shape $S$ in $\mathbb R^3$, the medial axis is represented as a set of centers of maximal spheres inscribed within the shape (Figure \ref{fig:medial_axis_vs_medial_enveloping},  left). \setlength{\columnsep}{10pt}

Each of these medial spheres is tangent to at least two points
on the boundary $\partial S$ of S and does not contain any other boundary
points in its interior. The medial axis transformation (MAT) comprises both the medial axis and the radius associated
with each sphere center \shortcite{TagliasacchiSOTAReport}. The MAT can be used to compactly capture both topological shape information and its geometry \cite{Li2005}, yet as recently noted in \cite{guo2023medial}, its ability to capture local surface details, such as shape protrusions and high curvature regions, is limited, since a substantial number of medial spheres are often needed (Figure \ref{fig:medial_axis_vs_medial_enveloping},  left).

\paragraph{Enveloping Primitives}
To circumvent the above limitation,  Guo et al. \shortcite{guo2023medial} introduced an implicit function, called ``generalized enveloping primitive'', which represents the shape's surface as a directional distance field around the medial axis. Specifically, given a query point $\bx$ in $\mathbb R^3$ and a medial mesh $S$ discretizing the medial axis of a given shape, their enveloping function outputs a signed distance value for query points depending on their closest medial elements (medial mesh vertex, edge, or face in their case):
\begin{equation}
E_S(\bx) = ||\bx - \bs_{\bx}|| - r(\bd_{\bs,\bx})
\end{equation}
where $\bs_x$ is the closest medial mesh element to the query point $\bx$ in Euclidean sense, $\bd_{\bs,\bx} = (\bx - \bs_{\bx})/\lVert  \bx - \bs_{\bx} \rVert $ represents a unit direction vector from the closest medial mesh element towards the query point, and $r(\cdot)$ is a radius function. The radius function essentially defines an envelope, or displacement, around the medial axis, which is modulated differently depending on different directions around it (Figure \ref{fig:medial_axis_vs_medial_enveloping}, right), providing much better surface approximation compared to using medial spheres. Guo et al. \shortcite{guo2023medial}  estimates the above enveloping function for a given input 3D shape through a global optimization and iterative refinement procedure.  

\subsection{Surface generation}
\label{sec:surface_decoder}

\begin{wrapfigure}{R}{0.57\columnwidth}
\centering
  \includegraphics[width=1\linewidth]{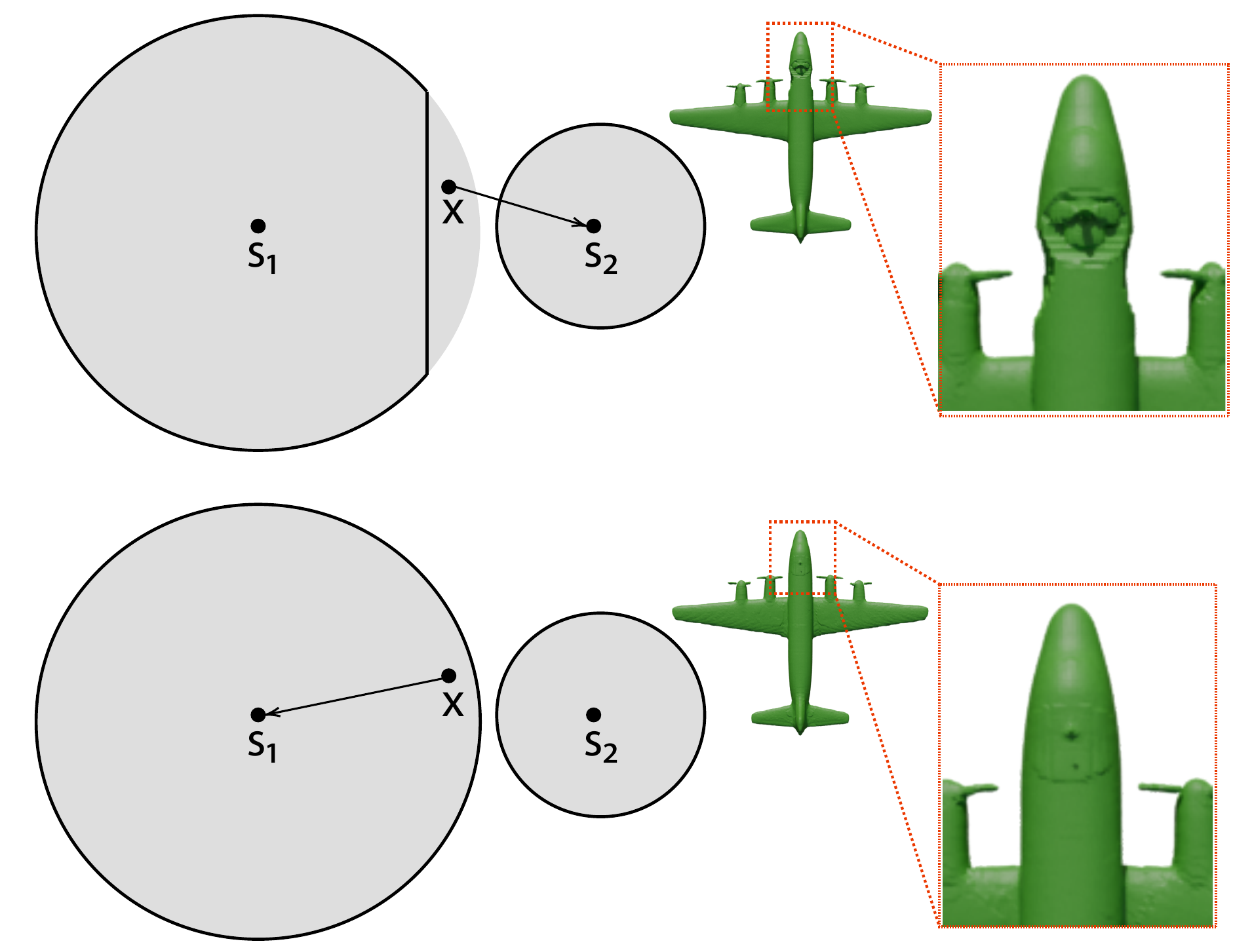}
  \vspace{-5mm}
  \caption{ (Top) Using the closest medial point for queries yields wrong ball reconstructions. (Bottom) Using the closest envelope yields the right result. Surface reconstruction is shown in green for both cases.
  \label{fig:closest_medial_elem_vs_prim} }
\vspace{-3mm}       
\end{wrapfigure}

\paragraph{Neural enveloping}  We  extend the above primitives to ``neural envelopes'', such that can be used in our neural generative model. We employ a neural network, parameterized by learnable parameters $\btheta$, to approximate the  radius function. In our case, the radius function depends not only on the direction from a medial element to the query point but also a medial latent code, which is specific to the medial element and aims to encode surface information around the medial axis. The latent code helps decoding towards a more accurate surface, since it is trained to encode shape information. In addition, instead of finding the  medial element closest to the query point, as done in \cite{guo2023medial}, we found that a significantly better surface approximation can be achieved by finding the medial element whose surface envelope is closest to the query point (Figure \ref{fig:closest_medial_elem_vs_prim}, see also our ablation). Specifically, our neural enveloping function defines the following implicit:
\begin{equation}
F(\bx; \btheta) = \min\limits_{i} \Big( ||\bx - \bs_i||  -  r(\bd_{i,\bx}, \bz_i; \btheta) \Big)
\label{eq:decoder}
\end{equation}
where $\bs=\{\bs_i\}_{i=1}^N$ is a set of $N$ medial element positions in $\mathbb R^3$ produced by our generative model, $\bz=\{\bz_i\}_{i=1}^N$ are corresponding latent codes ($256$-dimensional in our implementation) also produced by our generative model, and $\bd_{i,\bx} = (\bx - \bs_{i})/\lVert  \bx - \bs_{i} \rVert $ are unit vectors from each medial element towards the query point. 

The neural network function $r$  has the form of a multi-layer MLP. Given the values of the implicit function extracted at a dense grid of query points ($256^3$ in our implementation), the zero-level iso-surface can be extracted using the marching cubes algorithm \cite{lorensen1987marching}.  Note that evaluating the Eq. \ref{eq:decoder} is computationally expensive given that it requires evaluating the $\min$ over the envelopes of all queries and medial elements. In practice, to accelerate computation, we sample a dense number of unit directions ($1000$ in our implementation) around each medial element, estimate the largest value for the radius across all directions for each medial element i.e., form a sphere bounding the envelope of each medial element, then compute the $\min$ over only the medial elements whose bounding spheres include the query point. In this manner, we find for each query point, the medial elements that approximately lie in the vicinity of the query point. 

\paragraph{Training} Given the surfaces of training shapes with their medial elements, the parameters of the MLP can be trained to minimize the $L_1$ loss measuring error between predicted locations of sample points and 
the location of training surface sample points sampled at different directions around each medial element:
\begin{align}
L_{surf}(\btheta) = 
\frac{1}{N} \sum_{i=1}^N 
\frac{1}{|\mD(i)|} \sum_{d \in \mD(i)} 
\lVert \hat{\bx}_{i,d} - \bx_{i,d} \rVert_1, where \label{eq:surface_loss}\\
\bx_{i,d} =  \bs_i +  \bd_{i} \cdot r(\bd_{i}, \bz_i; \btheta) \label{eq:point_rec}
\end{align}
where $\mD(i)$ represents a set of training surface points found by casting rays from each medial element $i$ along several sample directions ($1000$ in our implementation), $\hat{\bx}_{i,d}$ is the 3D location of each training surface point, and $\bd_i$ is a sample unit direction vector. The loss is averaged over all shapes of the training datasets. Note that in contrast to other neural implicit surface formulations that often use point samples all over in $\mathbb R^3$ to avoid trivial solutions (i.e., zero implicit values everywhere), our MLP only needs surface sample points to be trained on.

\subsection{Generative model}
\label{sec:generative_model}
\label{sec:generative}

Central to our approach is the generation of medial elements along with their latent codes. Our generative model proceeds in two stages: first, we generate the positions of medial elements through a denoising diffusion process \cite{ho2020denoising,karras2022elucidating}, then we generate their latent codes conditioned on their position through another subsequent diffusion process. 
Our two-stage process allows the generation of a multitude of different shapes conforming to the same skeleton structure, including user-specified skeletons, as discussed in our results and applications.

\paragraph{Generation of medial elements} Our first stage synthesizes a point-based representation of the medial axis i.e., a simplified skeleton form including only points as  medial elements (as opposed to a mesh). The diffusion process starts by sampling $N$ 3D point positions from Gaussian noise $\bs \sim \mN(0, \sigma_\text{max}^2 \bI)$, where $\sigma_\text{max}$ is an initially prescribed standard deviation. Then an iterative denoising procedure is initiated to synthesize a point-based skeleton. The point denoising is executed by iteratively solving a probability flow ODE \cite{karras2022elucidating} in a series of time steps:
\mbox{$d\bs= -\dot{\sigma}(t)  \sigma(t) \nabla_{\bs} \log p\big(\bs; \sigma(t) \big)dt$}
where $\sigma(t)$ is a schedule defining the desired noise level at time $t$, $\dot{\sigma}(t)$ is its derivative, and \mbox{$\nabla_{\bs} \log p\big(\bs; \sigma(t) \big)dt$} is the score function for diffusion models \cite{song2021scorebased}. Following this vector field nudges the sample towards areas of higher density of the data distribution i.e., the distribution over plausible point-based skeletons in our case. The vector field is approximated with the help of a denoiser neural network. The network takes as input: (i) a noisy skeleton sample $\bs$ consisting of $2048$ noisy points in $\mathbb R^3$ in our implementation, (ii) the noise level $\sigma_t$ at a time step $t$, (iii) and a learnable embedding vector $\bc$  representing a desired category of shapes (e.g., $55$ different embedding vectors for $55$ categories in ShapeNet). The embedding is used for category-conditioned generation i.e., generate skeletons conditioned on a desired category, such as ``airplanes''. With the help of the denoiser network, the diffusion process outputs a skeletal approximation consisting of $2048$ sample points. The denoiser  network consists of a set of blocks each producing a feature representation for each medial point based on self-attention and cross-attention operations \cite{vaswani2017attention,3DShape2VecSet}. Specifically, each block performs the following operations:
\begin{eqnarray}
\{\bff_i^{(l)} \}_{i=1}^N = \text{SelfAttn}\big( \{\bff_i^{(l-1)}\}_{i=1}^N \big) 
\label{eq:self-attn}
\\
\{\bff_i^{\prime(l)} \}_{i=1}^N = \text{CrossAttn}\big( \{\bff_i^{(l)}\}_{i=1}^N,  \bc \big)
\label{eq:cross-attn}
\end{eqnarray}
where $\bff_i^{(l)}$, $\bff_i^{\prime(l)}$ are feature vectors ($256$-dimensional in our implementation)
 for each medial point computed after self attention and cross attention respectively from the block $l$. The first block uses as features the current sample positions per medial point. The noise level is taken into account the network through $\sigma$-dependent skip connections \cite{karras2022elucidating}. We  refer to the same paper for more details on noise variance scheduling and hyperparameters.
 
\paragraph{Training of the medial point denoiser network}
The parameters $\bphi_1$ of denoising network $D_1(\bs, \sigma_t, \bc; \bphi_1)$
are trained by minimizing the  $L_2$ denoising loss for samples drawn from training skeletons:
\begin{equation}
L_{\text{diff},1}(\bphi_1)=\mathbb{E}_{\hat{\bs} \sim p_\text{data}} \mathbb{E}_{\bn \sim \mN(\bzero, \sigma_t^2 \bI)}  \lVert D_1(\hat{\bs} + \bn, \sigma_t, \bc; \bphi_1) - \bs \rVert^2_2
\end{equation}
where $\hat{\bs}$ are training point-based skeletons and $\bn$ is sampled Gaussian noise. The training requires obtaining reference point-based skeletons for shapes.  

We developed a fast, distance field-based skeletonization algorithm whose basic idea is to move surface sample points set towards the negative gradient of the signed distance function through an iterative gradient descent procedure such that surface shrinks and moves towards the medial axis. We provide more details about our skeletonization in the supplementary material. The skeletonization was performed automatically without any manual parameter tuning per shape category or testing scenario. 

\paragraph{Generation of medial latents} Given a generated point-based skeleton
$\bs=\{\bs_i\}_{i=1}^N$ from the first diffusion model, or a point-sampled skeleton provided as input for skeleton-based shape synthesis, our second diffusion stage aims at generating a set of latent vectors  $\bz=\{\bz_i\}_{i=1}^N$. Each latent vector $\bz_i$ is generated such that it corresponds to the medial point $\bs_i$. This diffusion stage proceeds in a similar manner to the first one. We start by sampling $N$ 3D latent vectors from  Gaussian noise, then these are iteratively denoised  by iteratively solving the same probability flow ODE \cite{karras2022elucidating} as in our first stage (i.e., substituting medial points with medial latents). An important difference is that the denoiser network used to compute the score function is additionally conditioned on the medial point positions so that the network outputs latents tailored to each medial point. Specifically, the denoiser  network $D_2(\bz, \bs, \sigma_t, \bc; \bphi_2)$ consists of a set of blocks implementing the self-attention and cross-attention operations with the input category embedding, similarly to Eq. \ref{eq:self-attn} and \ref{eq:cross-attn} respectively.

As an additional operation, each block adds a  positional embedding to the input feature vector $\bh_i$ of each medial point based on its position:  $\bh_i^{\prime(l)} = \bh_i^l + g( \bs_i)$, where $g$ represents a frequency-based positional embedding of the medial point positions \cite{sitzmann2020siren},
followed by a fully connected layer. We observed that adding this positional embedding offered significantly better reconstruction results, since it made each latent aware of its corresponding medial position. The feature representations of the last block is processed through a block of self-attention layers to exchange the information within the latent set -- it produces the final latents used in our surface decoder (Eq. \ref{eq:decoder}).

\paragraph{Training of the medial latent denoiser network}
The parameters $\bphi_2$ of denoising network $D_2(\bz, \bs, \sigma_t, \bc; \bphi_2)$ are trained by minimizing the expected $L_2$ denoising error loss for training latents:
\begin{equation}
L_{\text{diff},2}(\bphi_2)=\mathbb{E}_{\hat{z} \sim p_{\text{data}}} \mathbb{E}_{\bn \sim \mN(\bzero, \sigma_t^2 \bI)}  \lVert D_2(\hat{\bz} + \bn, \sigma_t, \bc; \bphi_2) - \bz \rVert^2_2
\end{equation}
where $\hat{\bz}$ are training skeletal latent codes. To provide these latent codes, we devised an autoencoder-based,  unsupervised learning strategy where the training latents are estimated such that they yield an optimal surface reconstruction error for the training shapes. More specifically, in a pre-training step that aims to estimate latents for training shapes, we train an encoder with learnable parameters $\bomega$ that takes as input a set of dense surface sample points  $\{\hat{\bx}_k\}_{k=1}^K$ ($200K$ in our implementation) along with the training medial points $\{\hat{\bs}_i\}_{i=1}^N$,  encodes them into latent codes,  then decodes them back to predict surface points:
\begin{eqnarray}
\{\hat{\bz}_i\}_{i=1}^N = \text{Encoder}( \{\hat{\bx}_k\}_{k=1}^K, \{\hat{\bs}_i\}_{i=1}^N  ; \bomega) \\
\{\bx_k\}_{k=1}^K = \text{Decoder}( \{\hat{\bs}_i\}_{i=1}^N, \{\hat{\bz}_i\}_{i=1}^N; \btheta)
\end{eqnarray}
The encoder consists of cross-attention blocks that estimate features by taking into account both the surface samples and medial points so that  the  medial latents encode surface information. The decoder implements Eq. \ref{eq:decoder}, i.e., it computes surface points based on the radius function, given medial points and latents. The encoder and decoder parameters
are trained to minimize the surface loss of Eq. \ref{eq:surface_loss} along with a KL regularization loss, as commonly used in variational autoencoders \cite{Kingma2014}.

\section{Results \& Applications}
\label{sec:results}
\label{sec:validation}

We discuss the experiments and validation of our method for three applications: category-conditioned shape generation, surface reconstruction from point clouds, and skeleton-driven shape generation. 

\subsection{Category-driven shape generation}

In this application, the input is a given category (e.g., ``lamp'', ``chair'', and so on) and the output is a sample set of 3D generated shapes from this category. Our method here makes use of the diffusion stages discussed in Section \ref{sec:generative}: the first stage generates a point-based MAT given the input category and the second stage generates the medial latents given the input category. Then our surface decoder generates the surface under the guidance of the medial points and latents (Section \ref{sec:surface_decoder}). 

\paragraph{Baselines and metrics}
We compare our method with the following recent neural 3D\ generative methods: the autoregressive transformer-based \emph{3DILG} model \cite{Zhang_3dilg} and latent diffusion-based \emph{3DS2VS} \cite{3DShape2VecSet}. For both competing methods, we use the source code provided by the authors.  

As generative evaluation metrics, we employ the metrics also used in 3DS2VS. First, we report the Maximum Mean Discrepancy  based on Chamfer Distance (\textbf{MMD-CD}) and Earth Mover's Distance (\textbf{MMD-EMD}) -- the MMD metrics quantify the \emph{fidelity} of generated examples i.e., how well the generated shapes match the reference ShapeNet test splits. In addition, we report \emph{coverage} measured as  the fraction of the generated shapes matching the reference test splits in terms of  Chamfer distance (\textbf{COV-CD}) and Earth Mover's Distance (\textbf{COV-EMD}). For implementation details of these metrics, we refer readers to \cite{achlioptas2018learning} who  introduced them in the context of 3D generation. 

 In addition, as proposed in the literature of image generative models, we use the Fr\'echet Inception Distance and Kernel Inception Distance. While prior work \cite{3DShape2VecSet,Zhang_3dilg} relies on PointNet++ embeddings to evaluate these metrics, we use the 768-D embeddings taken by the more recent and more discriminative model of PointBert \cite{yu2022point}. We refer to these two metrics as  \textbf{PointBert-FID} and \textbf{PointBert-KID}. Finally, we report generative \textbf{precision} and \textbf{recall} \cite{precision_recall_distributions}  as alternative measures to assess how well generated data covers the reference test splits and vice versa. For precision and recall, the similarities are assessed via PointBert embeddings. 

\paragraph{Experimental setup}
We follow the experimental setup of 3DS2VS \cite{3DShape2VecSet}, which was also demonstrated in the same application. We use the same dataset of ShapeNet-v2 \cite{chang2015shapenet} with the same splits. We use the same watertight ShapeNet meshes provided by \cite{Zhang_3dilg}. In constrast to 3DS2VS that evaluated category-conditioned generation in $2$ categories,  in our case we evaluate on the largest $10$ categories from ShapeNet, namely: tables, cars, chairs, airplanes, sofas, rifles, lamps, watercrafts, benches, and speakers. To further improve our evaluation, for each method and category we generate a number of samples equal to $3$x the size of the ground truth test split per category, which is also consistent with Achlioptas et al. \shortcite{achlioptas2018learning} originally proposed.

\paragraph{Quantitative results}
Table \ref{tab:generative_eval} shows quantitative evaluation of all competing methods for all above measures. The measures are averaged over the $10$ largest categories of ShapeNet. For all metrics, our method outperforms the competing work demonstrating results of better fidelity, coverage, precision and recall. In addition, we observe that according to FID/KID, our method outputs samples whose feature distributions match better the reference splits. Our supplement includes per-category evaluation for all measures -- again, for the majority of categories, \method outperforms the competing methods for all metrics. In particular, we observe that the gap widens for categories containing shapes with thin or tubular-like parts, such as benches, rifles, airplanes, chairs, and tables.
\begin{table}[t!]

\begin{tabular}{r|c|c|c}
\toprule
Metric     & 3DILG & 3DS2VS & Ours \\
\midrule
MMD-CD ($\times 10^{2}$) ($\downarrow$) & 9.64 &  9.34  & \textbf{8.64}\\
MMD-EMD ($\times 10^{2}$) ($\downarrow$)& 11.0 &  10.9 & \textbf{10.1}\\
COV-CD ($\times 10^{2}$) ($\uparrow$)   & 54.8  & 50.6   & \textbf{58.5}\\
COV-EMD ($\times 10^{2}$) ($\uparrow$)  & 53.1  & 49.9   & \textbf{57.2}\\
precision (\%) ($\uparrow$)             & 73.1  & 72.4   & \textbf{80.6}\\
recall (\%) ($\uparrow$)                & 79.4  & 80.3   & \textbf{89.2}\\
PointBert-FID ($\times 10^{-2}$) ($\downarrow$)   & 1.46  & 1.62   & \textbf{1.25}\\
PointBert-KID ($\times 1$) ($\downarrow$)         & 4.06  & 4.77   & \textbf{2.54}\\
\bottomrule
\end{tabular}
\caption{Evaluation measures on category-conditioned shape generation. The measures are averaged over the $10$ largest categories of ShapeNet. For evaluation per-category, please see the supplement.  }
\label{tab:generative_eval}
\vspace{-10mm}
\end{table}

\begin{figure*}[!ht]
\setlength\tabcolsep{-1pt}
\begin{tabular}{ccccccccccccccc}
\includegraphics[width=.09\linewidth,valign=m]{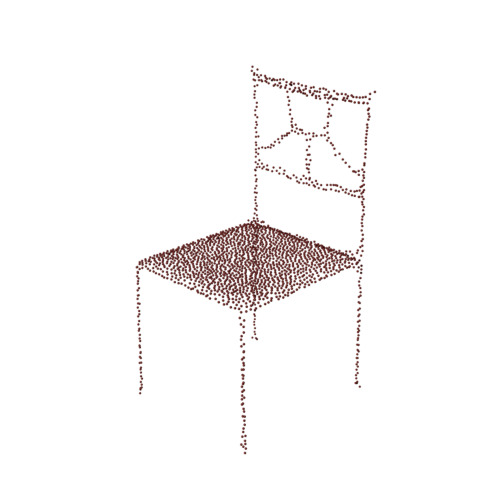} & 
\includegraphics[width=.09\linewidth,valign=m]{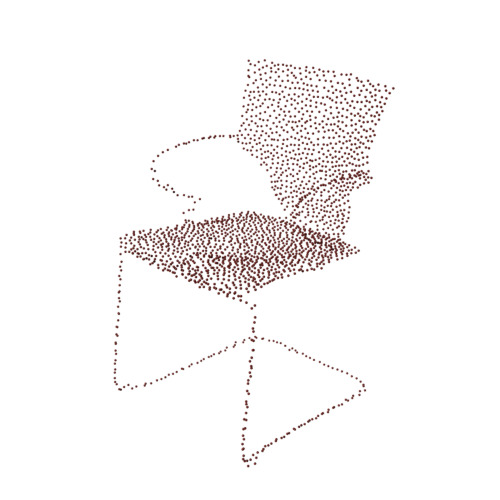} & 
\includegraphics[width=.09\linewidth,valign=m]{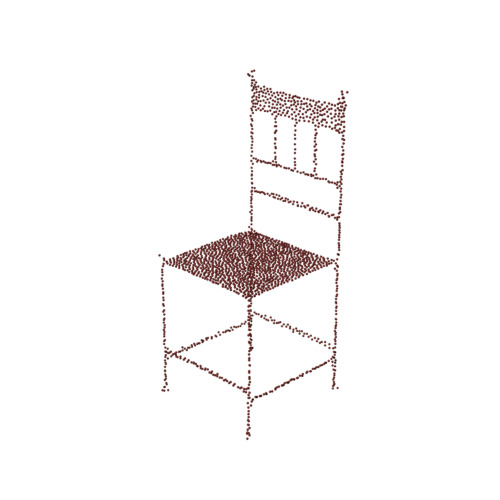} & 
\includegraphics[width=.09\linewidth,valign=m]{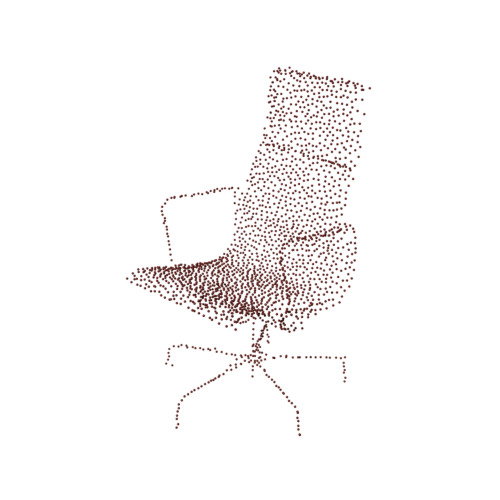} & 
\includegraphics[width=.09\linewidth,valign=m]{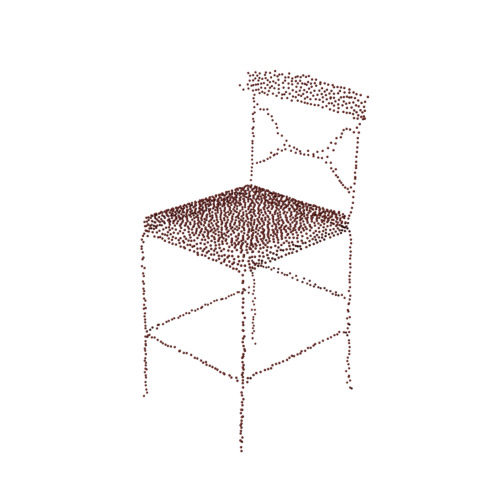} & 
\includegraphics[width=.09\linewidth,valign=m]{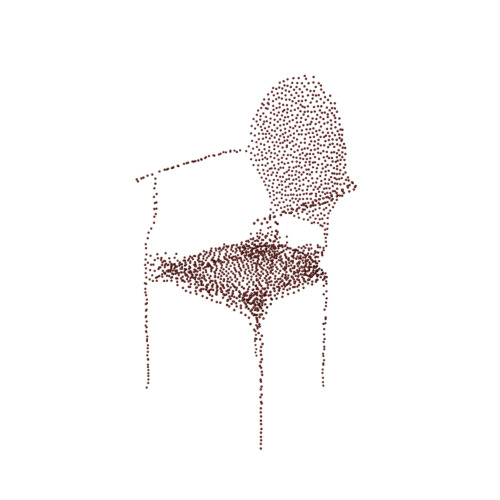} & 
\includegraphics[width=.09\linewidth,valign=m]{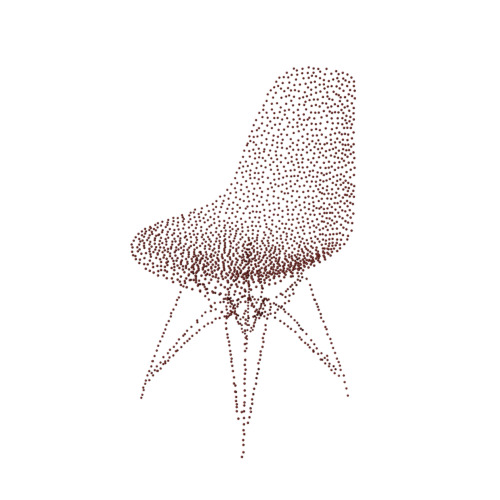} & 
\includegraphics[width=.09\linewidth,valign=m]{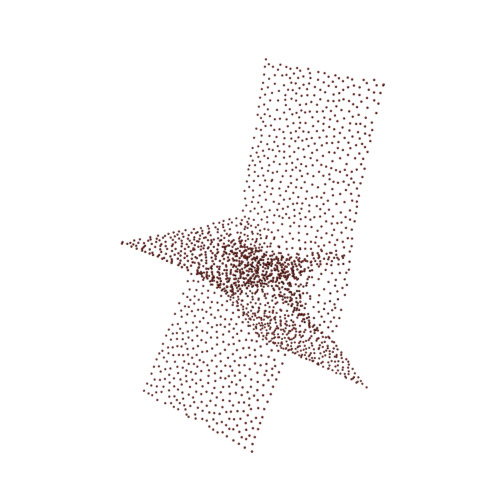} & 
\includegraphics[width=.09\linewidth,valign=m]{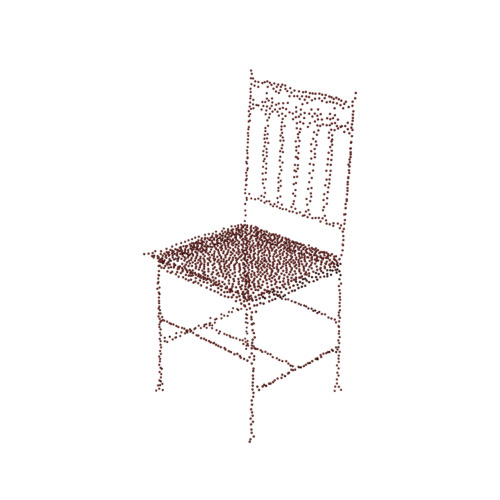} & 
\includegraphics[width=.09\linewidth,valign=m]{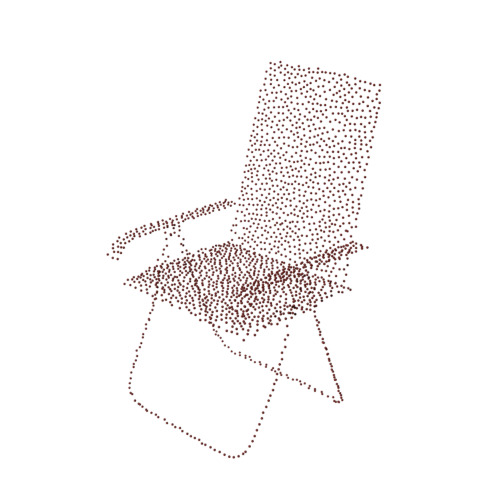} & 
\includegraphics[width=.09\linewidth,valign=m]{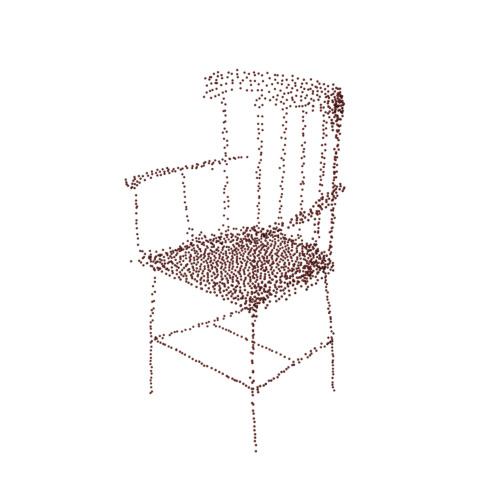} & 
\includegraphics[width=.09\linewidth,valign=m]{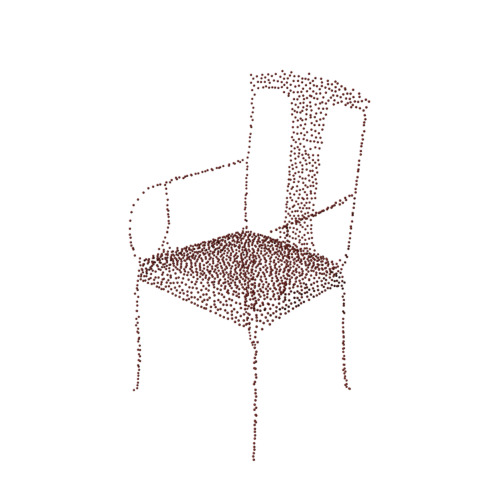} & 
 \\
\includegraphics[width=.09\linewidth,valign=m]{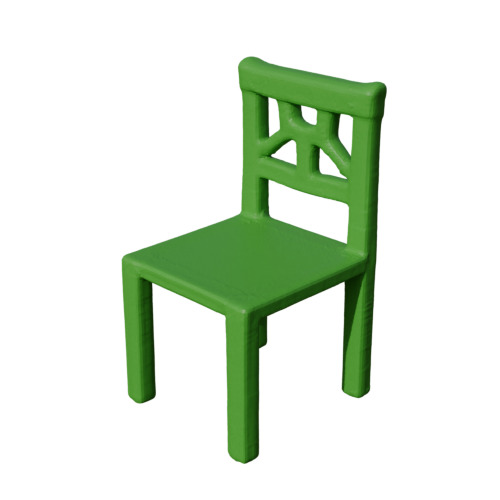} & 
\includegraphics[width=.09\linewidth,valign=m]{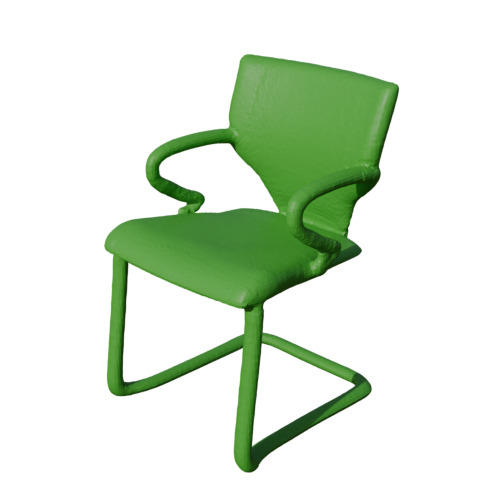} & 
\includegraphics[width=.09\linewidth,valign=m]{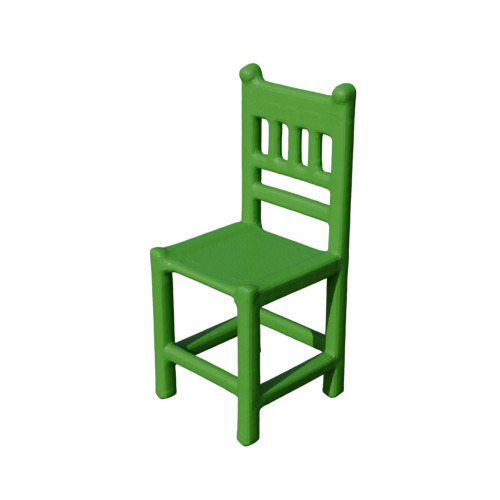} & 
\includegraphics[width=.09\linewidth,valign=m]{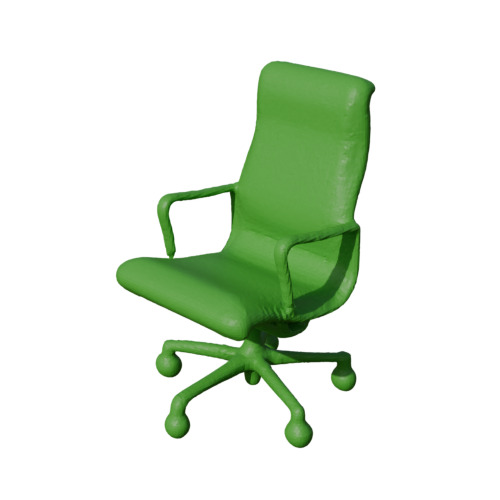} & 
\includegraphics[width=.09\linewidth,valign=m]{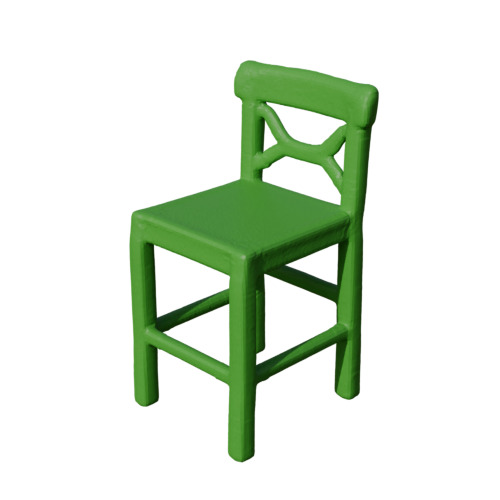} & 
\includegraphics[width=.09\linewidth,valign=m]{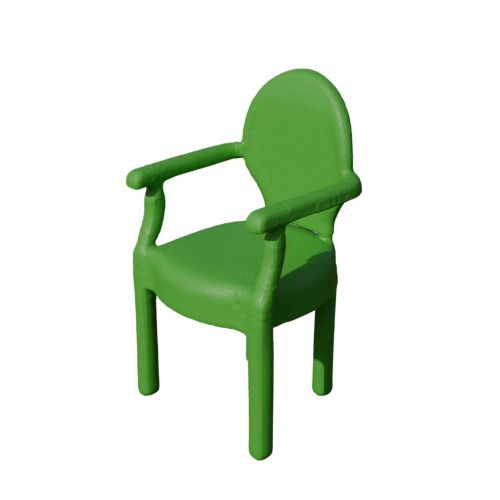} & 
\includegraphics[width=.09\linewidth,valign=m]{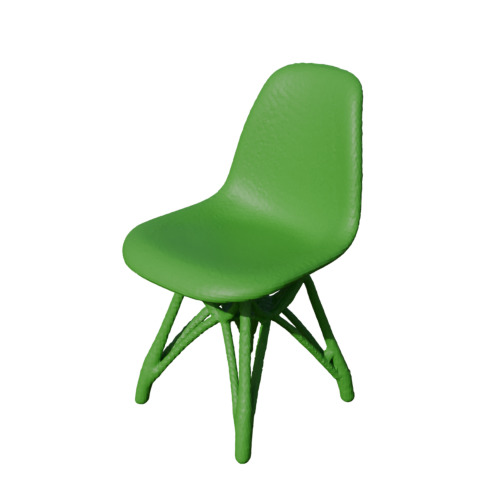} & 
\includegraphics[width=.09\linewidth,valign=m]{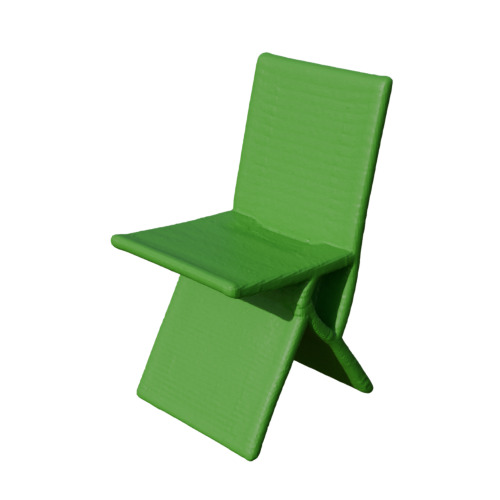} & 
\includegraphics[width=.09\linewidth,valign=m]{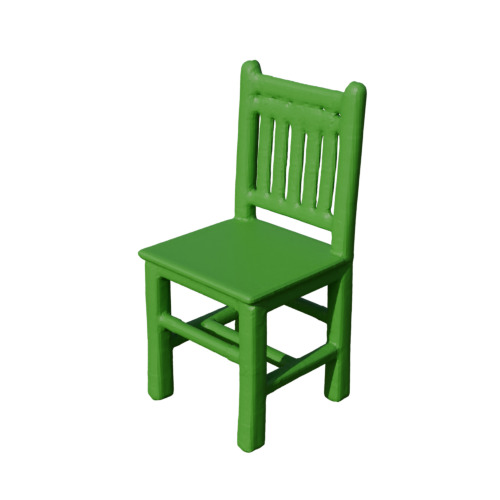} & 
\includegraphics[width=.09\linewidth,valign=m]{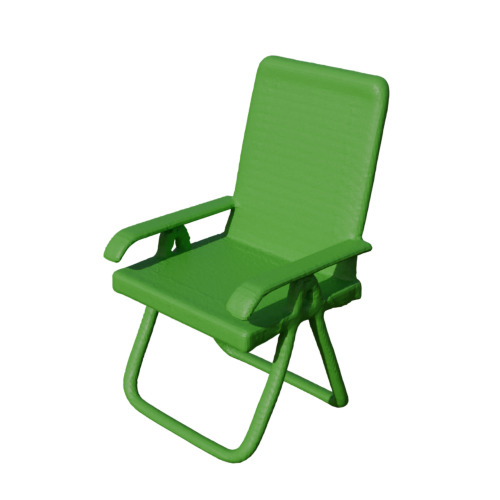} & 
\includegraphics[width=.09\linewidth,valign=m]{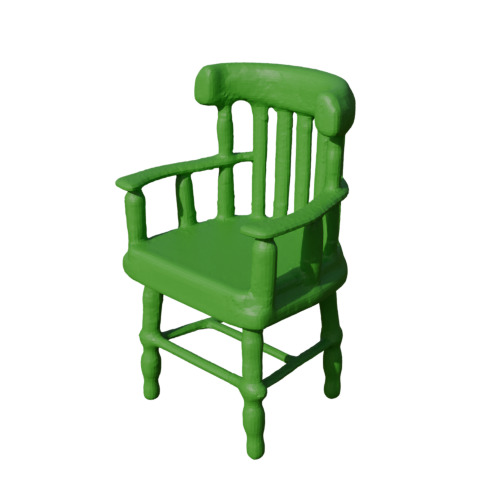} & 
\includegraphics[width=.09\linewidth,valign=m]{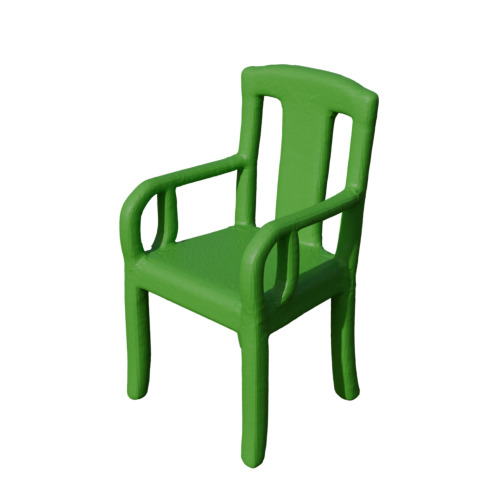} & 
 \\

\hline
\includegraphics[width=.09\linewidth,valign=m]{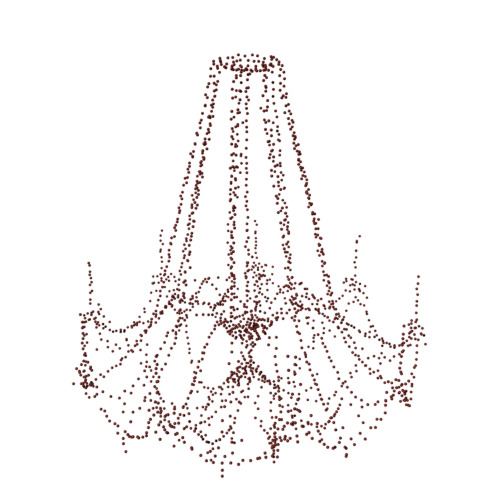} & 
\includegraphics[width=.09\linewidth,valign=m]{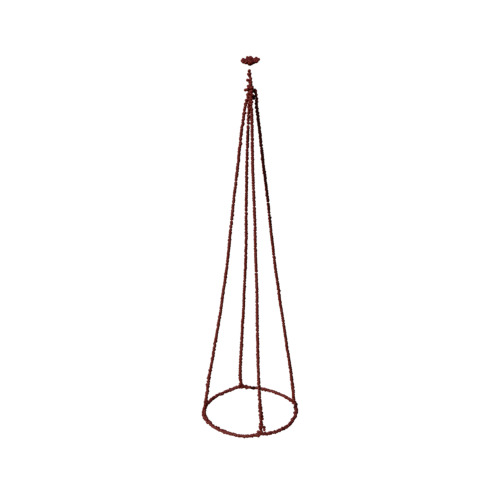} & 
\includegraphics[width=.09\linewidth,valign=m]{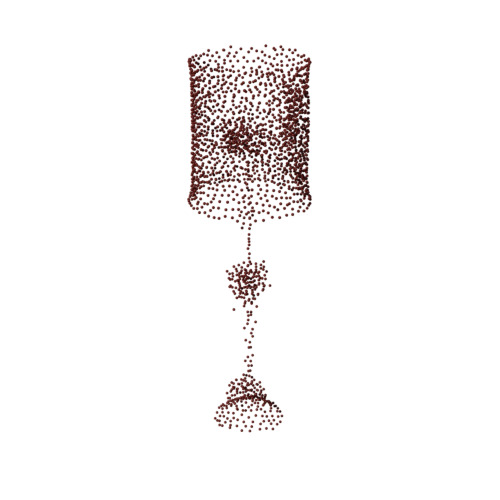} & 
\includegraphics[width=.09\linewidth,valign=m]{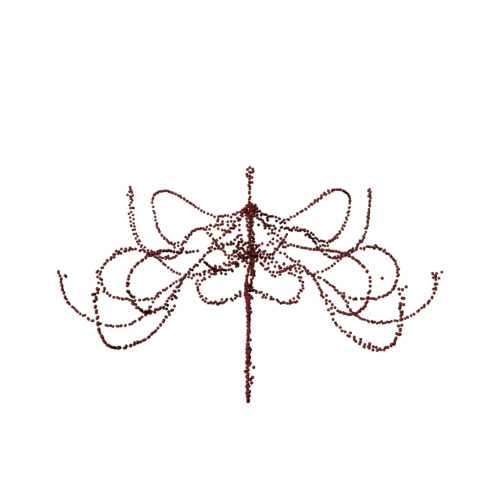} & 
\includegraphics[width=.09\linewidth,valign=m]{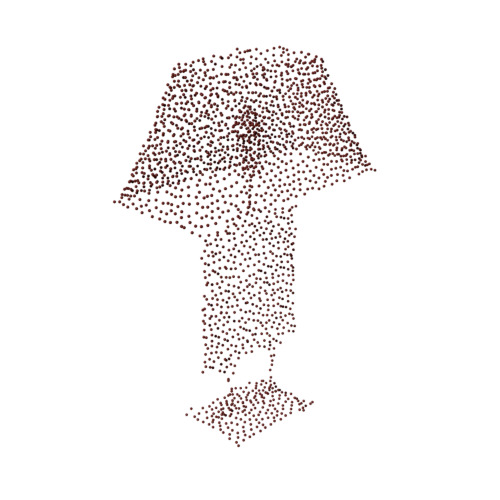} & 
\includegraphics[width=.09\linewidth,valign=m]{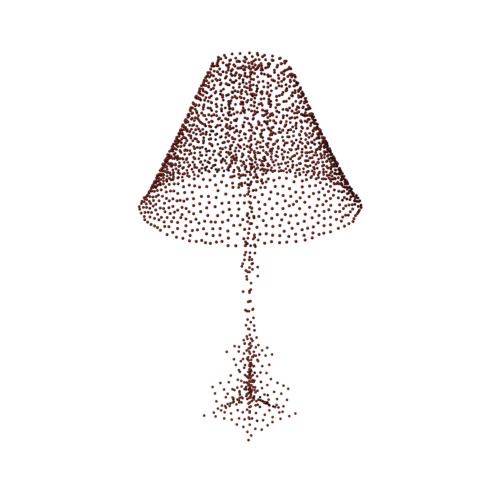} & 
\includegraphics[width=.09\linewidth,valign=m]{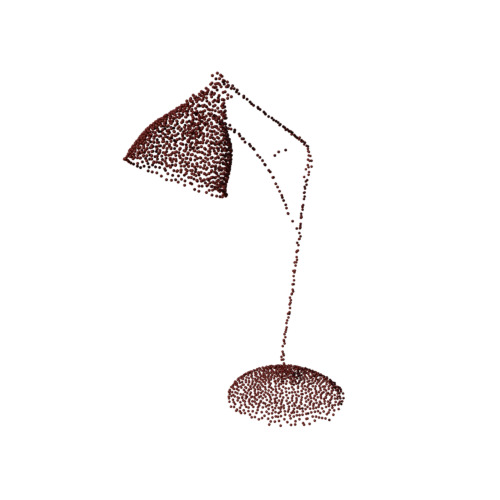} & 
\includegraphics[width=.09\linewidth,valign=m]{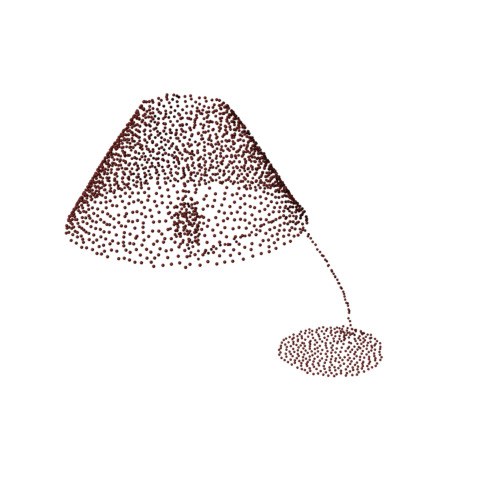} & 
\includegraphics[width=.09\linewidth,valign=m]{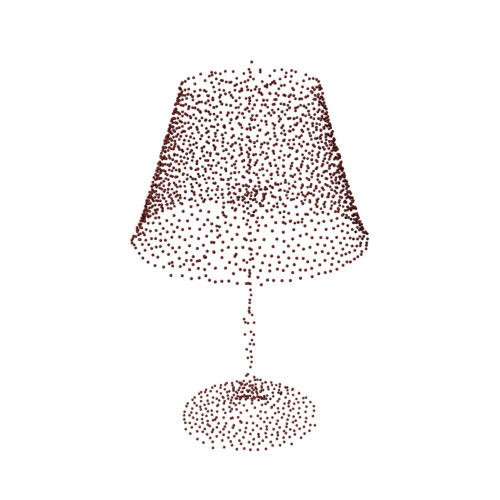} & 
\includegraphics[width=.09\linewidth,valign=m]{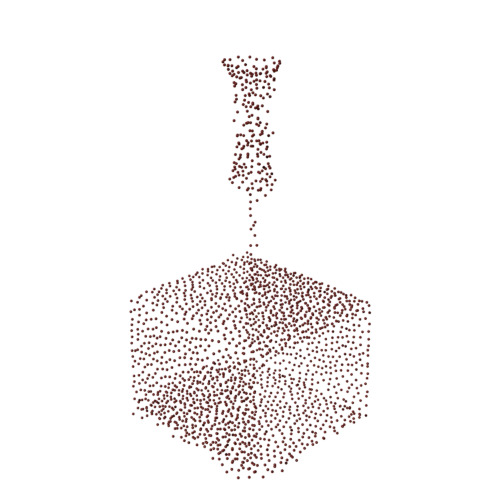} & 
\includegraphics[width=.09\linewidth,valign=m]{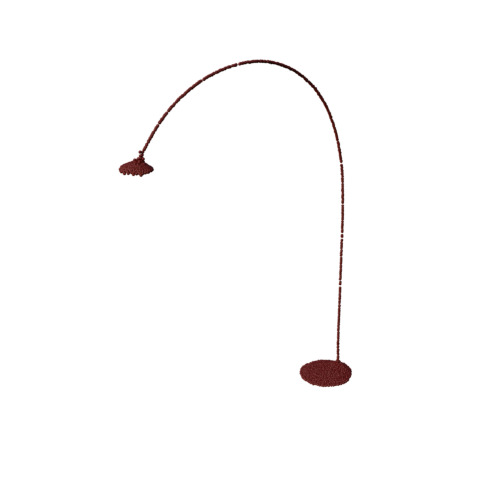} & 
\includegraphics[width=.09\linewidth,valign=m]{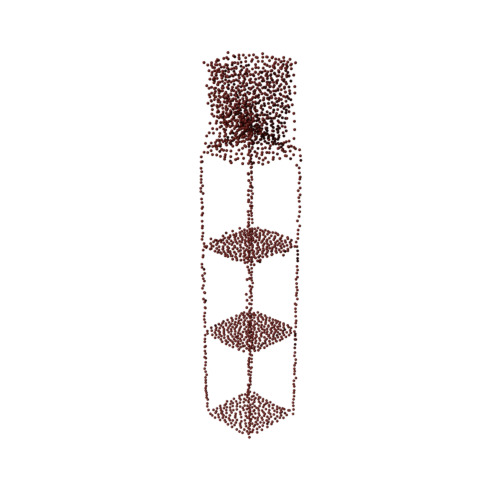} & \\
\includegraphics[width=.09\linewidth,valign=m]{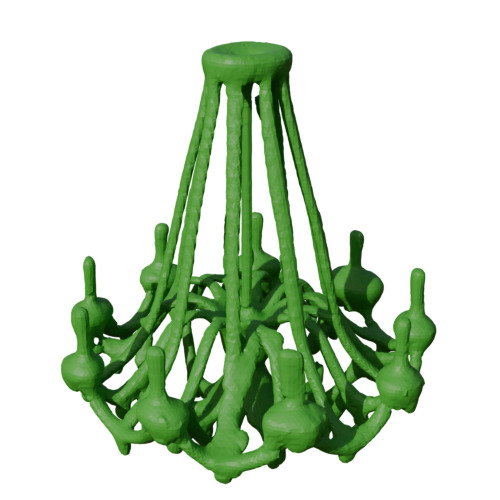} & 
\includegraphics[width=.09\linewidth,valign=m]{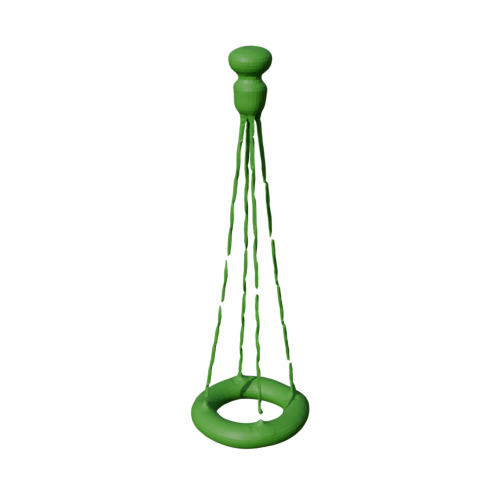} & 
\includegraphics[width=.09\linewidth,valign=m]{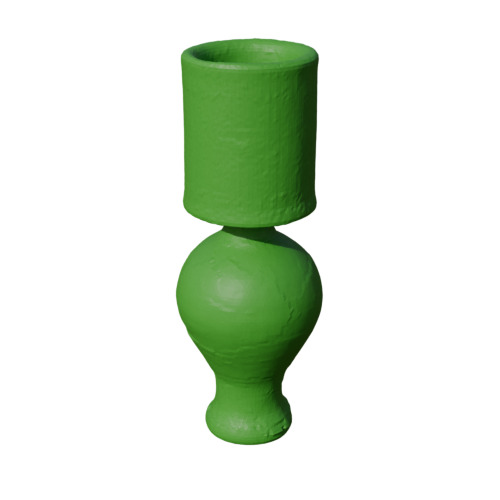} & 
\includegraphics[width=.09\linewidth,valign=m]{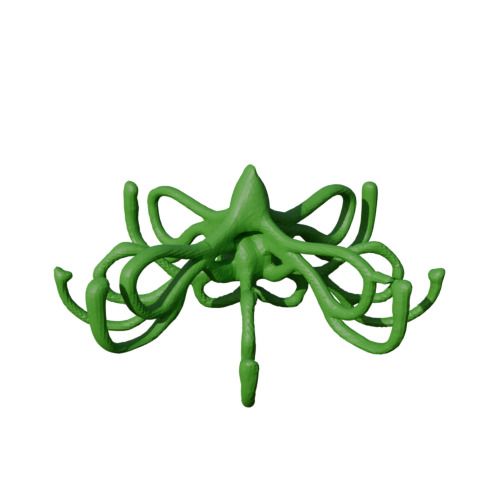} & 
\includegraphics[width=.09\linewidth,valign=m]{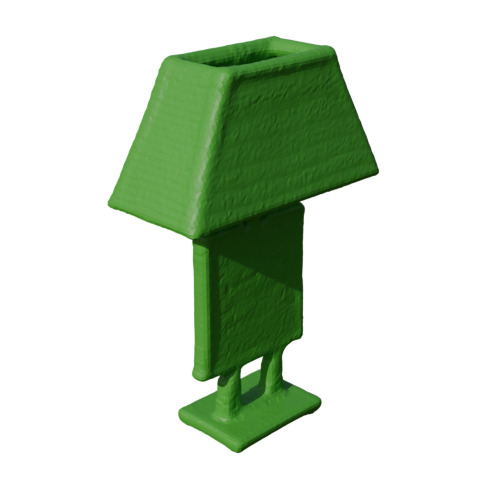} & 
\includegraphics[width=.09\linewidth,valign=m]{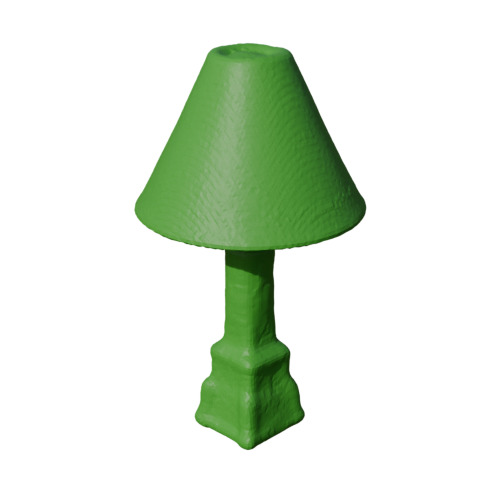} & 
\includegraphics[width=.09\linewidth,valign=m]{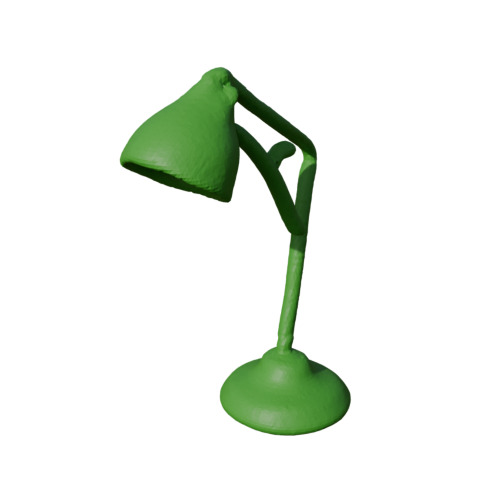} & 
\includegraphics[width=.09\linewidth,valign=m]{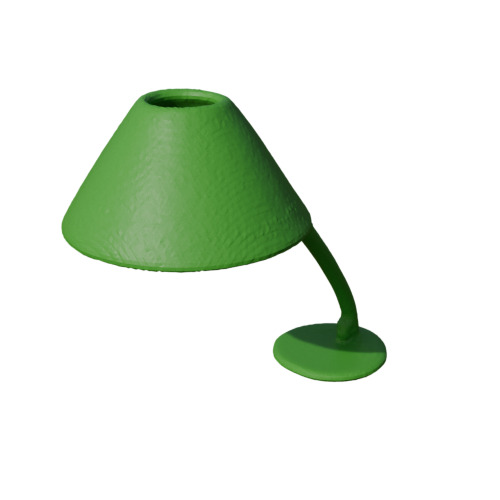} & 
\includegraphics[width=.09\linewidth,valign=m]{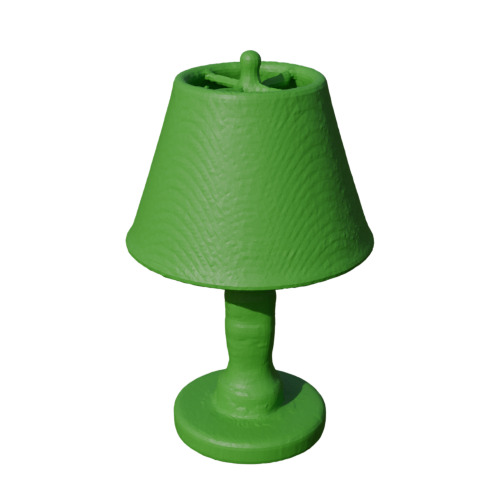} & 
\includegraphics[width=.09\linewidth,valign=m]{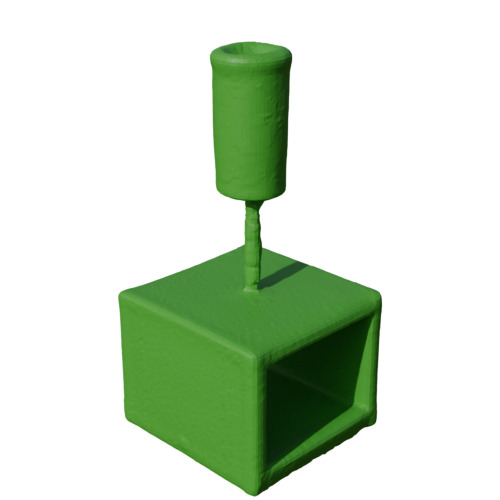} & 
\includegraphics[width=.09\linewidth,valign=m]{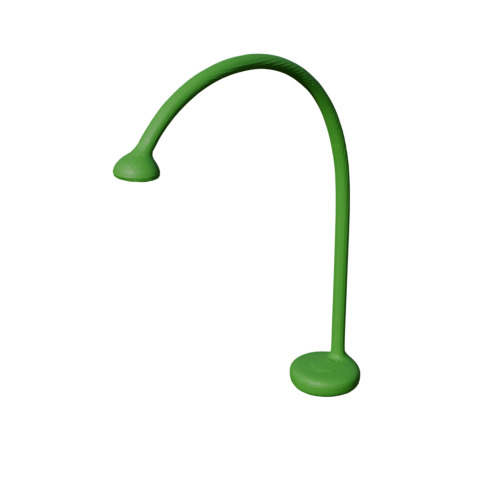} & 
\includegraphics[width=.09\linewidth,valign=m]{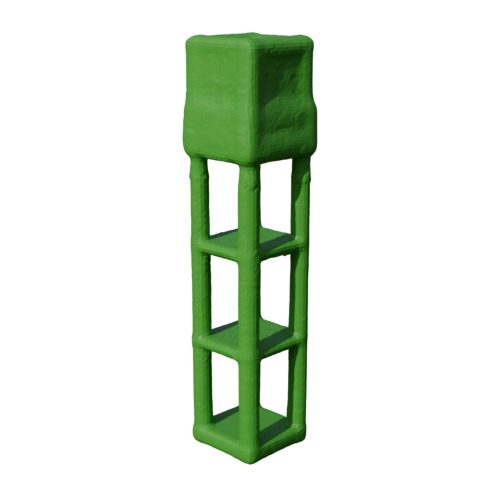} & \\

\hline
\includegraphics[width=.09\linewidth,valign=m]{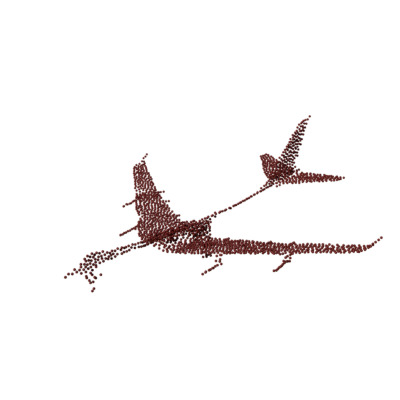} & 
\includegraphics[width=.09\linewidth,valign=m]{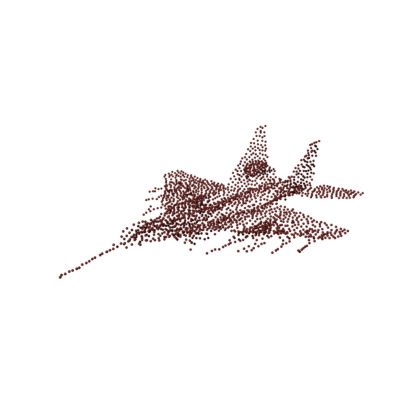} & 
\includegraphics[width=.09\linewidth,valign=m]{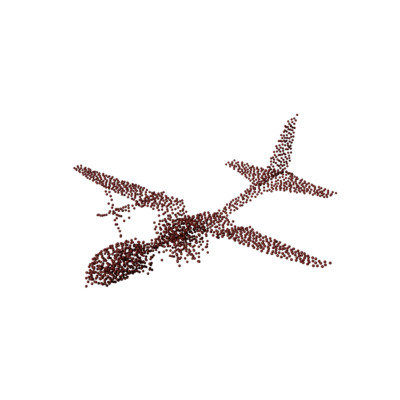} & 
\includegraphics[width=.09\linewidth,valign=m]{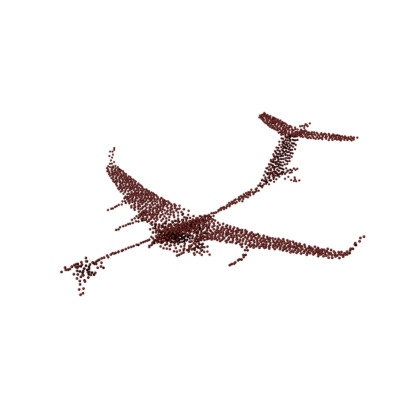} & 
\includegraphics[width=.09\linewidth,valign=m]{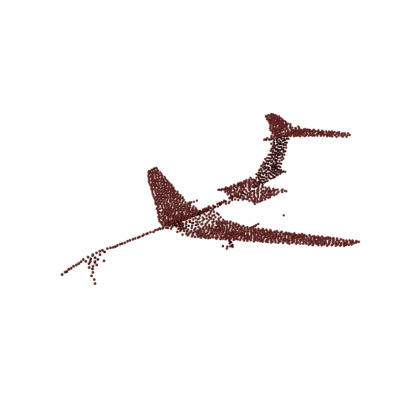} & 
\includegraphics[width=.09\linewidth,valign=m]{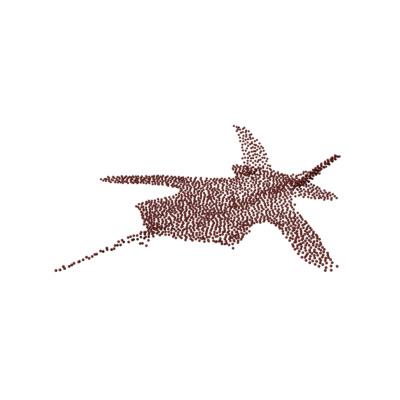} & 
\includegraphics[width=.09\linewidth,valign=m]{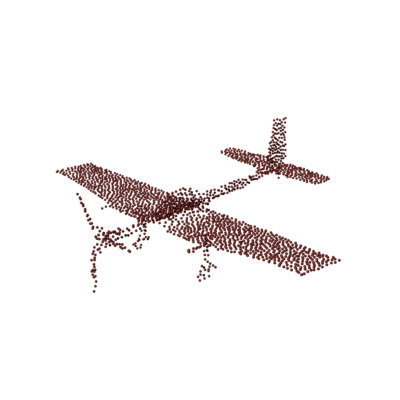} & 
\includegraphics[width=.09\linewidth,valign=m]{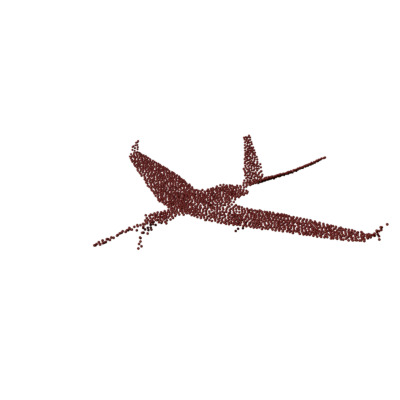} & 
\includegraphics[width=.09\linewidth,valign=m]{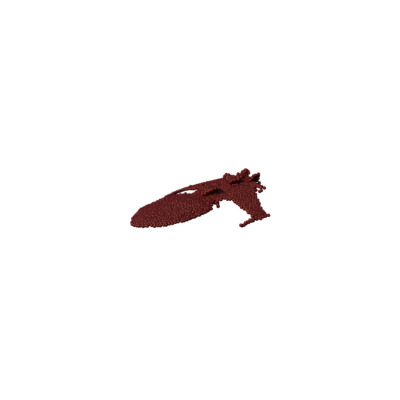} & 
\includegraphics[width=.09\linewidth,valign=m]{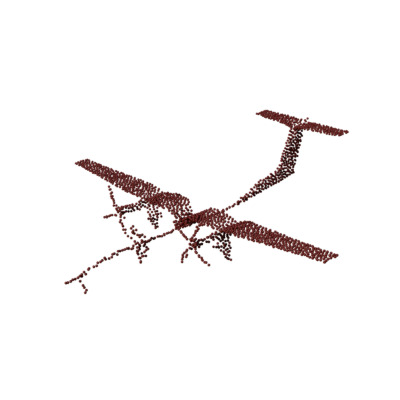} & 
\includegraphics[width=.09\linewidth,valign=m]{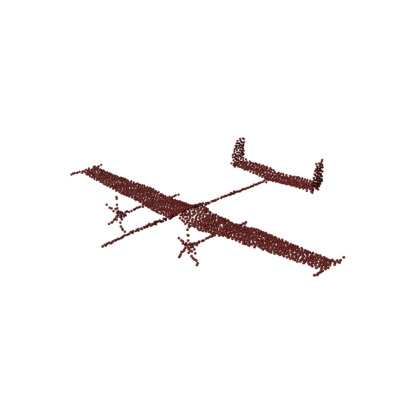} & 
\includegraphics[width=.09\linewidth,valign=m]{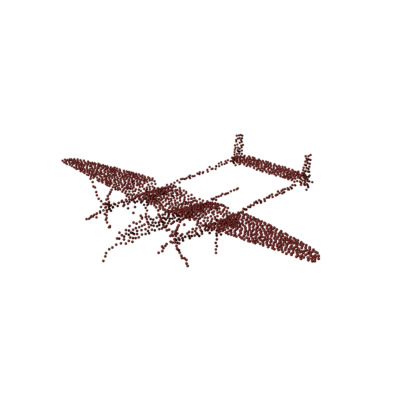} & \\
\includegraphics[width=.09\linewidth,valign=m]{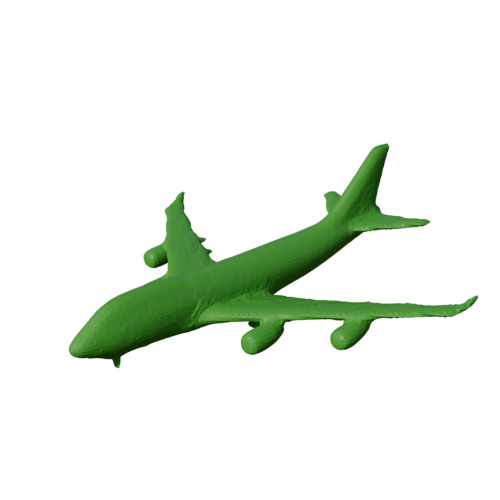} & 
\includegraphics[width=.09\linewidth,valign=m]{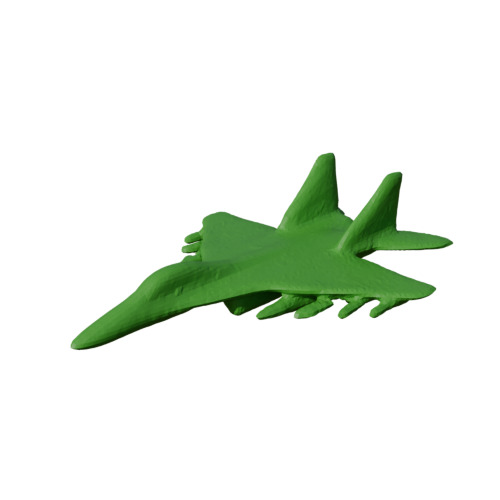} & 
\includegraphics[width=.09\linewidth,valign=m]{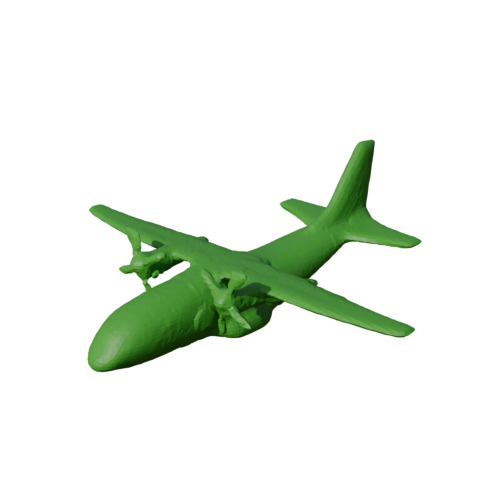} & 
\includegraphics[width=.09\linewidth,valign=m]{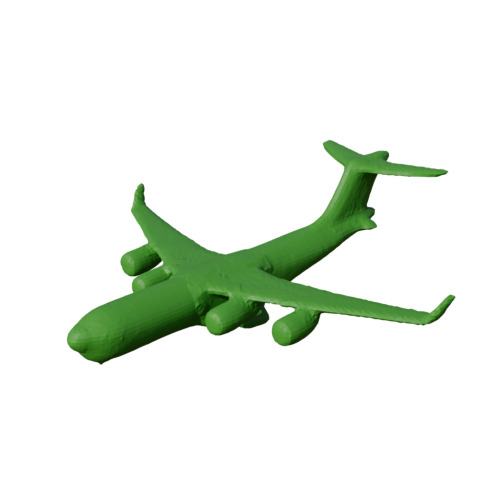} & 
\includegraphics[width=.09\linewidth,valign=m]{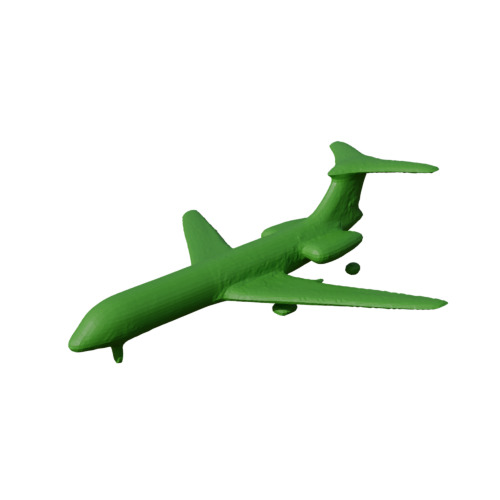} & 
\includegraphics[width=.09\linewidth,valign=m]{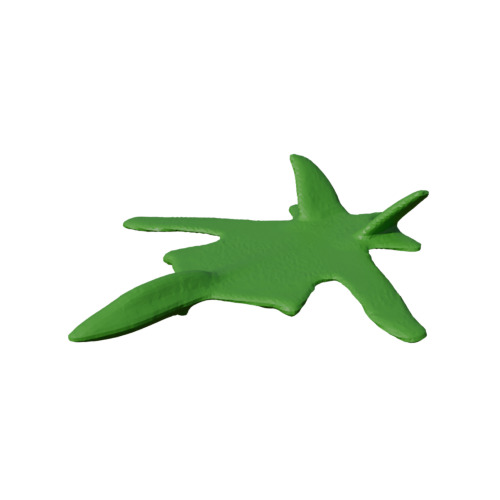} & 
\includegraphics[width=.09\linewidth,valign=m]{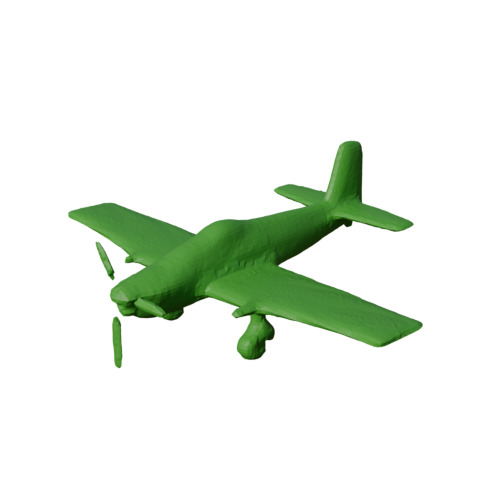} & 
\includegraphics[width=.09\linewidth,valign=m]{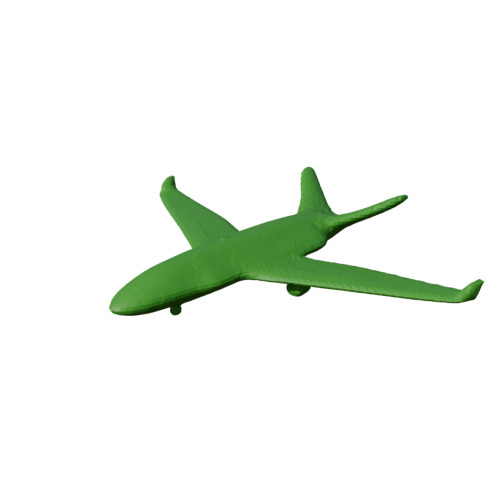} & 
\includegraphics[width=.09\linewidth,valign=m]{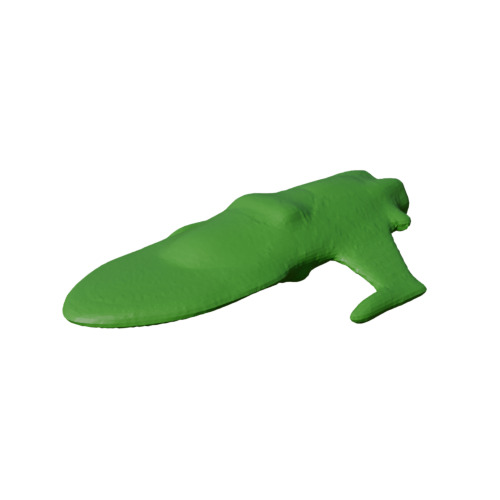} & 
\includegraphics[width=.09\linewidth,valign=m]{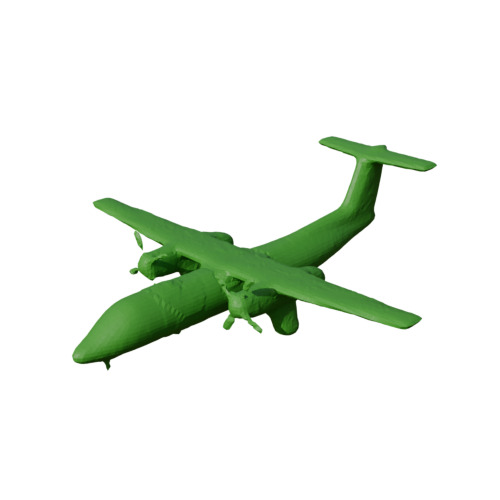} & 
\includegraphics[width=.09\linewidth,valign=m]{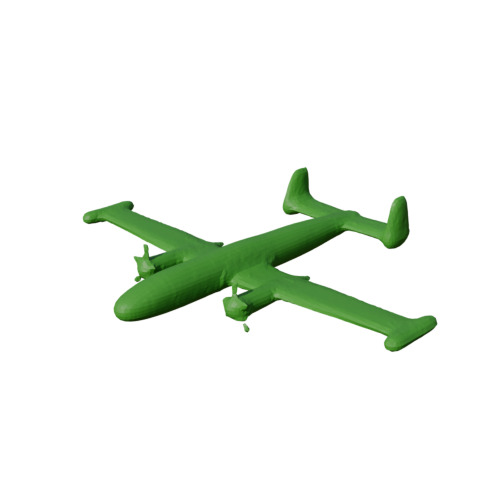} & 
\includegraphics[width=.09\linewidth,valign=m]{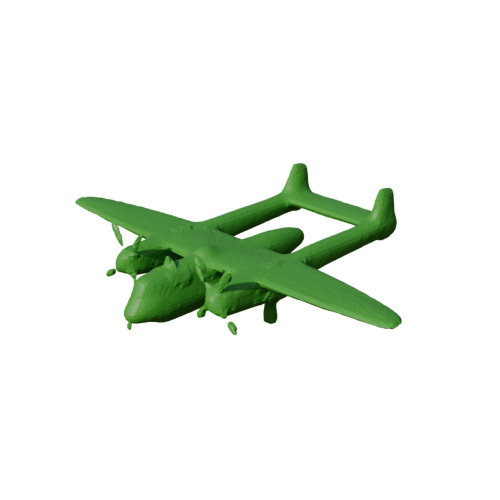} & \\
\hline
\includegraphics[width=.09\linewidth,valign=m]{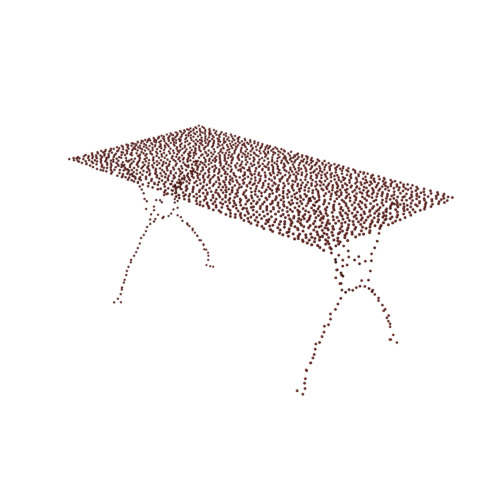} & 
\includegraphics[width=.09\linewidth,valign=m]{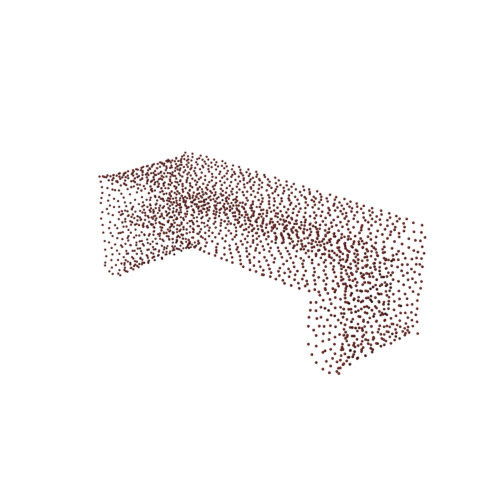} & 
\includegraphics[width=.09\linewidth,valign=m]{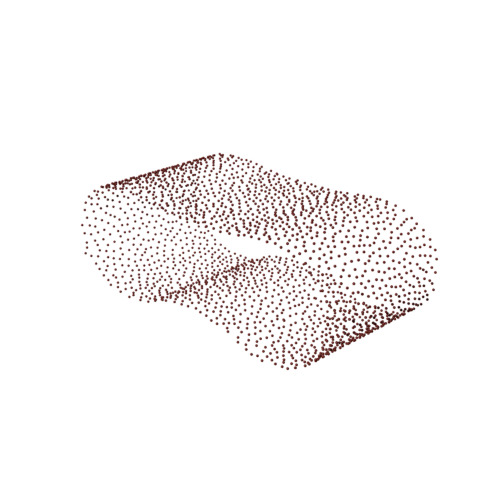} & 
\includegraphics[width=.09\linewidth,valign=m]{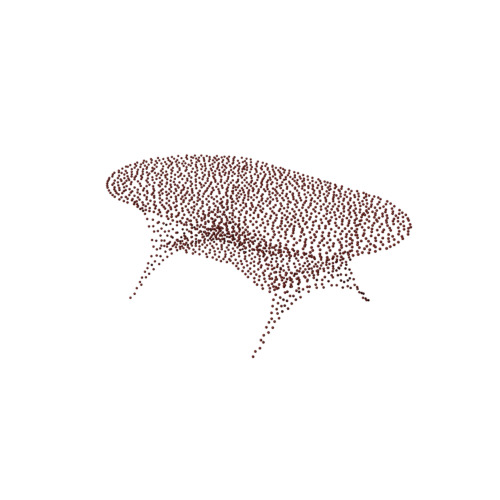} & 
\includegraphics[width=.09\linewidth,valign=m]{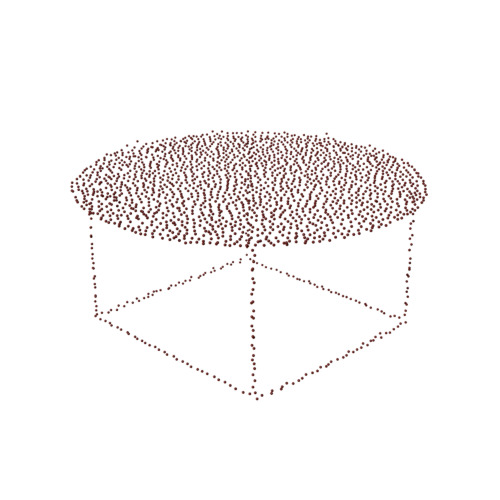} & 
\includegraphics[width=.09\linewidth,valign=m]{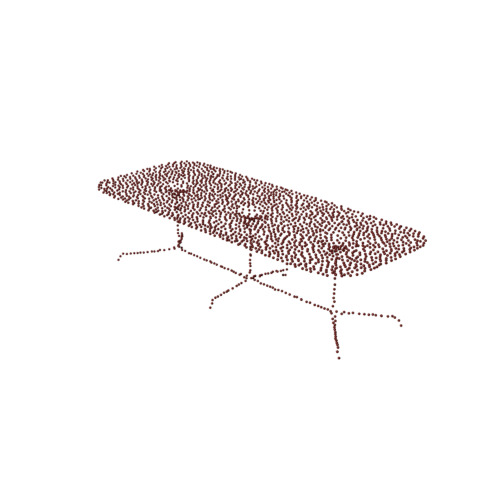} & 
\includegraphics[width=.09\linewidth,valign=m]{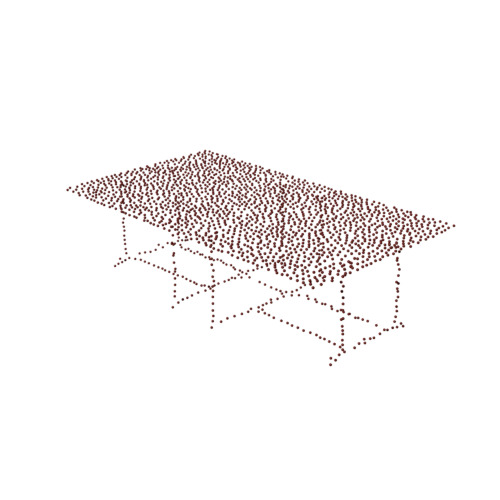} & 
\includegraphics[width=.09\linewidth,valign=m]{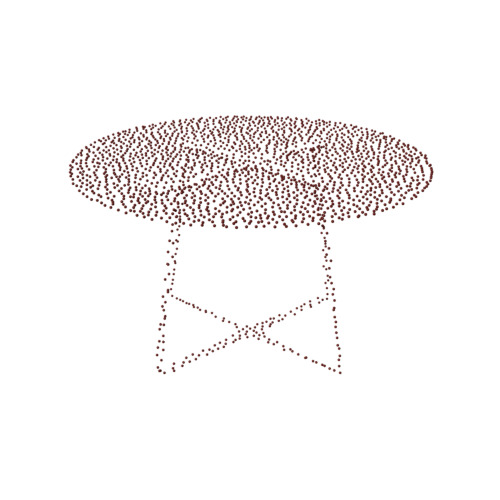} & 
\includegraphics[width=.09\linewidth,valign=m]{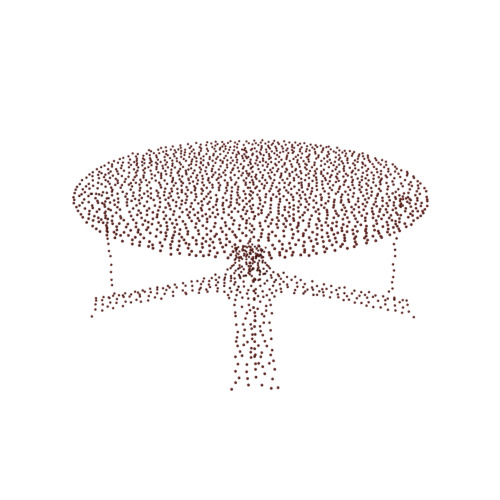} & 
\includegraphics[width=.09\linewidth,valign=m]{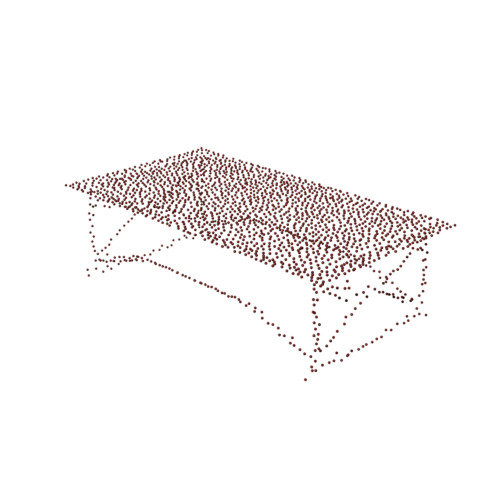} & 
\includegraphics[width=.09\linewidth,valign=m]{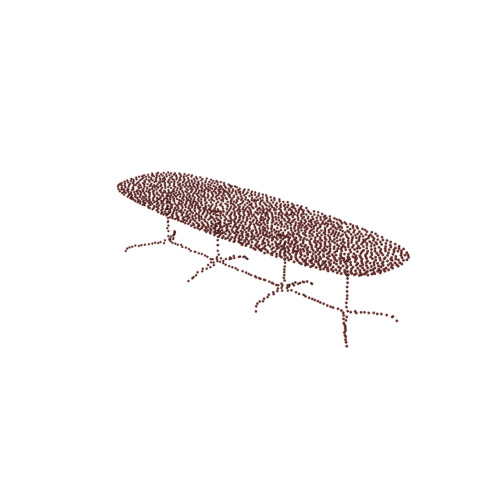} & 
\includegraphics[width=.09\linewidth,valign=m]{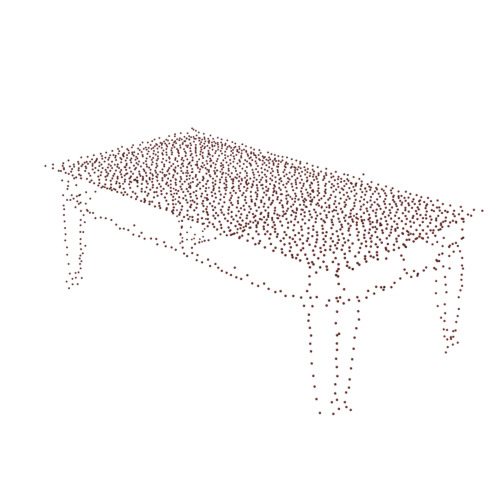} & \\

\includegraphics[width=.09\linewidth,valign=m]{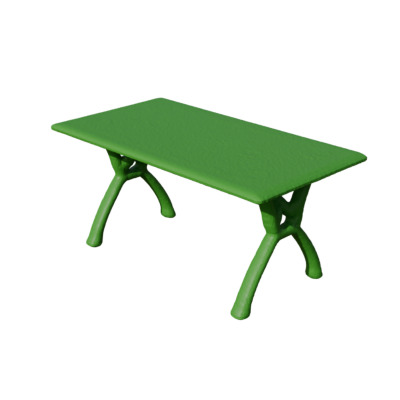} & 
\includegraphics[width=.09\linewidth,valign=m]{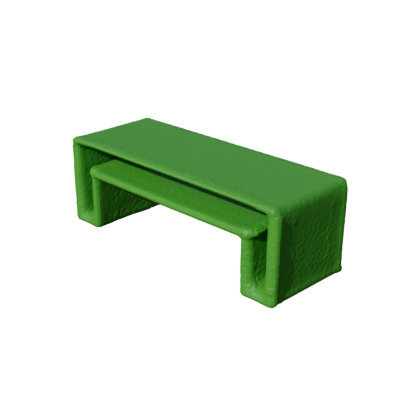} & 
\includegraphics[width=.09\linewidth,valign=m]{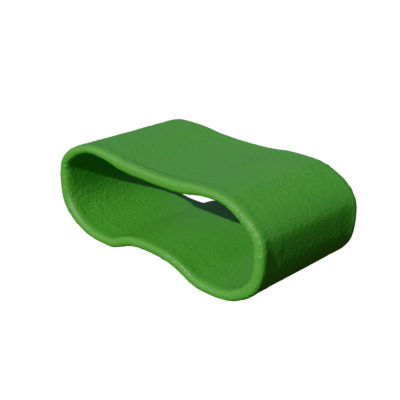} & 
\includegraphics[width=.09\linewidth,valign=m]{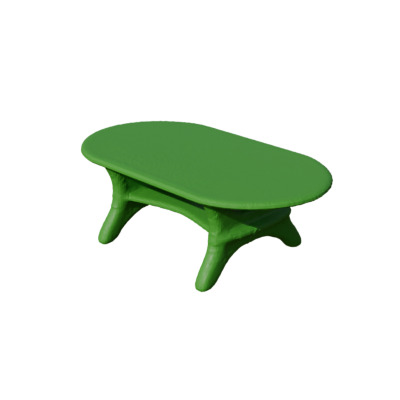} & 
\includegraphics[width=.09\linewidth,valign=m]{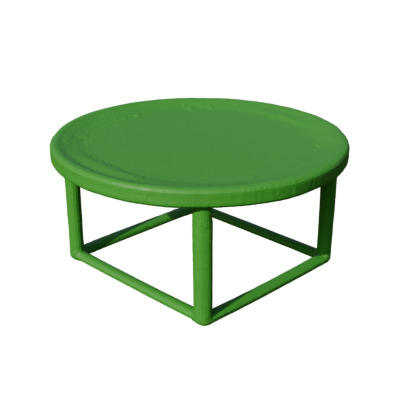} & 
\includegraphics[width=.09\linewidth,valign=m]{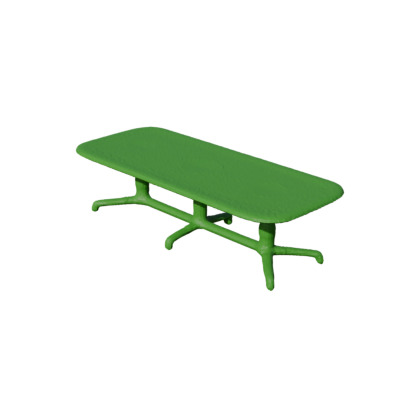} & 
\includegraphics[width=.09\linewidth,valign=m]{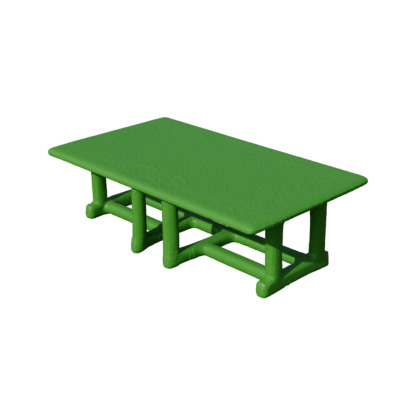} & 
\includegraphics[width=.09\linewidth,valign=m]{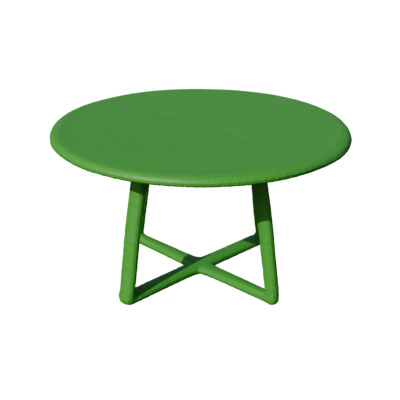} & 
\includegraphics[width=.09\linewidth,valign=m]{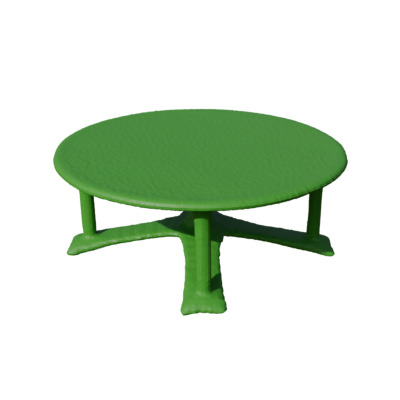} & 
\includegraphics[width=.09\linewidth,valign=m]{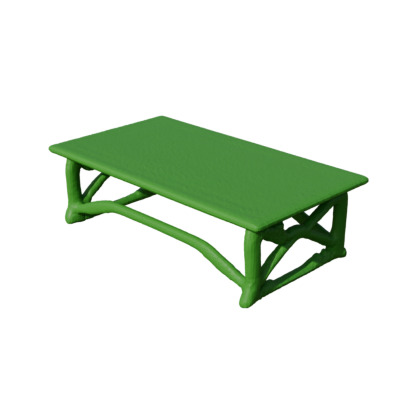} & 
\includegraphics[width=.09\linewidth,valign=m]{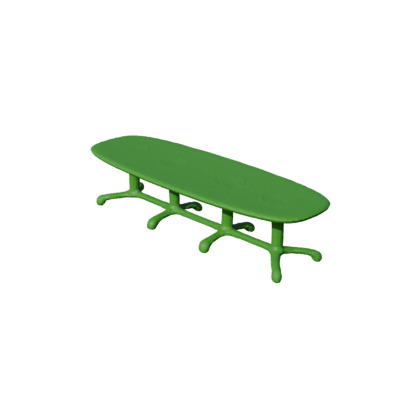} & 
\includegraphics[width=.09\linewidth,valign=m]{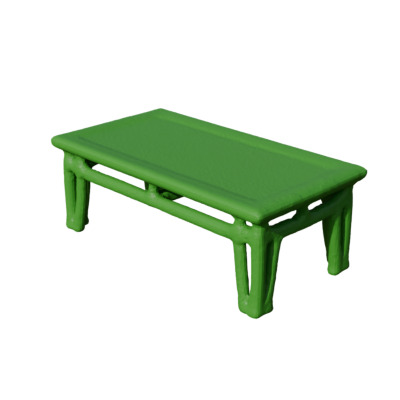} & \\
\hline
\includegraphics[width=.08\linewidth,valign=m]{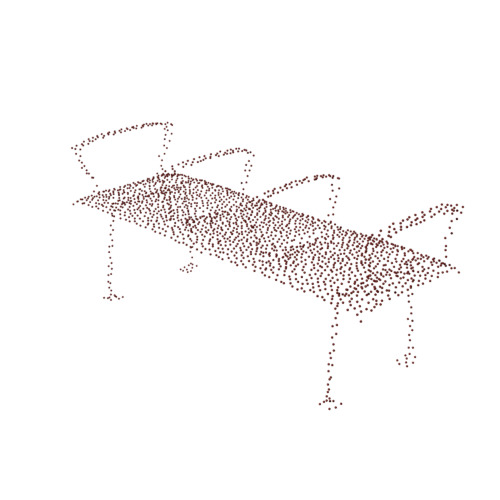} & 
\includegraphics[width=0.08\linewidth,valign=m]{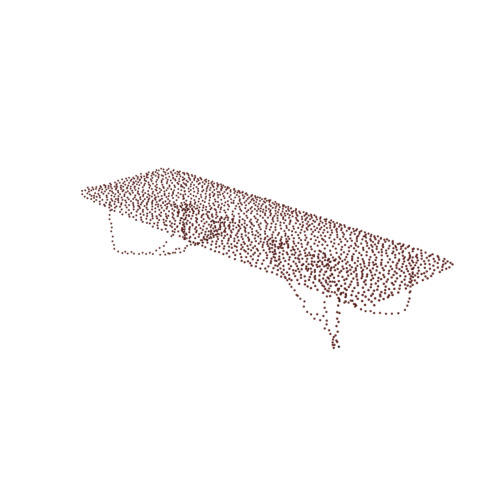} & 
\includegraphics[width=0.08\linewidth,valign=m]{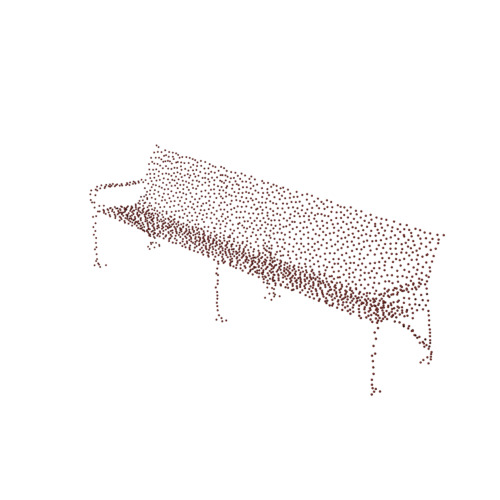} & 
\includegraphics[width=0.08\linewidth,valign=m]{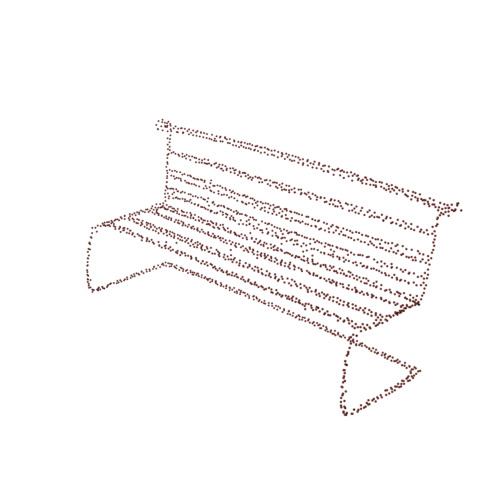} & 
\includegraphics[width=0.08\linewidth,valign=m]{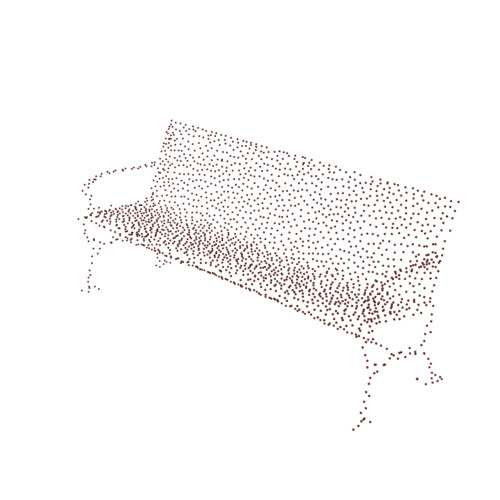} & 
\includegraphics[width=0.08\linewidth,valign=m]{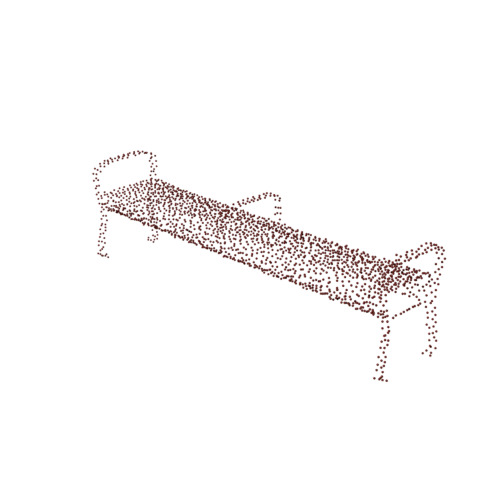} & 
\includegraphics[width=0.08\linewidth,valign=m]{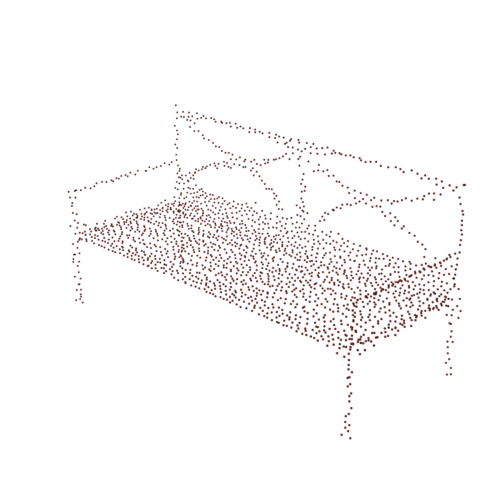} & 
\includegraphics[width=0.08\linewidth,valign=m]{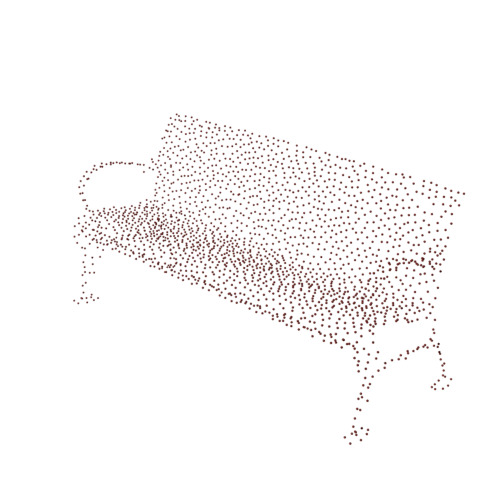} & 
\includegraphics[width=0.08\linewidth,valign=m]{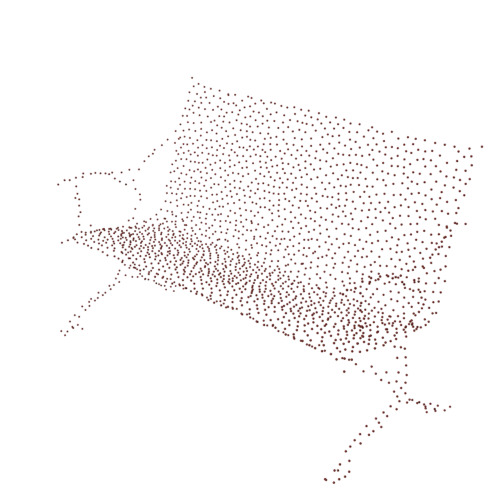} & 
\includegraphics[width=0.08\linewidth,valign=m]{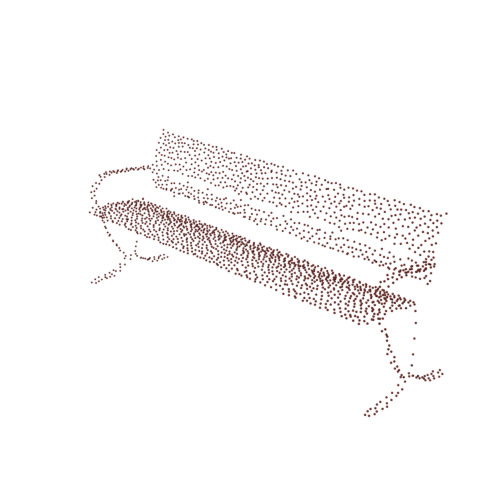} & 
\includegraphics[width=0.08\linewidth,valign=m]{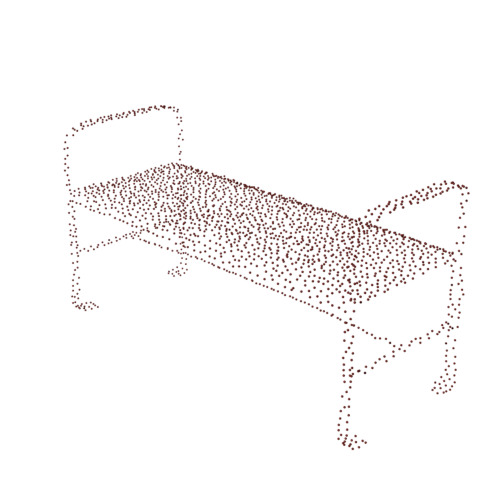} & 
\includegraphics[width=0.08\linewidth,valign=m]{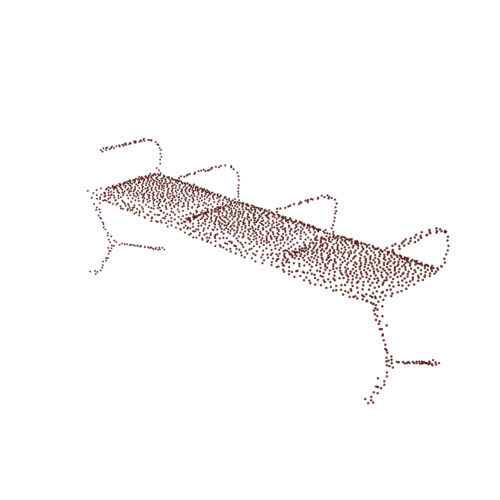} & \\
\includegraphics[width=0.08\linewidth,valign=m]{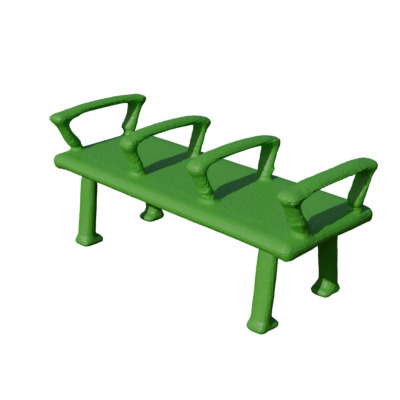} & 
\includegraphics[width=0.08\linewidth,valign=m]{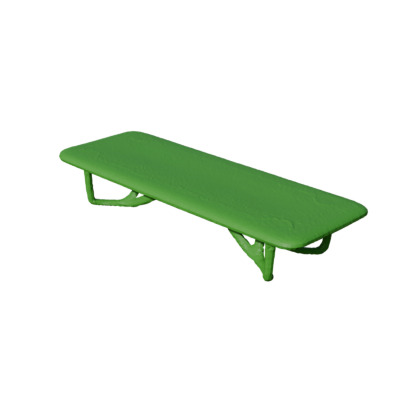} & 
\includegraphics[width=0.08\linewidth,valign=m]{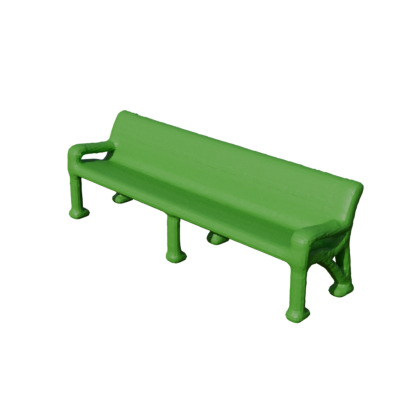} & 
\includegraphics[width=0.08\linewidth,valign=m]{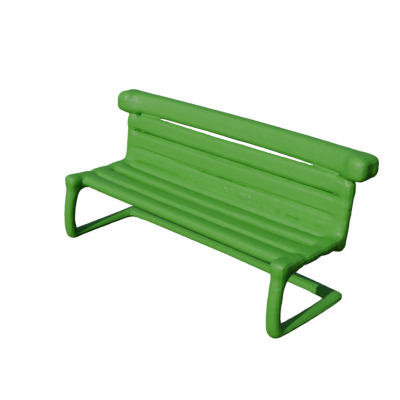} & 
\includegraphics[width=0.08\linewidth,valign=m]{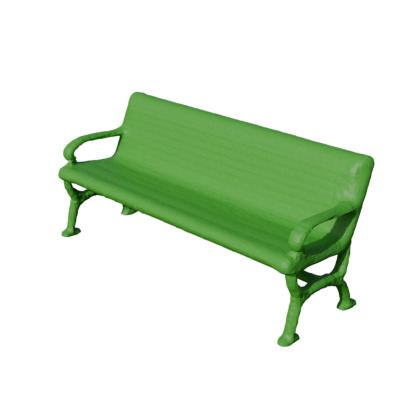} & 
\includegraphics[width=0.08\linewidth,valign=m]{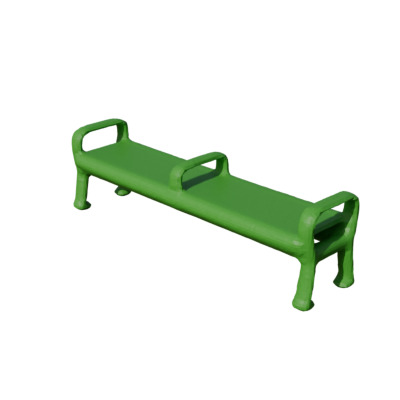} & 
\includegraphics[width=0.08\linewidth,valign=m]{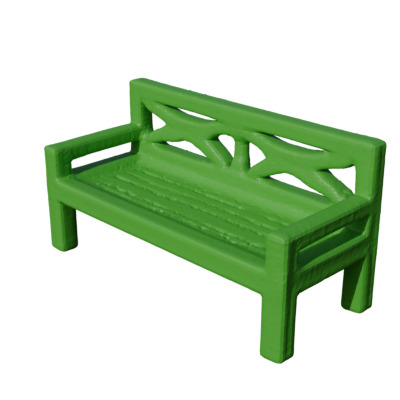} & 
\includegraphics[width=0.08\linewidth,valign=m]{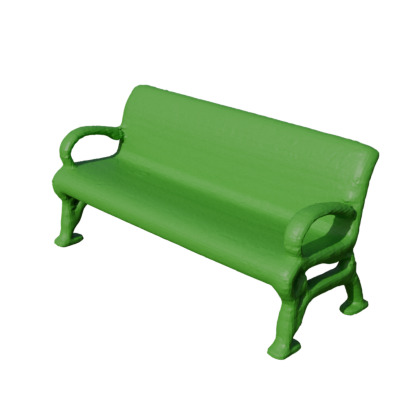} & 
\includegraphics[width=.08\linewidth,valign=m]{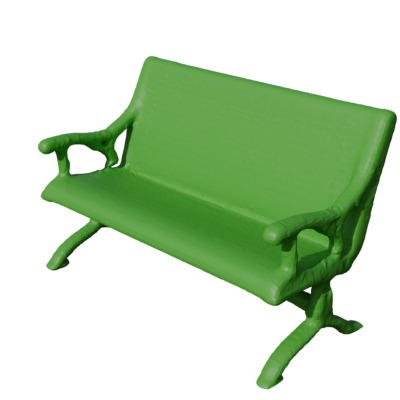} & 
\includegraphics[width=0.08\linewidth,valign=m]{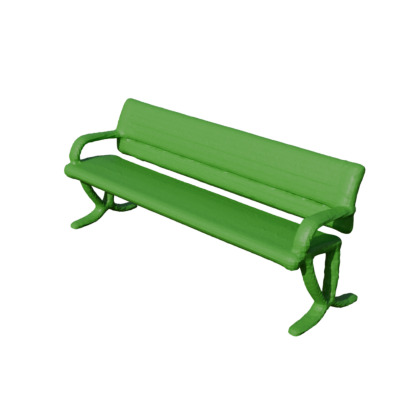} & 
\includegraphics[width=0.08\linewidth,valign=m]{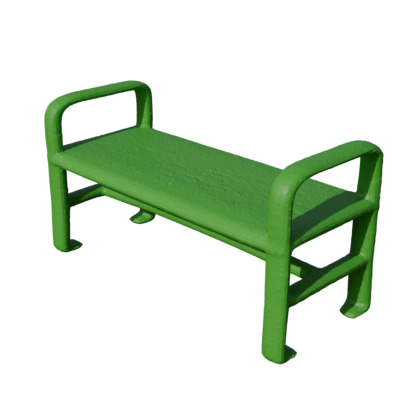} & 
\includegraphics[width=0.08\linewidth,valign=m]{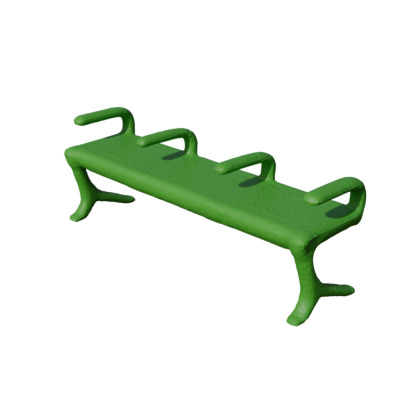} & \\
\hline
\includegraphics[width=.09\linewidth,valign=m]{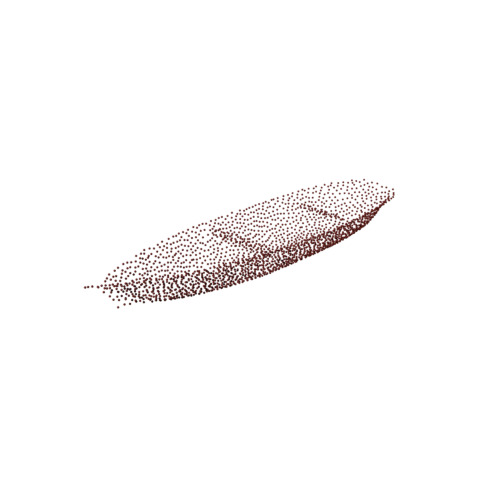} & 
\includegraphics[width=.09\linewidth,valign=m]{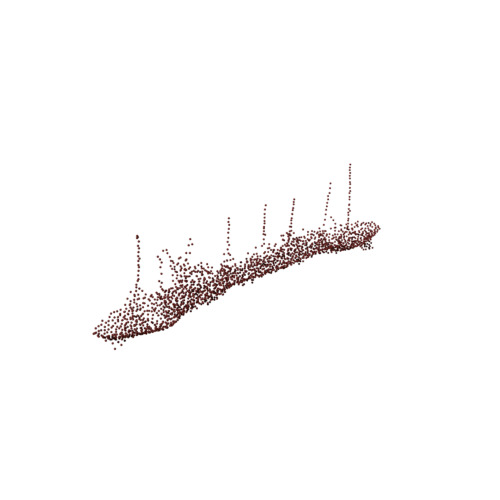} & 
\includegraphics[width=.09\linewidth,valign=m]{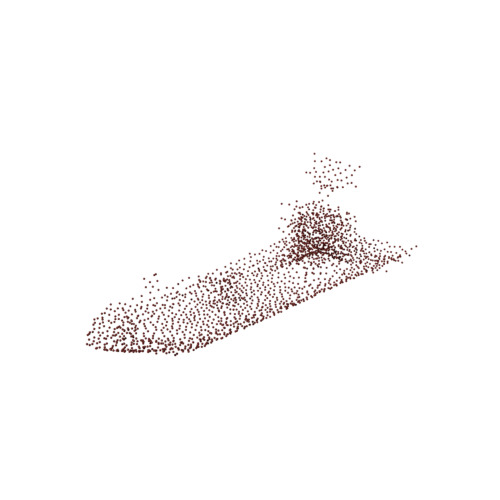} & 
\includegraphics[width=.09\linewidth,valign=m]{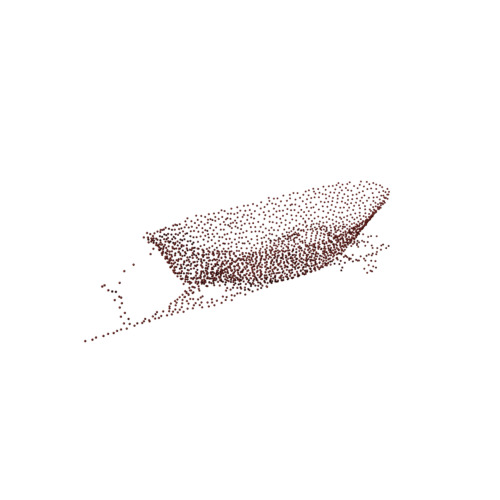} & 
\includegraphics[width=.09\linewidth,valign=m]{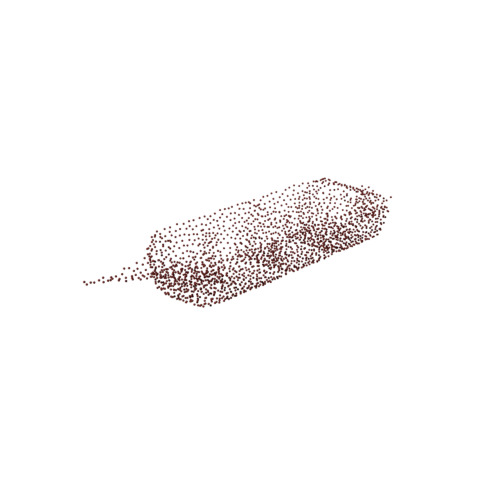} & 
\includegraphics[width=.09\linewidth,valign=m]{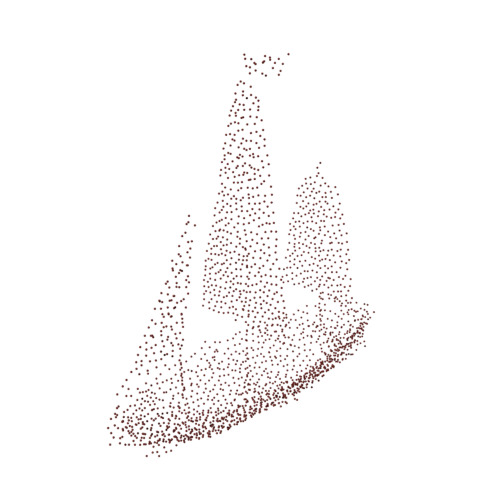} & 
\includegraphics[width=.09\linewidth,valign=m]{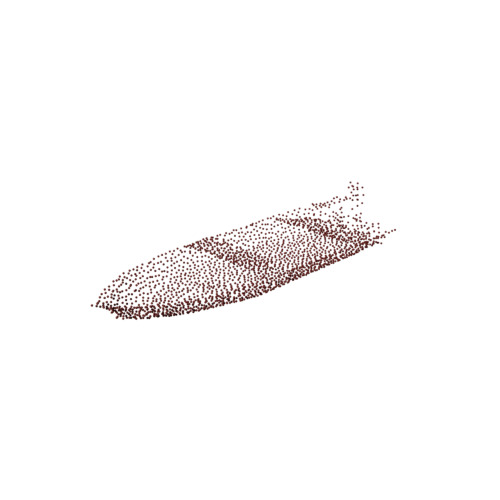} & 
\includegraphics[width=.09\linewidth,valign=m]{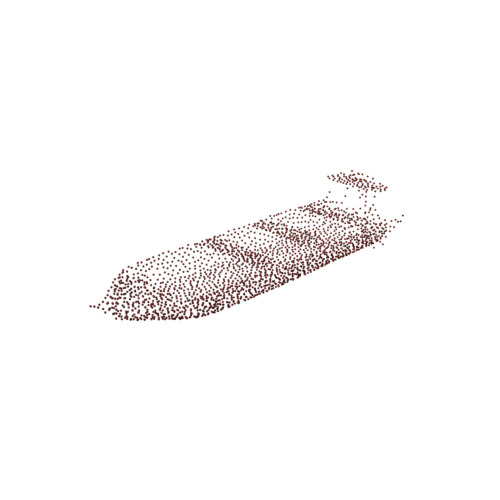} & 
\includegraphics[width=.08\linewidth,valign=m]{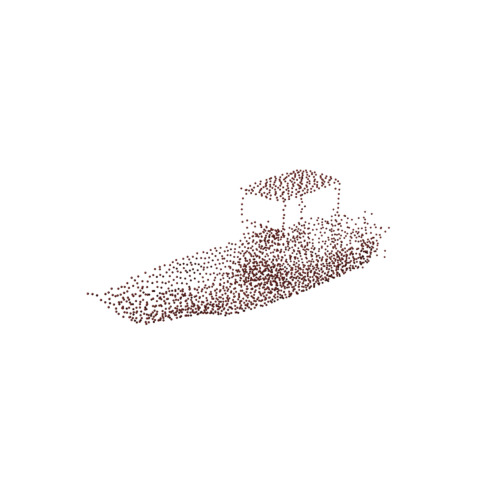} & 
\includegraphics[width=.09\linewidth,valign=m]{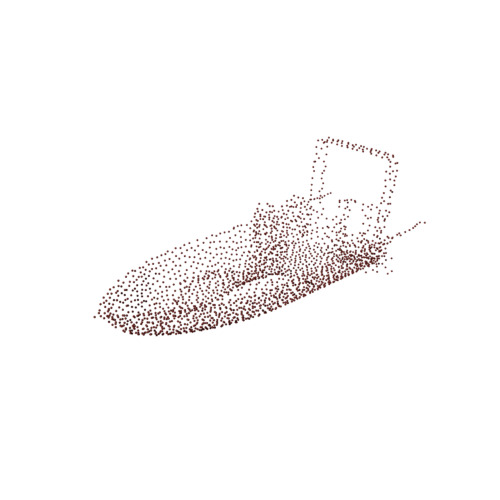} & 
\includegraphics[width=.09\linewidth,valign=m]{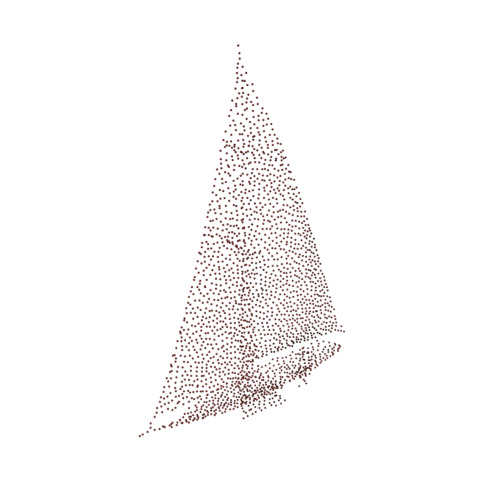} & 
\includegraphics[width=.09\linewidth,valign=m]{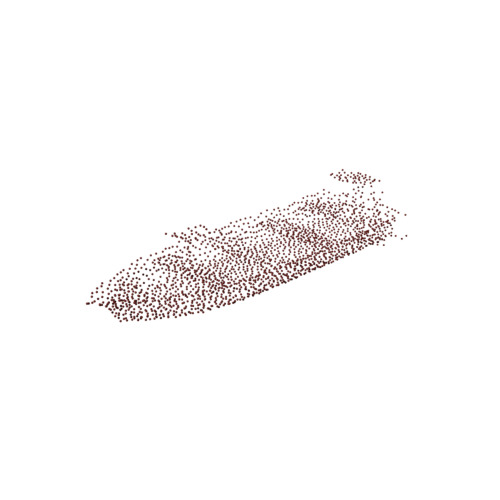} & \\

\includegraphics[width=.09\linewidth,valign=m]{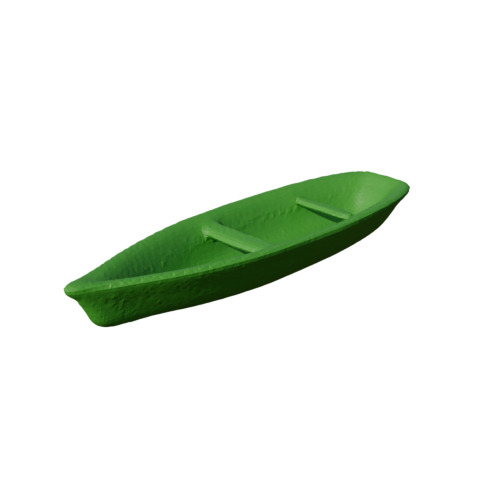} & 
\includegraphics[width=.09\linewidth,valign=m]{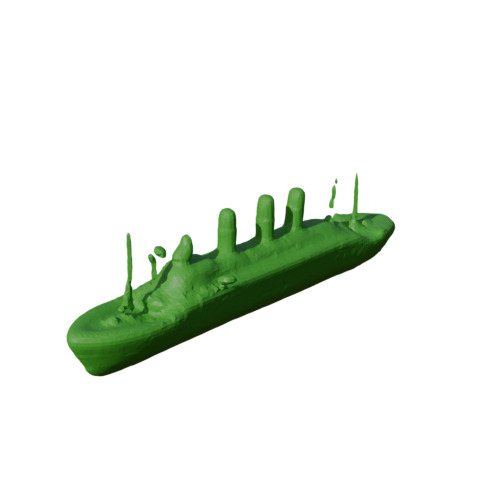} & 
\includegraphics[width=.09\linewidth,valign=m]{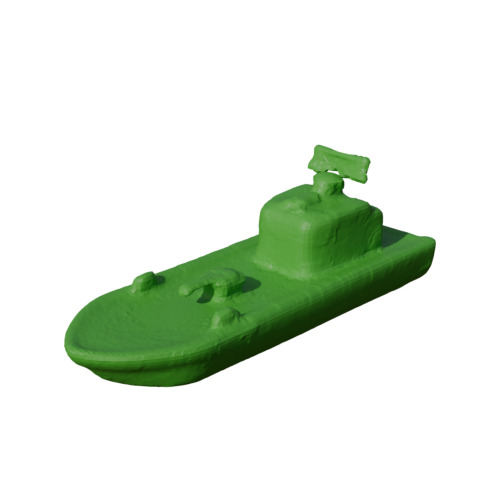} & 
\includegraphics[width=.09\linewidth,valign=m]{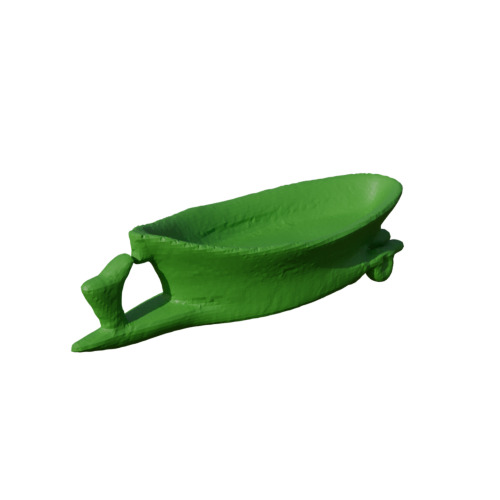} & 
\includegraphics[width=.09\linewidth,valign=m]{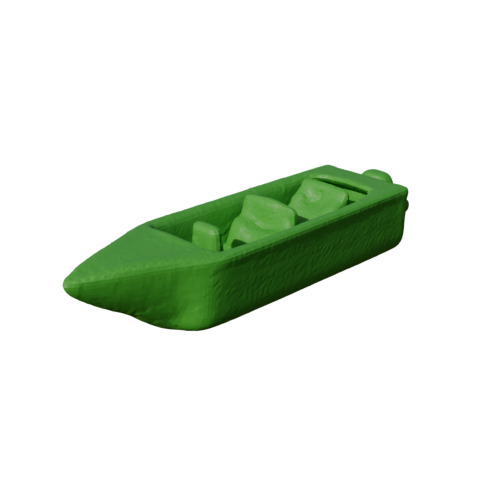} & 
\includegraphics[width=.09\linewidth,valign=m]{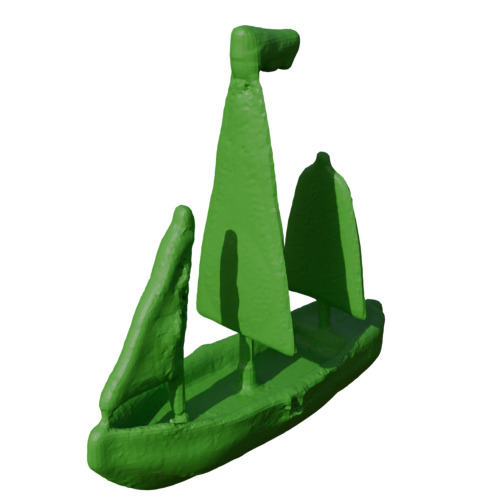} & 
\includegraphics[width=.09\linewidth,valign=m]{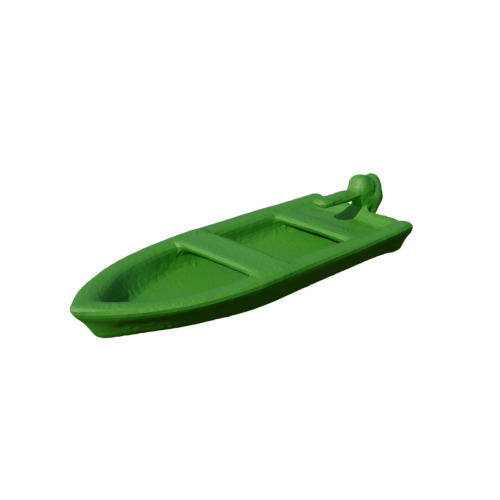} & 
\includegraphics[width=.09\linewidth,valign=m]{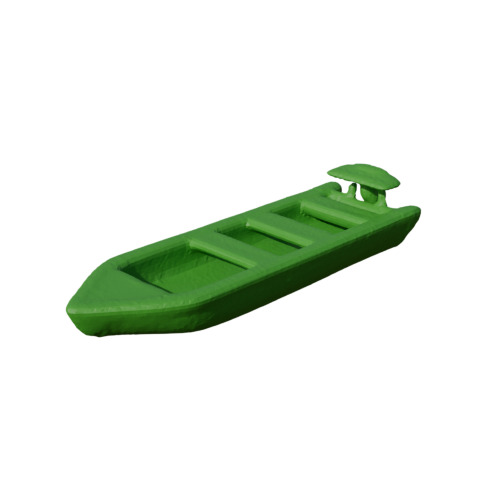} & 
\includegraphics[width=.08\linewidth,valign=m]{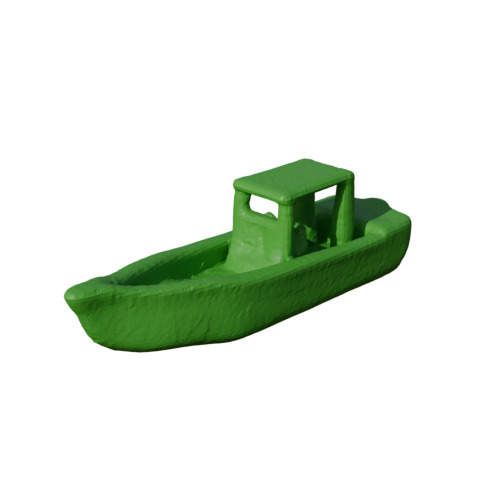} & 
\includegraphics[width=.09\linewidth,valign=m]{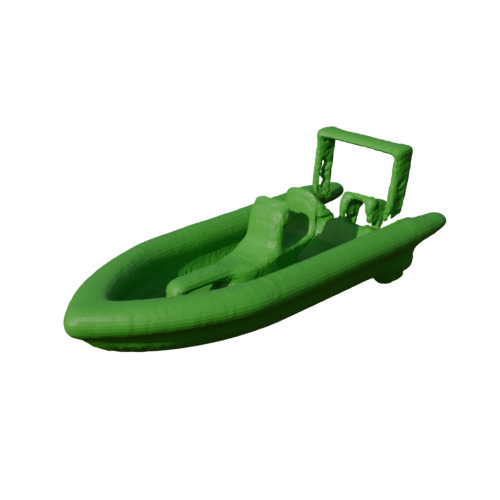} & 
\includegraphics[width=.09\linewidth,valign=m]{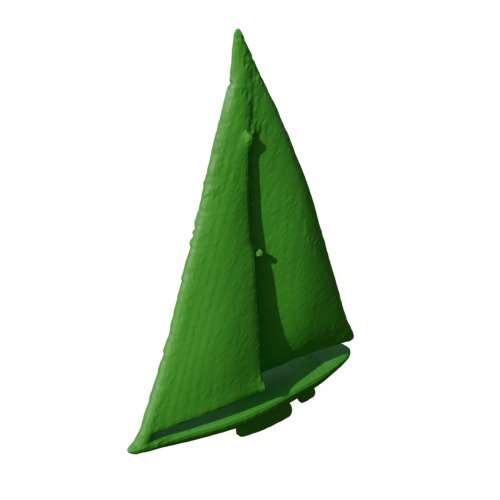} & 
\includegraphics[width=.09\linewidth,valign=m]{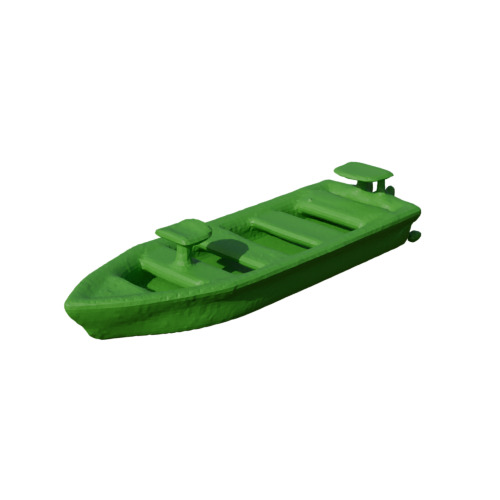} & \\

\end{tabular}
\caption{\textbf{Category-conditioned shape generation on ShapeNet.} We show generated shapes from \method for five categories: chair, lamp, airplane, table, bench and watercraft (from top to bottom). Odd rows are skeletons generated by our model; even rows are surfaces sampled from them.}
\label{fig:category_gen}

\vspace{12pt}

\end{figure*}
\paragraph{Qualitative results}
Figure \ref{fig:category_gen} demonstrates generated samples from our method for various categories from ShapeNet. We observe that our method is able to generate structurally challenging patterns (e.g., grid-like patterns for chair backs), thin parts (e.g, wings for airplanes), tubular or generalized cylinder-like parts (e.g., lamp wires or other connecting pieces). 

\paragraph{Timings} It takes around 40 seconds for GEM3D to generate the skeleton and surface, while 3DS2VS takes about 20 seconds. Reconstruction takes 6s for GEM3D vs 12s for 3DS2VS ($128^3$ grid; all measured on a NVidia RTX 2080Ti). 


\subsection{Surface reconstruction from point clouds}
In this application, the input to our method is a sparse point cloud of a shape and the output is a reconstructed surface. Here, we also follow the ``auto-encoding'' setting also used in 3DILG~\cite{Zhang_3dilg} and 3DS2VS \cite{3DShape2VecSet}, where the point cloud is obtained after sampling the original input surface. In this autoencoding setting, we only use the second diffusion stage of our method i.e., given a point-based skeleton, the second stage generates the medial latents, then our surface decoder uses them to reconstruct the surface. To apply our method on this task, we need as input a point-sampled skeleton. To obtain such a skeleton, we use a variant of P2PNet \cite{yin2018p2pnet}, an encoder-decoder network that takes as input a point cloud of a shape's surface and converts it to a point-based skeleton representation. Compared to the original P2PNet, we replaced the original PointNet++ encoder  \cite{qi2017pointnet++} with the cross-attention-based shape encoder from 3DS2VS \cite{3DShape2VecSet}, including  its positional-based embedding functions. We train this P2PNet variant in the training split of ShapeNet. Then at test time, we use its output skeletons as input to our model decoder. We emphasize that no manual input or supervision were needed to obtain the skeletons -- P2PNet is trained in a unsupervised manner. 

\paragraph{Baselines and metrics}
We again compare our method with \emph{3DILG}  \cite{Zhang_3dilg} and 3DS2VS \cite{3DShape2VecSet}, which were also applied in the same auto-encoding setting. All methods use the same number of latent codes ($2048$).

For evaluation purposes, we compare reconstructed shapes to their corresponding reference (``ground-truth'') shapes, which the input point clouds were sampled from. Following 3DS2VS \cite{3DShape2VecSet}, we use the Chamfer Distance (\textbf{CD}), Intersection Over Union (\textbf{IoU}) and F-Score (\textbf{F1}). Compared to 3DS2VS'  protocol, we made the following changes: (i) for computing IoU, we use a higher-resolution grid of $256^3$ instead of $128^3$ to better characterize reconstruction of small-scale surface details, thin parts, and holes, (ii) input point clouds are sampled with farthest point sampling instead to better cover small and thin parts. In the supplement, we provide comparisons using the original 3DS2VS' protocol. 

\begin{figure}[!b]

\setlength\tabcolsep{-3pt}
\begin{tabular}{ccccccccc}
Input & 3DILG & 3D2VS & Ours(S) & Ours(M) & GT \\
\addlinespace[5pt]
\includegraphics[width=.18\linewidth,valign=m]{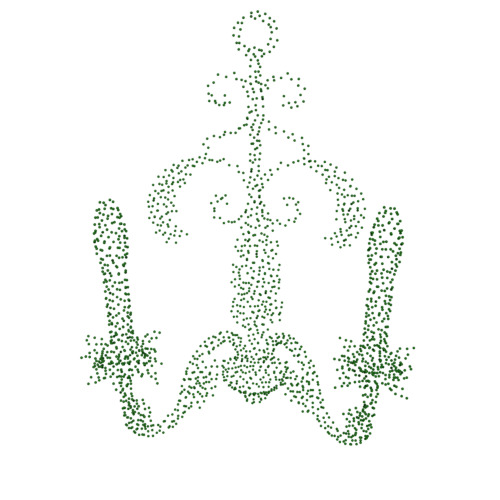} & 
\includegraphics[width=.18\linewidth,valign=m]{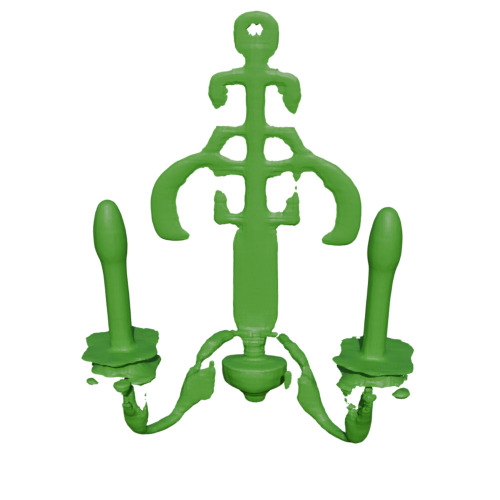} & 
\includegraphics[width=.18\linewidth,valign=m]{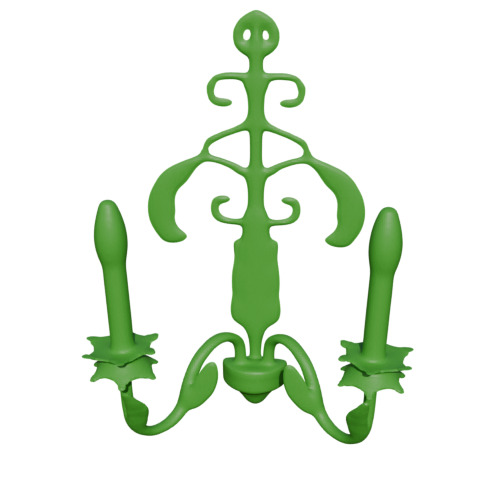} & 
\includegraphics[width=.18\linewidth,valign=m]{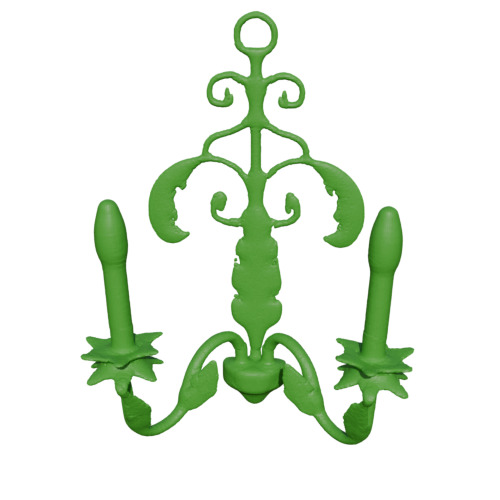} & 
\includegraphics[width=.18\linewidth,valign=m]{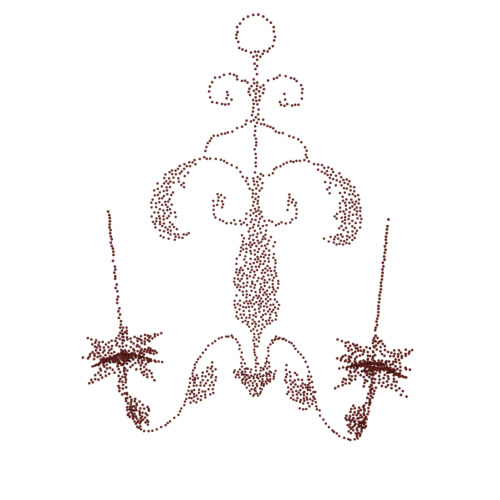} &
\includegraphics[width=.18\linewidth,valign=m]{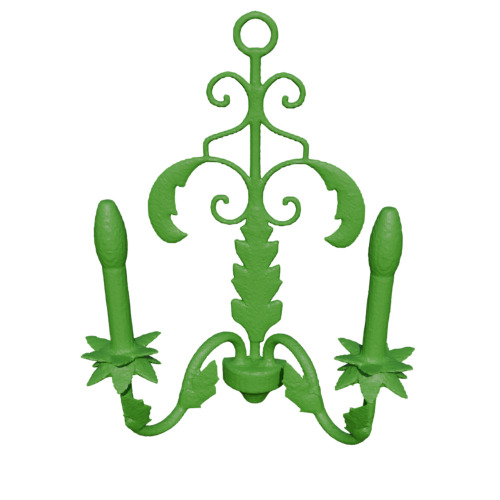} &
\\
\addlinespace[-4pt]
\includegraphics[width=.18\linewidth,valign=m]{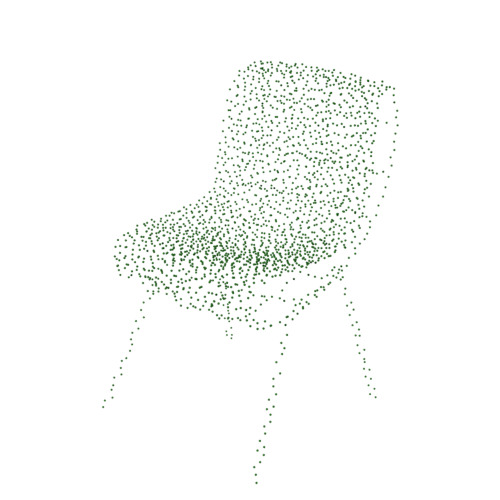} & 
\includegraphics[width=.18\linewidth,valign=m]{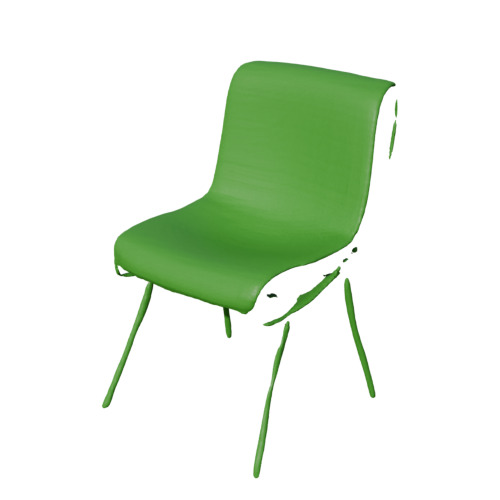} & 
\includegraphics[width=.18\linewidth,valign=m]{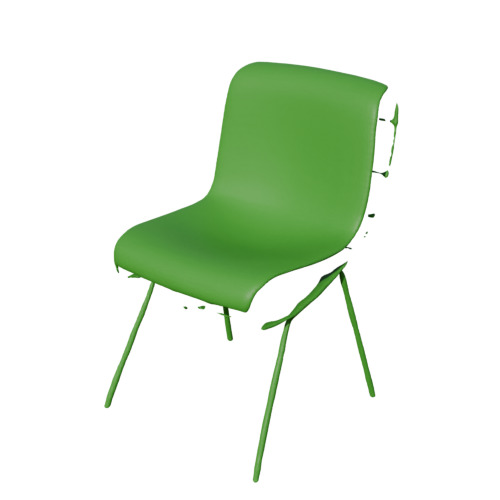} & 
\includegraphics[width=.18\linewidth,valign=m]{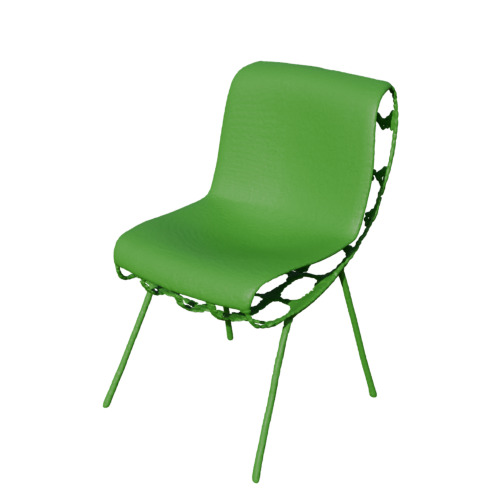} & 
\includegraphics[width=.18\linewidth,valign=m]{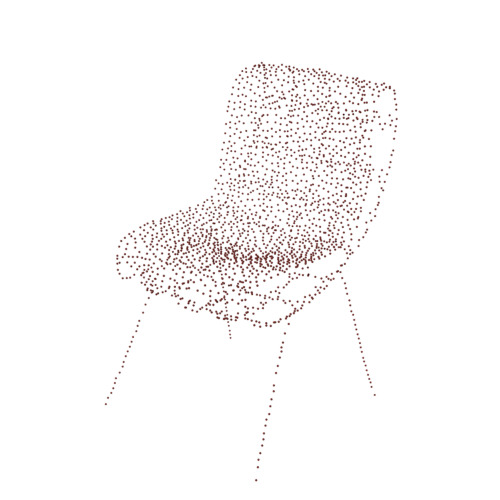} &
\includegraphics[width=.18\linewidth,valign=m]{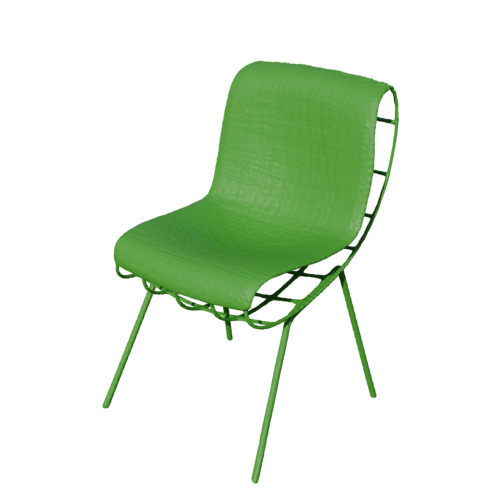} &
\\
\addlinespace[-5pt]
\includegraphics[width=.18\linewidth,valign=m]{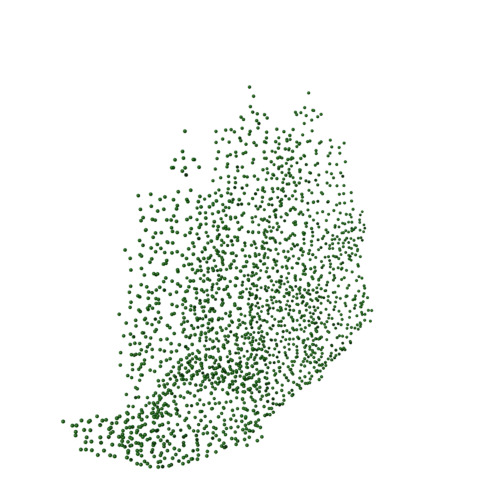} & 
\includegraphics[width=.18\linewidth,valign=m]{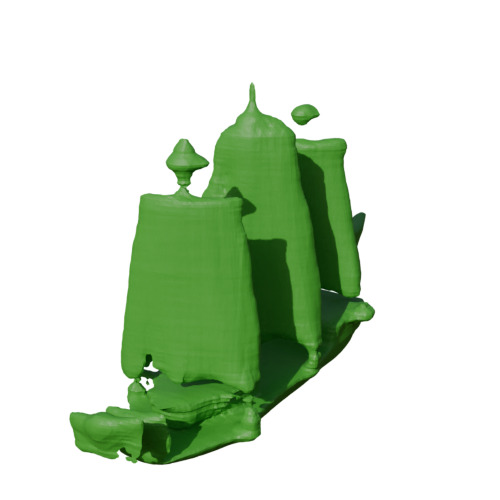} & 
\includegraphics[width=.18\linewidth,valign=m]{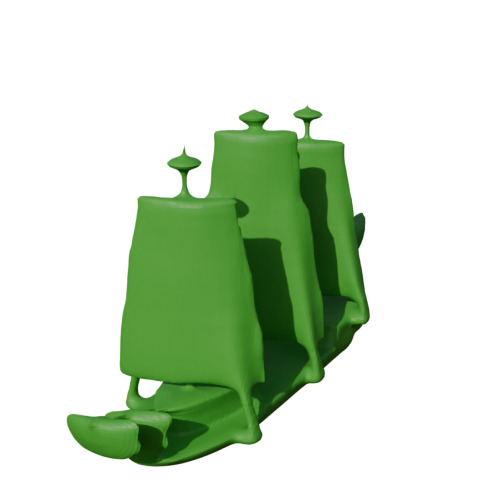} & 
\includegraphics[width=.18\linewidth,valign=m]{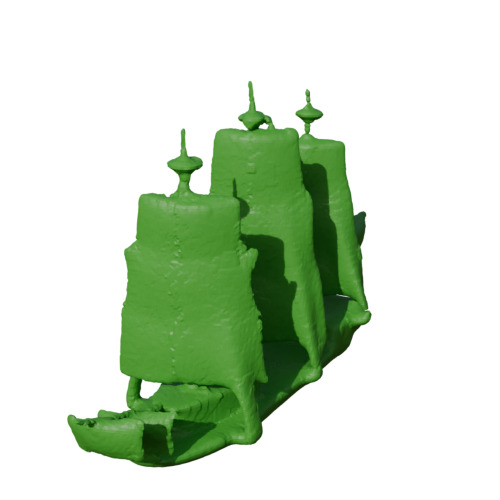} & 
\includegraphics[width=.18\linewidth,valign=m]{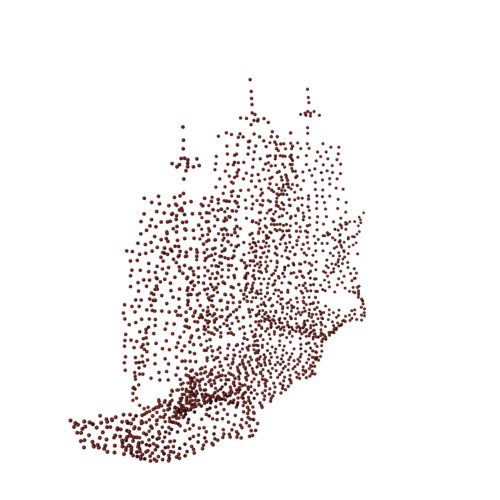} &
\includegraphics[width=.18\linewidth,valign=m]{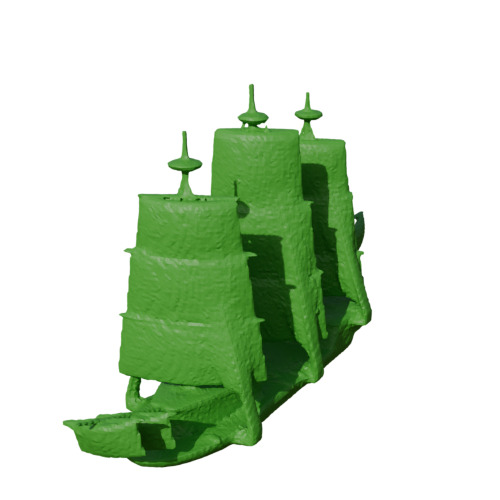} &
\\
\includegraphics[width=.18\linewidth,valign=m]{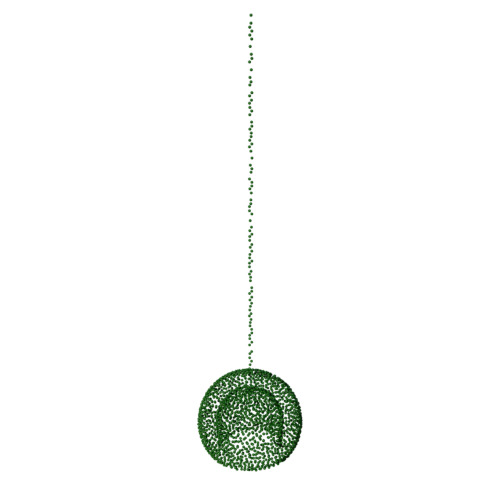} & 
\includegraphics[width=.18\linewidth,valign=m]{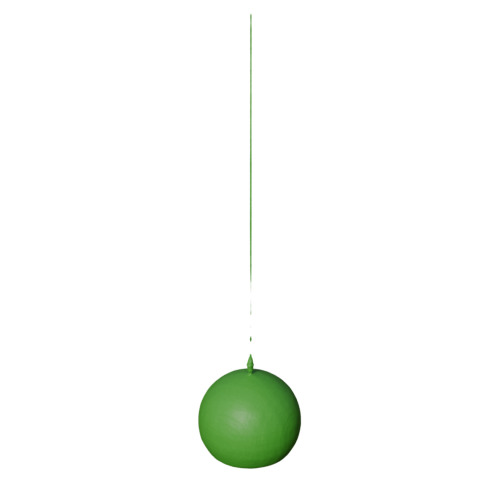} & 
\includegraphics[width=.18\linewidth,valign=m]{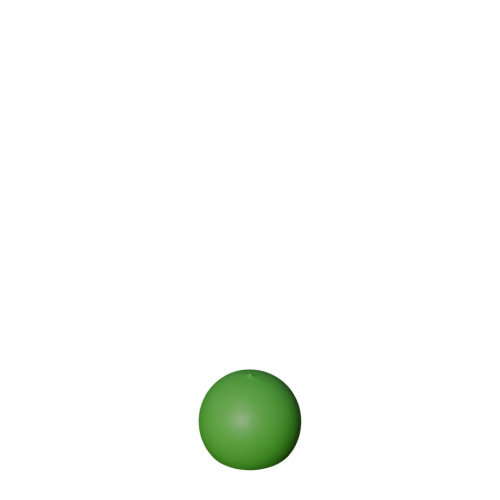} & 
\includegraphics[width=.18\linewidth,valign=m]{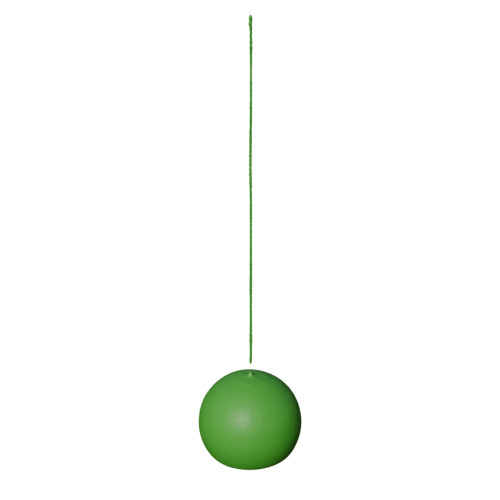} & 
\includegraphics[width=.18\linewidth,valign=m]{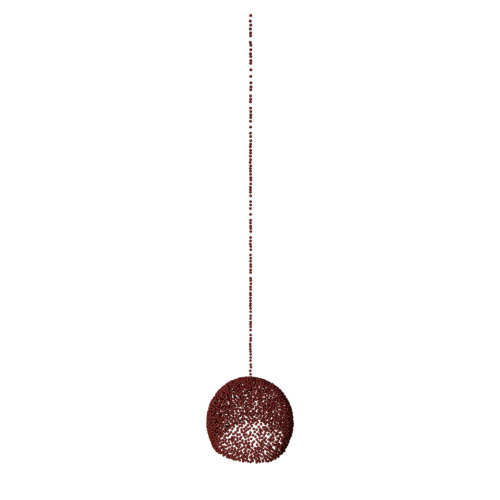} &
\includegraphics[width=.18\linewidth,valign=m]{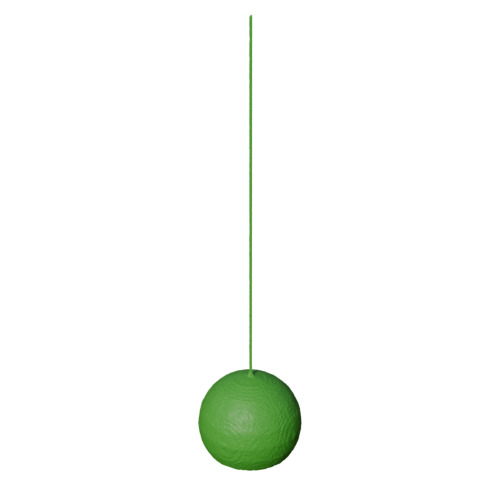} &
\\
\addlinespace[-1.0pt]
\includegraphics[width=.18\linewidth,valign=m]{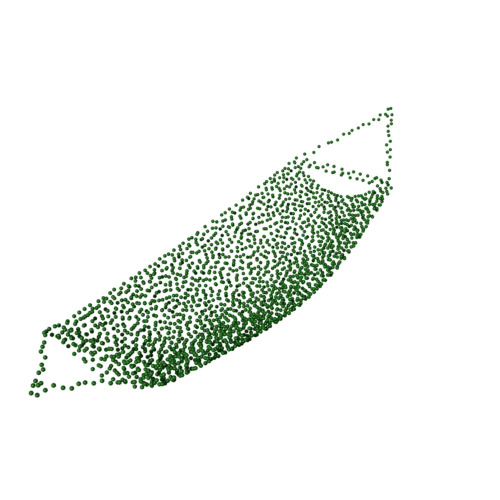} & 
\includegraphics[width=.18\linewidth,valign=m]{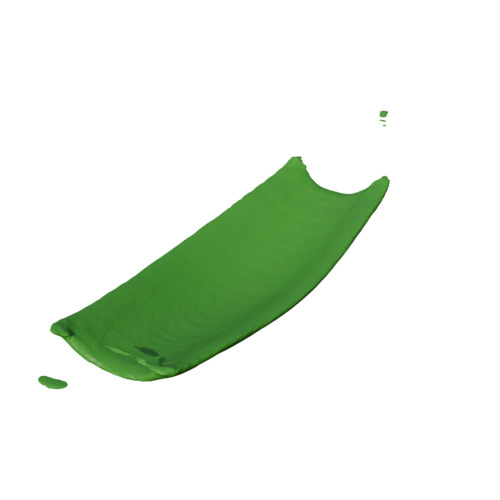} & 
\includegraphics[width=.18\linewidth,valign=m]{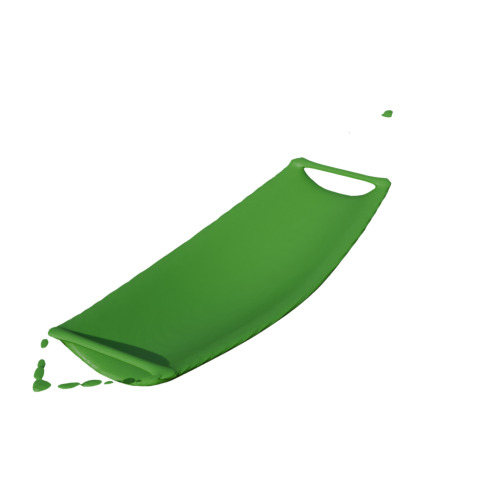} & 
\includegraphics[width=.18\linewidth,valign=m]{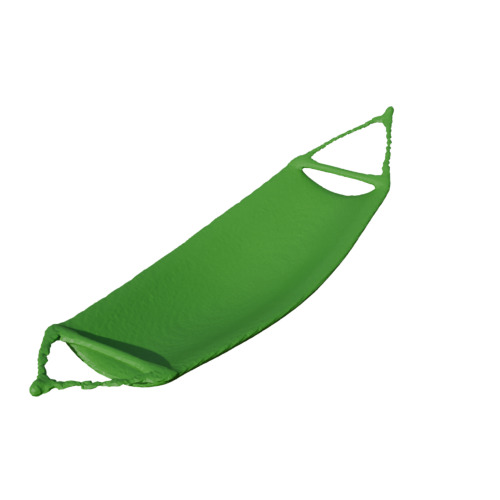} & 
\includegraphics[width=.18\linewidth,valign=m]{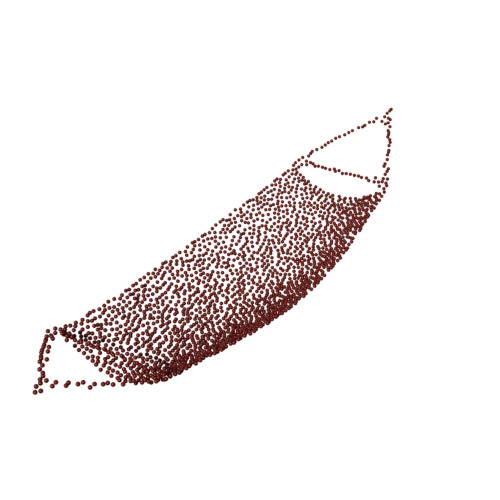} &
\includegraphics[width=.18\linewidth,valign=m]{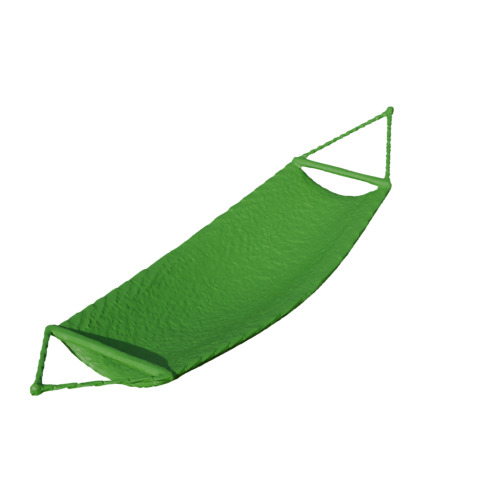} &
\\
\includegraphics[width=.16\linewidth,valign=m]{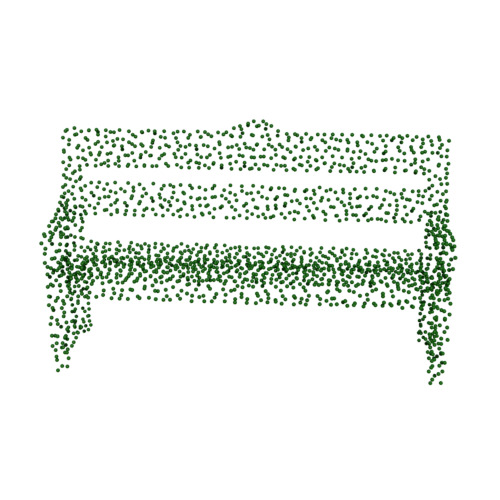} & 
\includegraphics[width=.16\linewidth,valign=m]{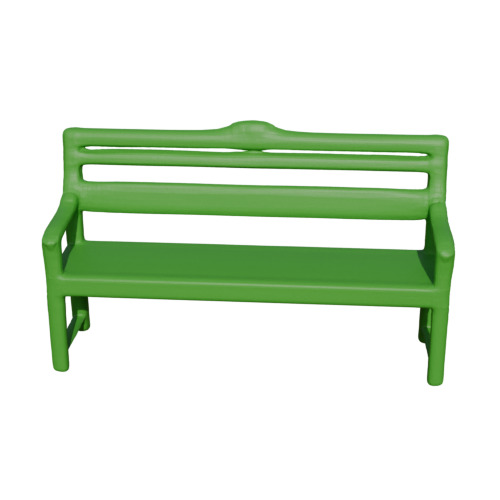} & 
\includegraphics[width=.16\linewidth,valign=m]{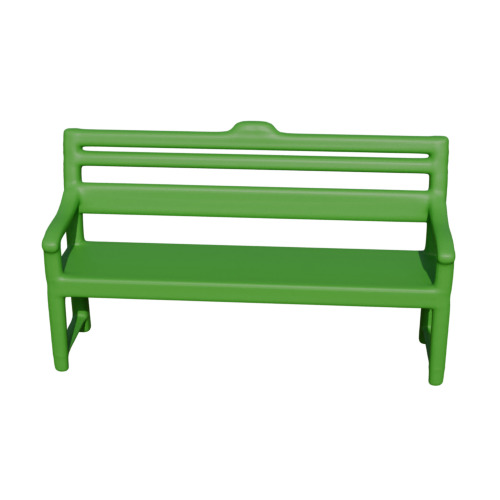} & 
\includegraphics[width=.16\linewidth,valign=m]{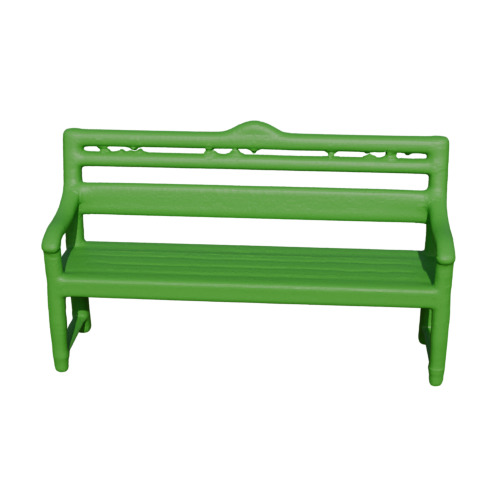} & 
\includegraphics[width=.16\linewidth,valign=m]{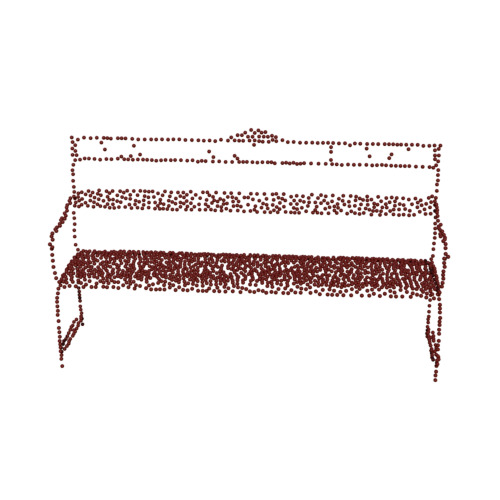} &
\includegraphics[width=.16\linewidth,valign=m]{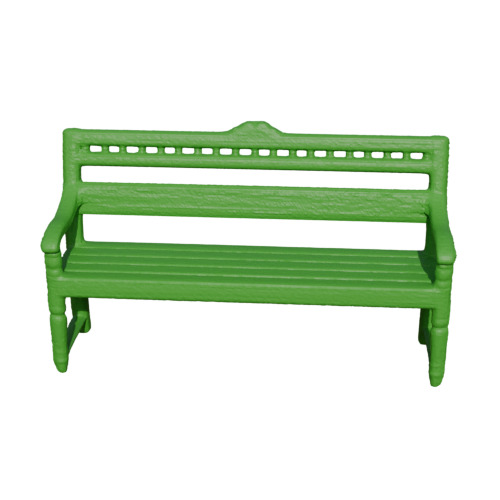} &
\\
\end{tabular}

\vspace{-15pt}
\caption{\textbf{Point cloud reconstruction on ShapeNet.} \method yields better reconstruction esp for thin and tubular parts, and with better connectivity.} 
\label{fig:comparison_pc_ae_shapenet}
\vspace{-2mm}
\end{figure}
\begin{figure}[!ht]

\setlength\tabcolsep{-1pt}
\begin{tabular}{ccccccccc}
Input & 3DILG & 3D2VS & Ours(S) & Ours(M) & GT \\
\addlinespace[-2pt]
\includegraphics[width=.15\linewidth,valign=m]{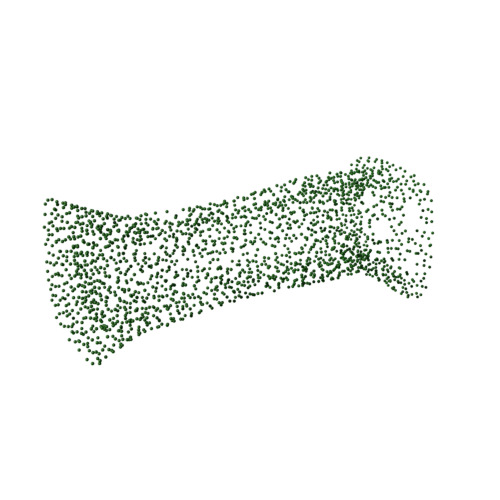} & 
\includegraphics[width=.15\linewidth,valign=m]{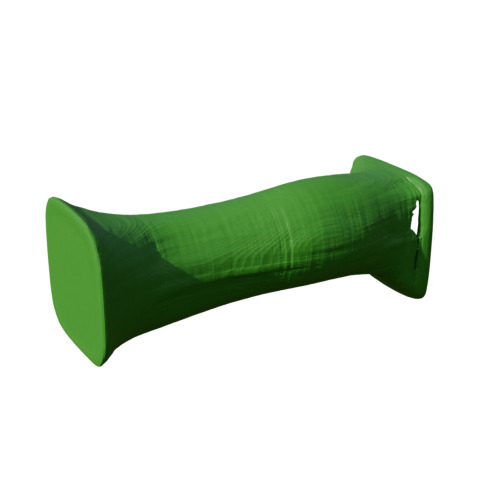} & 
\includegraphics[width=.15\linewidth,valign=m]{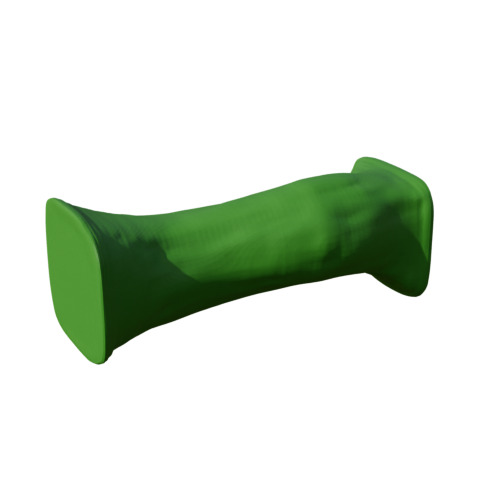} & 
\includegraphics[width=.15\linewidth,valign=m]{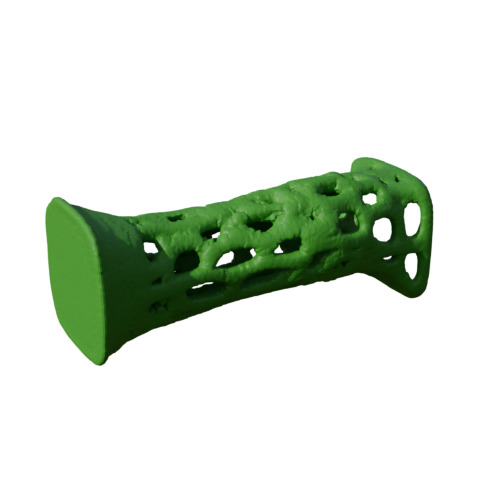} & 
\includegraphics[width=.15\linewidth,valign=m] {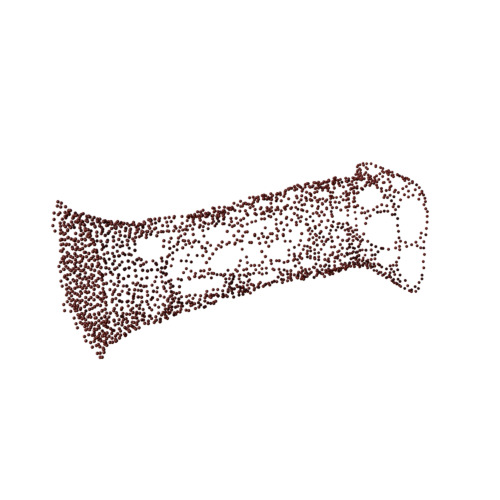} &
\includegraphics[width=.15\linewidth,valign=m]{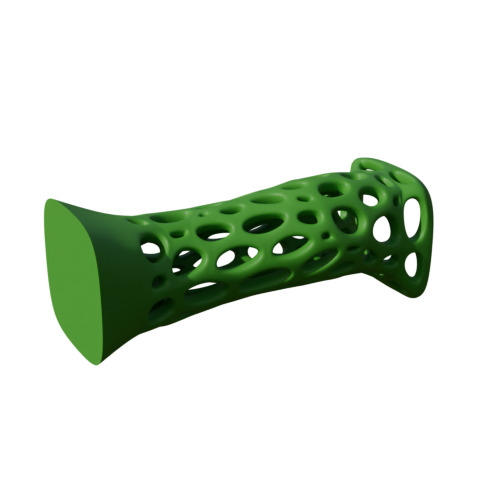} &
\\
\addlinespace[-2pt]
\includegraphics[width=.15\linewidth,valign=m]{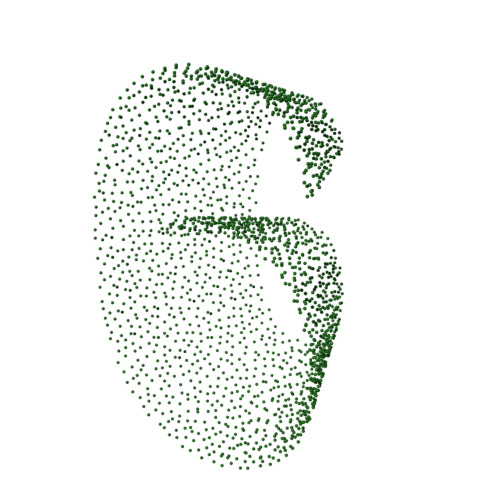} & 
\includegraphics[width=.15\linewidth,valign=m]{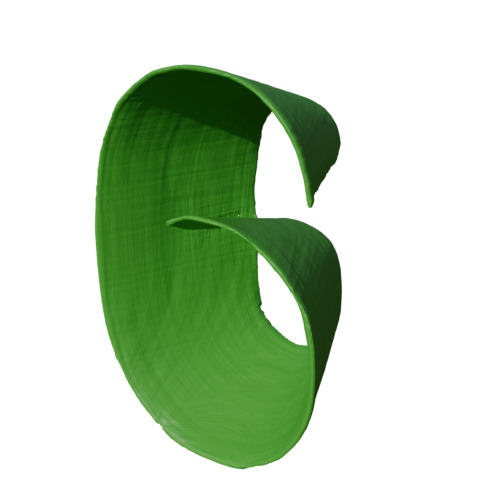} & 
\includegraphics[width=.15\linewidth,valign=m]{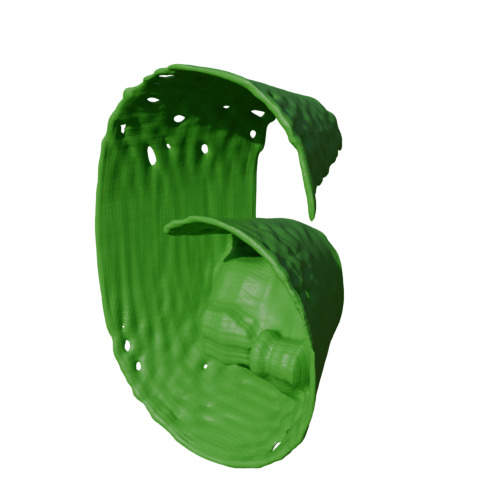} & 
\includegraphics[width=.15\linewidth,valign=m]{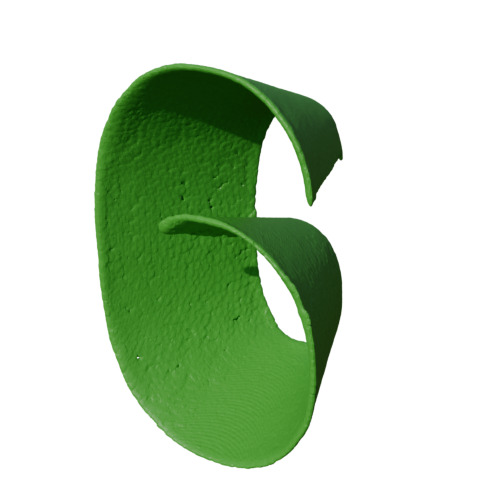} & 
\includegraphics[width=.15\linewidth,valign=m] {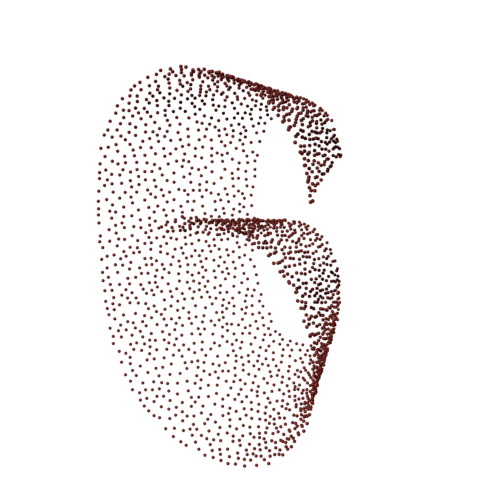} &
\includegraphics[width=.15\linewidth,valign=m]{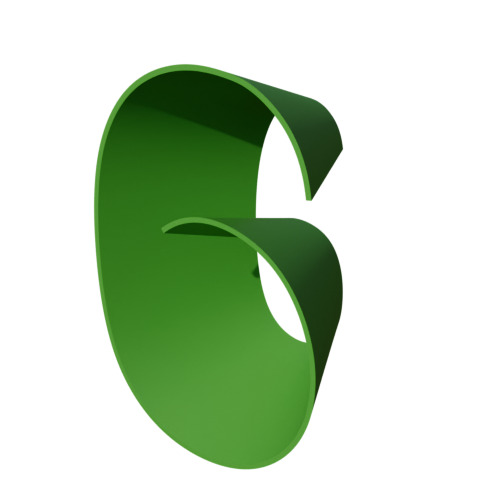} &
\\
\includegraphics[width=.15\linewidth,valign=m]{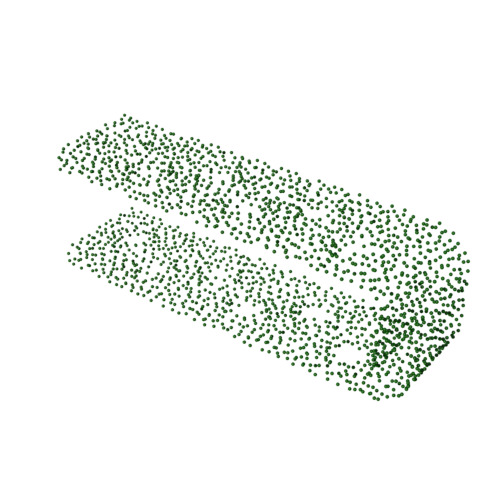} & 
\includegraphics[width=.15\linewidth,valign=m]{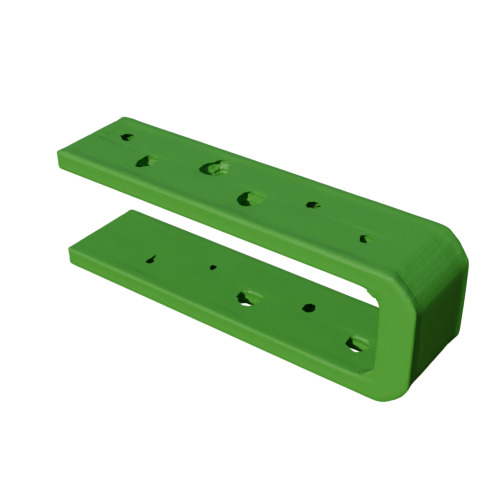} & 
\includegraphics[width=.15\linewidth,valign=m]{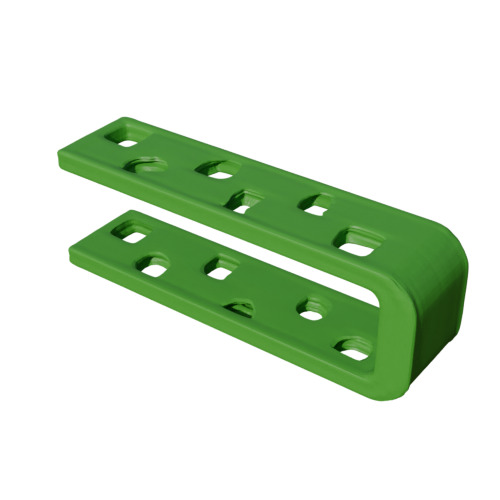} & 
\includegraphics[width=.15\linewidth,valign=m]{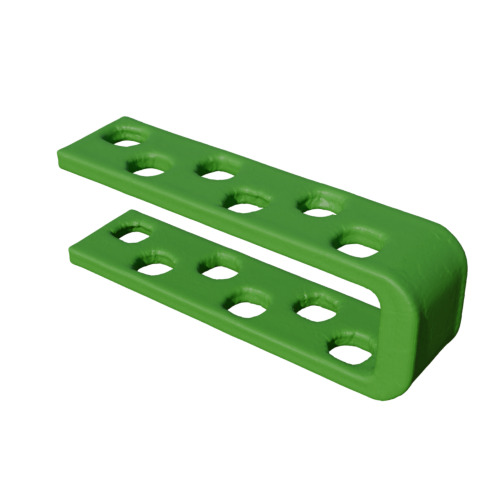} & 
\includegraphics[width=.15\linewidth,valign=m] {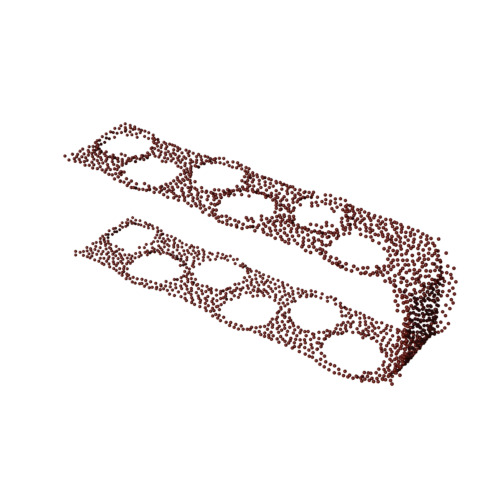} &
\includegraphics[width=.15\linewidth,valign=m]{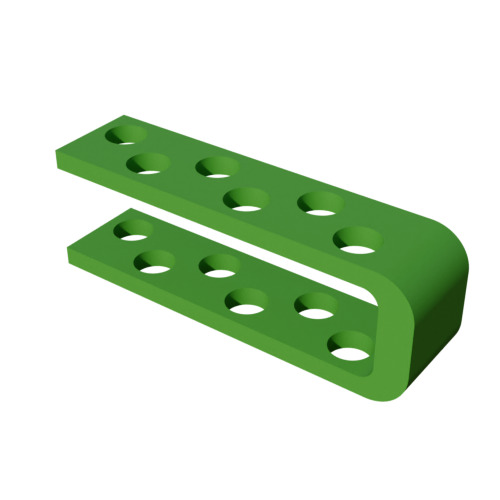} &
\\
\addlinespace[-4pt]
\includegraphics[width=.15\linewidth,valign=m]{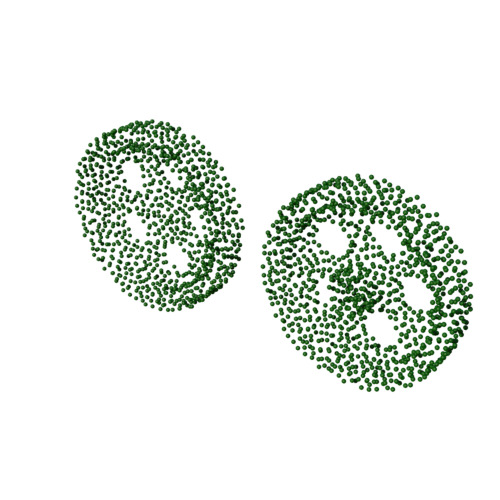} & 
\includegraphics[width=.15\linewidth,valign=m]{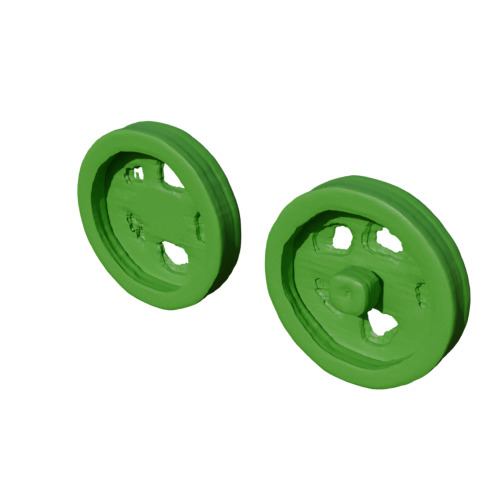} & 
\includegraphics[width=.15\linewidth,valign=m]{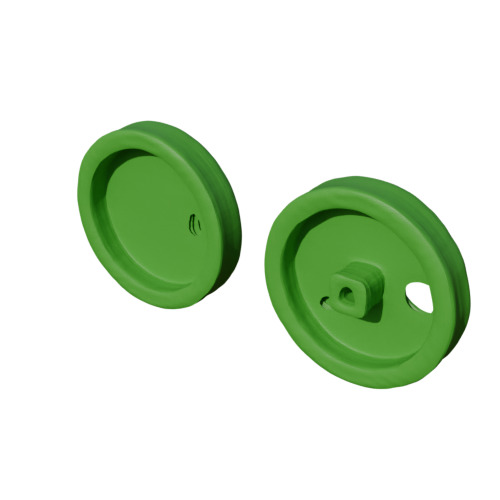} & 
\includegraphics[width=.15\linewidth,valign=m]{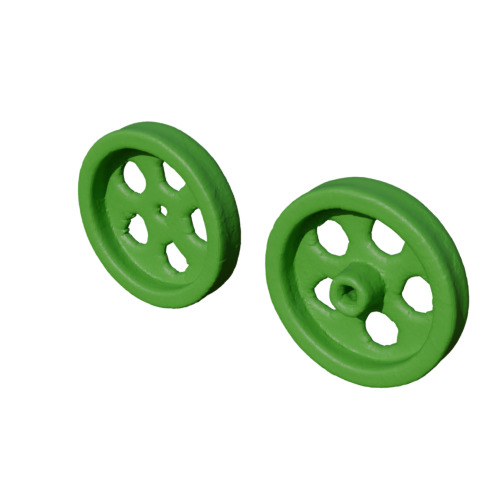} & 
\includegraphics[width=.15\linewidth,valign=m] {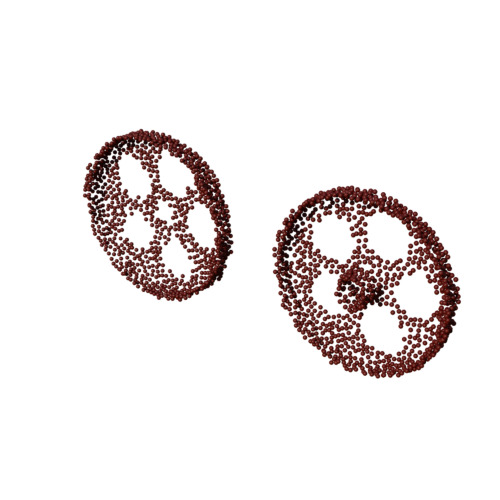} &
\includegraphics[width=.15\linewidth,valign=m]{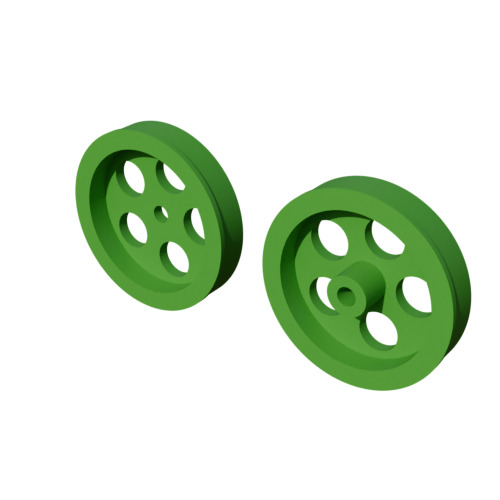} &
\\
\addlinespace[-2pt]
\includegraphics[width=.15\linewidth,valign=m]{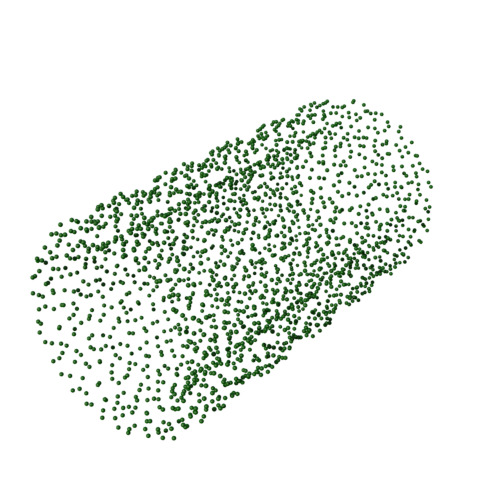} & 
\includegraphics[width=.15\linewidth,valign=m]{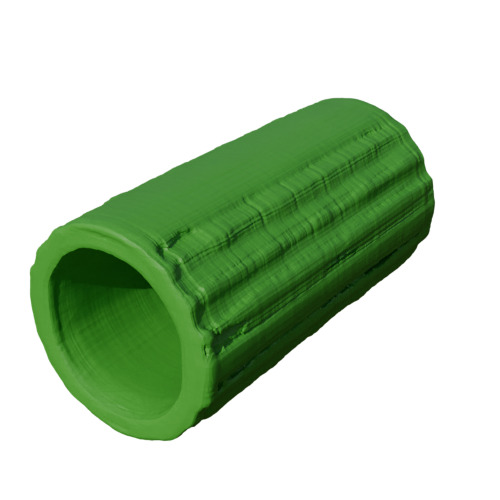} & 
\includegraphics[width=.15\linewidth,valign=m]{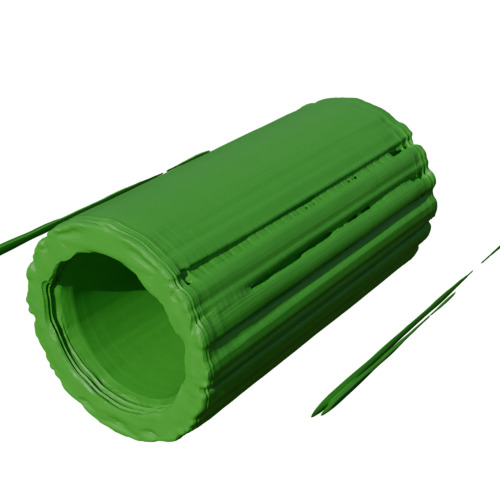} & 
\includegraphics[width=.15\linewidth,valign=m]{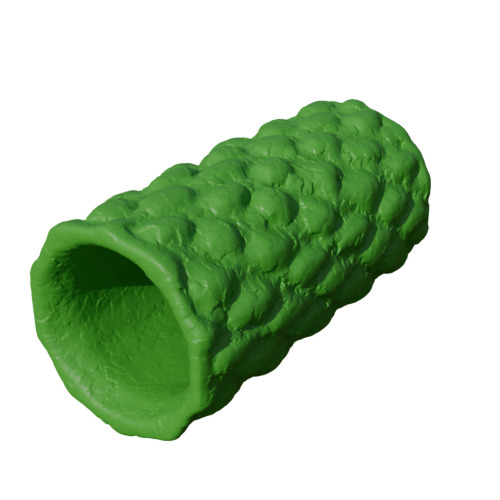} & 
\includegraphics[width=.15\linewidth,valign=m] {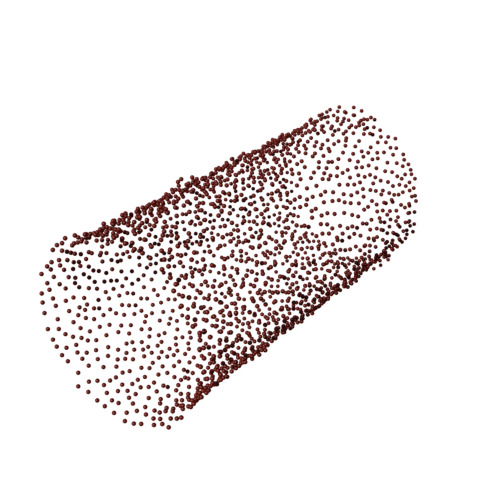} &
\includegraphics[width=.15\linewidth,valign=m]{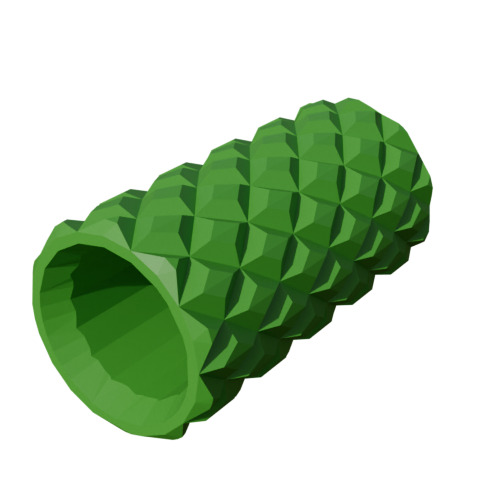} &
\\
\addlinespace[-4pt]
\includegraphics[width=.15\linewidth,valign=m]{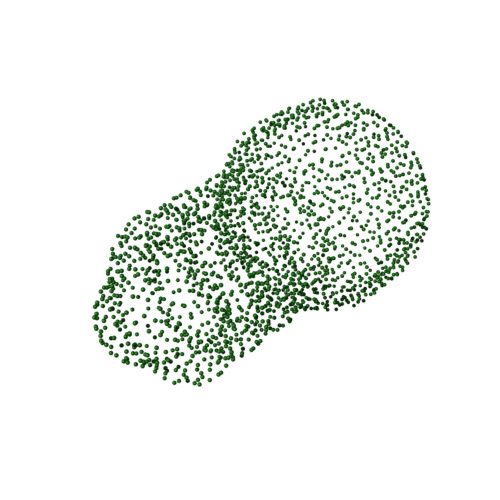} & 
\includegraphics[width=.15\linewidth,valign=m]{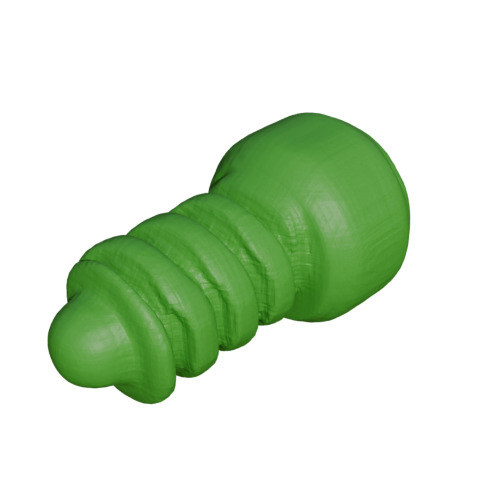} & 
\includegraphics[width=.15\linewidth,valign=m]{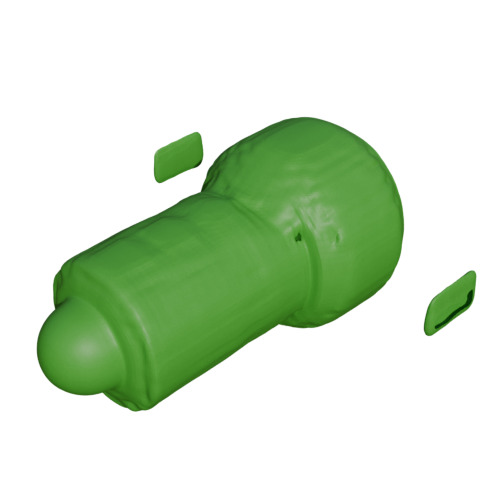} & 
\includegraphics[width=.15\linewidth,valign=m]{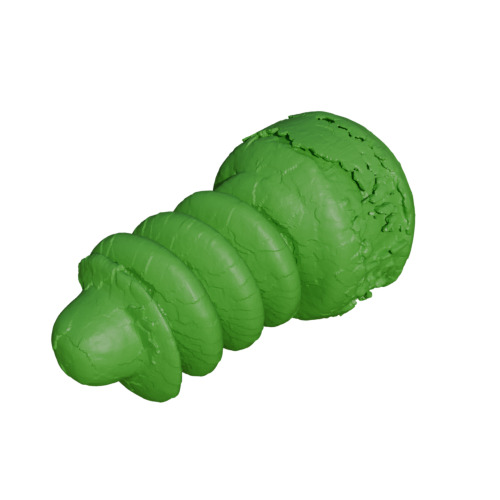} & 
\includegraphics[width=.15\linewidth,valign=m] {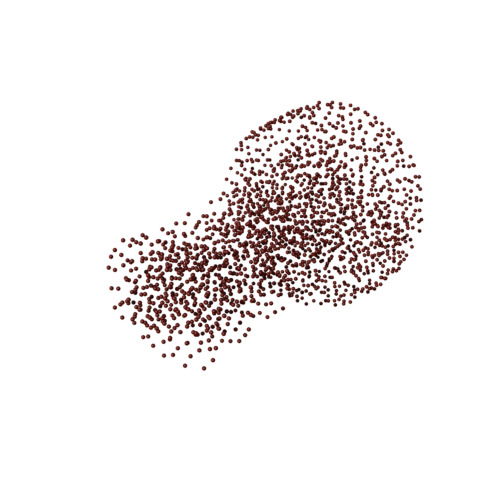} &
\includegraphics[width=.15\linewidth,valign=m]{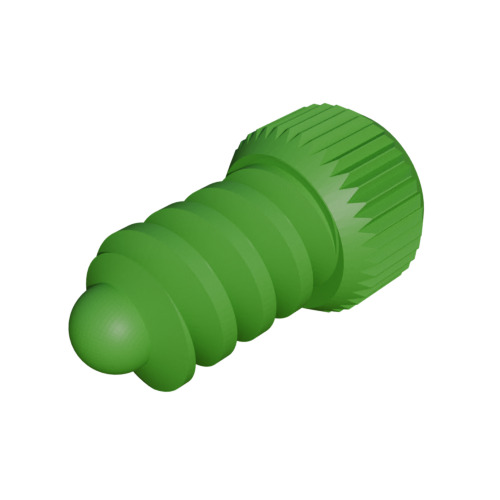} &
\\



\end{tabular}

\vspace{-12pt}
\caption{\textbf{Point cloud reconstruction on Thingi10K.} In this out-of-distribution test setting our model is able to reconstruct topology (e.g., holes, connectivity) and surface details with less artifacts than prior work.}
\label{fig:comparison_pc_thingi}
\end{figure}

\begin{table}[t!]
  \centering
\begin{tabular}{rc|c|c|c}
\toprule
Metric  & Test dataset &  3DILG &   3DS2VS & Ours    \\ 
\midrule
CD  ($\downarrow$) & Shapenet (10 cat.)  & 1.69 & 1.32 & \textbf{1.17} \\
CD  ($\downarrow$) & Shapenet (all cat.)  & 2.05 & 1.70 & \textbf{1.38} \\ 
\hline
IOU  ($\uparrow$) & Shapenet (10 cat.)   & 90.1 & \textbf{96.2} & 95.9\\
IOU  ($\uparrow$) & Shapenet (all cat.)   & 90.5 & 95.4 & \textbf{95.5} \\
\hline
F1  ($\uparrow$) & Shapenet (10 cat.)  & 97.7 & 98.5 & \textbf{99.3} \\
F1  ($\uparrow$) & Shapenet (all cat.)  & 96.3 & 97.4 & \textbf{98.7} \\  

\bottomrule
\end{tabular}
  \caption{Point cloud surface reconstruction (``auto-encoding'') evaluation on ShapeNet. The rows ``ShapeNet (10 cat.)'' report the measures averaged over the $10$ largest categories of ShapeNet, while the rows ``ShapeNet (all)'' report the measures averaged over all $55$ ShapeNet categories.
 For evaluation per-category, please see the supplement.   \label{tab:reconstruction_shapenet_eval} }
\vspace{-5mm}
\end{table}


\begin{table}[t!]
  \centering
\begin{tabular}{cccccccc}
\toprule
\midrule
 &     Thingi10K    &    $\# shapes$  & 3DILG          & 3DS2VS            &  Ours         \\ \hline
& all & 9997     & 2.12 &  1.74   &     \textbf{1.68}                    \\ 
& $genus\ge0$ & 5608       &  1.93   & 1.62  &     \textbf{1.61}                      \\ 
& $genus\ge5$ & 1037 &  2.41      & 1.96 &     \textbf{1.78} \\
& $genus\ge10$& 489 & 2.88    & 2.30 &   \textbf{1.98}  \\ 
 \multirow{-4}{*}{ CD $\downarrow$} &  $genus\ge20$  &229  & 3.67 &  2.91 & \textbf{2.32}                  \\ 
 \hline

  \hline&  all & 9997    & 93.6 &  96.4  &     \textbf{96.7}     \\
 & $genus\ge0$ & 5608     & 95.3 &  \textbf{97.4}    &     \textbf{97.4}  \\
 & $genus\ge5$ & 1037      & 91.1 &  94.8&     \textbf{95.9}  \\
 & $genus\ge10$ & 489      & 87.5   &  92.3  & \textbf{93.7}         \\
 \multirow{-4}{*}{ F1 $\uparrow$} & $genus\ge20$ & 229  & 82.6  & 88.0  &     \textbf{90.4}   \\
\bottomrule
\end{tabular}
\caption{Evaluation measures on point cloud surface reconstruction in the Thingi10K dataset. We note that none of the methods are trained on Thingi10K. We report performance in terms of CD and F1 scores for shapes of increasing genus. \label{tab:thingi_quant_eval} }
\vspace{-5mm}
\end{table}


\paragraph{Experimental setup}
Following \cite{Zhang_3dilg} and \cite{3DShape2VecSet}, we use the same training and test split from ShapeNet-v2  to train and test our method respectively. We also test all competing methods in a more challenging out-of-distribution testing scenario: after training on ShapeNet, we test all methods on Thingi10K \cite{Thingi10K}. We note that the whole dataset is $10K$ test shapes, $4$ times larger than the ShapeNet's test split.
The Thingi10K contains man-made objects that often possess highly complex topology. Vast majority of the Thingi10K shapes do not exists in ShapeNet and do not even relate to any of its categories. Thingi10K also provides topological complexity information (genus) for a large subset of the dataset. We use it as a proxy of topological complexity for additional evaluation.

\paragraph{Quantitative results}
Table \ref{tab:reconstruction_shapenet_eval} shows quantitative evaluation on ShapeNet-v2. The odd rows report the measures averaged over the $10$ largest categories of ShapeNet, and the even rows report the measures averaged over all $55$ categories.  In terms of the surface-based metrics of CD and F1 scores, our model is consistently better than the baselines.  In terms of IoU, our model has comparable performance. The results reflect a more accurate surface reconstruction, without compromising volumetric (IoU) accuracy. The supplement reports the performance for each of the largest $10$ categories -- again, we outperform prior methods, especially for topologically challenging categories, such as lamps and watercrafts. 
The out-of-distribution evaluation on Thingi10K is shown in Table \ref{tab:thingi_quant_eval}. Our model clearly outperforms the baselines -- the performance gap increases for topologically challenging shapes with higher genus. We note that this evaluation does not include IoU since some Thingi10K ground-truth shapes are not watertight, thus, volume-based metrics cannot be accurately assessed.

\begin{figure}[!b]

\vspace{-10pt}
\setlength\tabcolsep{-3pt}
\begin{tabular}{ccccccccc}
 Skeleton & Sample 1 & Sample 2 & Sample 3 & Sample 4 \\
\addlinespace[5pt]
\includegraphics[width=.21\linewidth,valign=m]{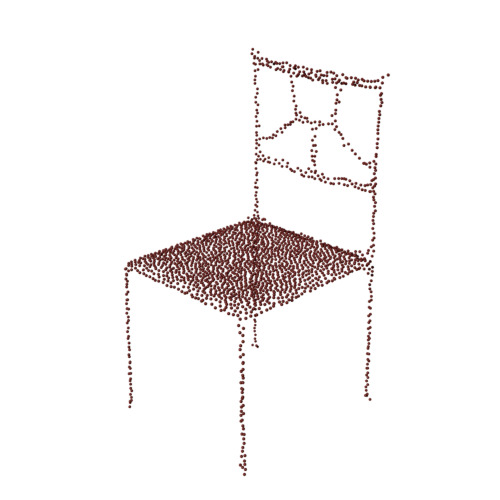} & 
\includegraphics[width=.21\linewidth,valign=m]{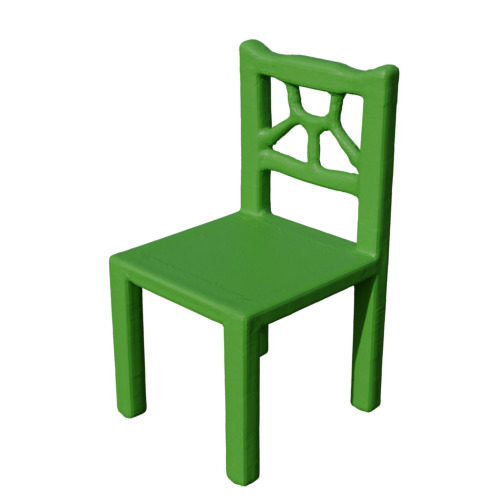} &
\includegraphics[width=.21\linewidth,valign=m]{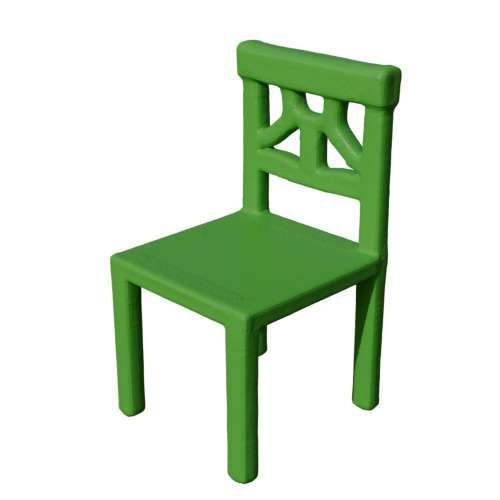} & 
\includegraphics[width=.21\linewidth,valign=m]{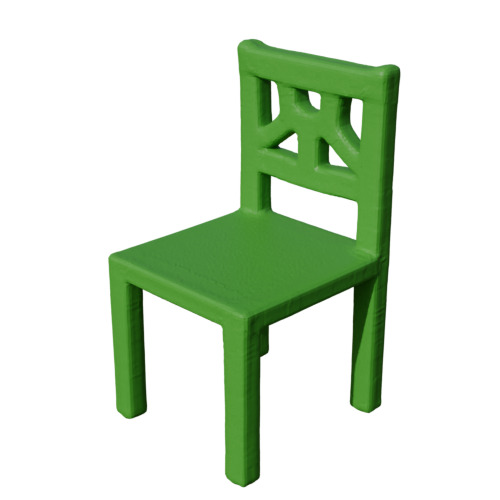} & 
\includegraphics[width=.21\linewidth,valign=m]{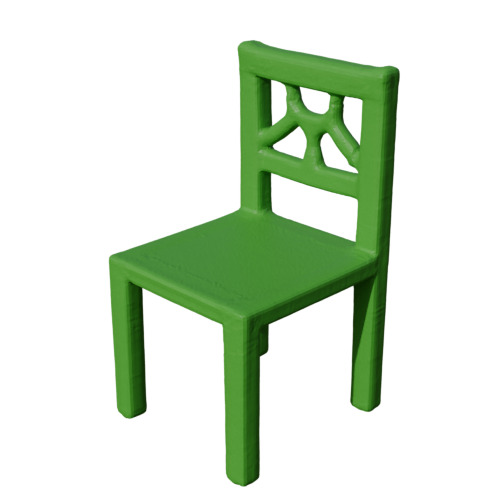} & 
\\
\includegraphics[width=.21\linewidth,valign=m]{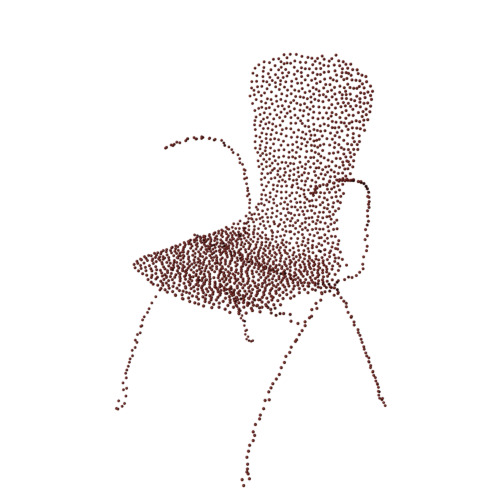} & 
\includegraphics[width=.21\linewidth,valign=m]{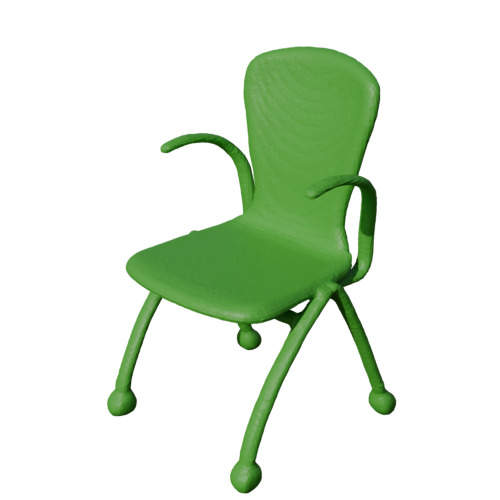} &
\includegraphics[width=.21\linewidth,valign=m]{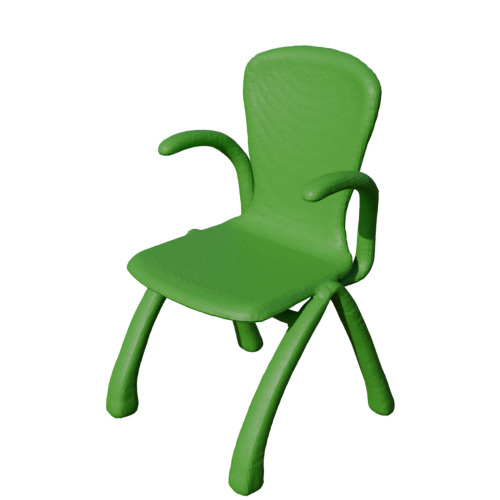} & 
\includegraphics[width=.21\linewidth,valign=m]{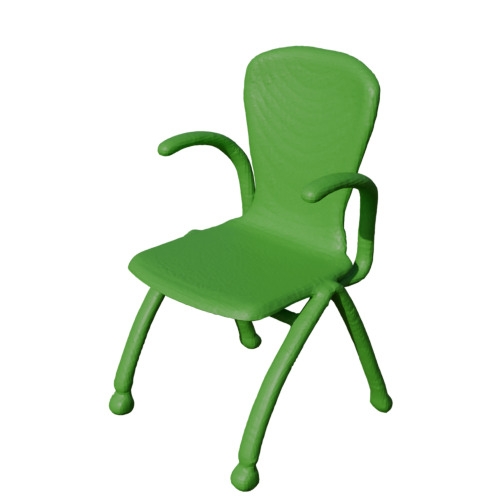} & 
\includegraphics[width=.21\linewidth,valign=m]{fig/img/surf_sampling_variance/03001627/skel_seed_48/03001627-48-19_render0001.jpg} & 
\\
\includegraphics[width=.21\linewidth,valign=m]{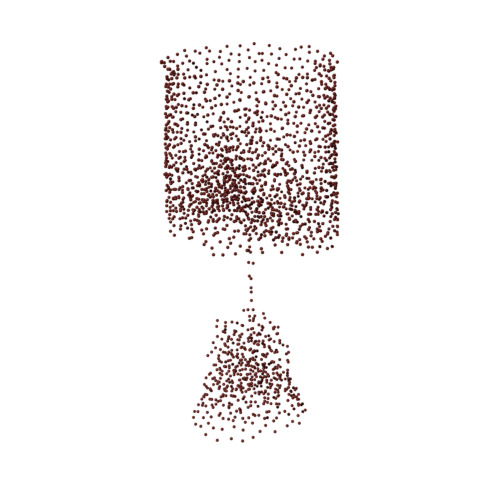} & 
\includegraphics[width=.21\linewidth,valign=m]{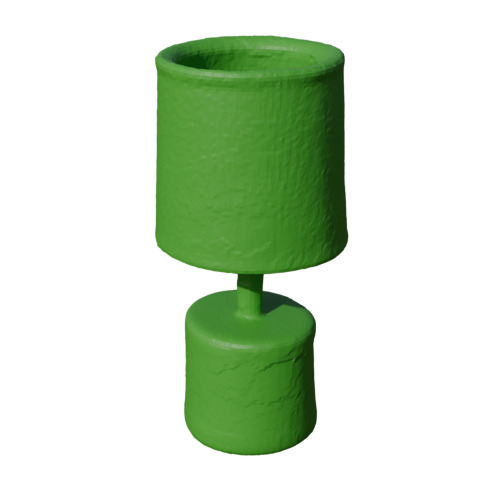} &
\includegraphics[width=.21\linewidth,valign=m]{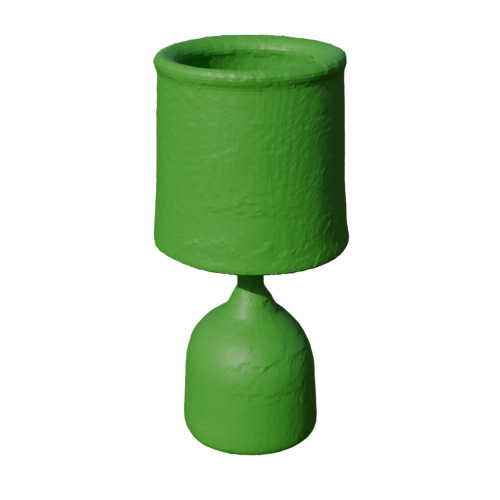} & 
\includegraphics[width=.21\linewidth,valign=m]{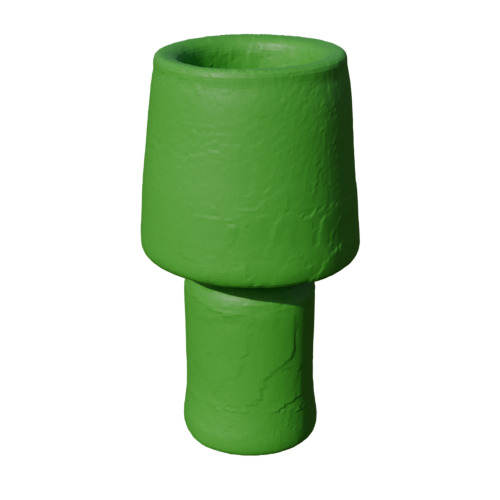} & 
\includegraphics[width=.21\linewidth,valign=m]{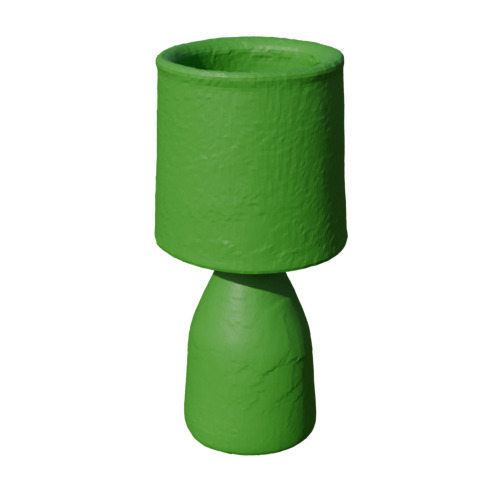} & 
\\
\includegraphics[width=.21\linewidth,valign=m]{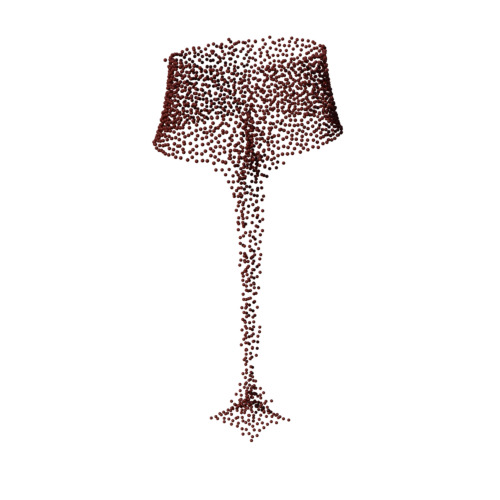} & 
\includegraphics[width=.21\linewidth,valign=m]{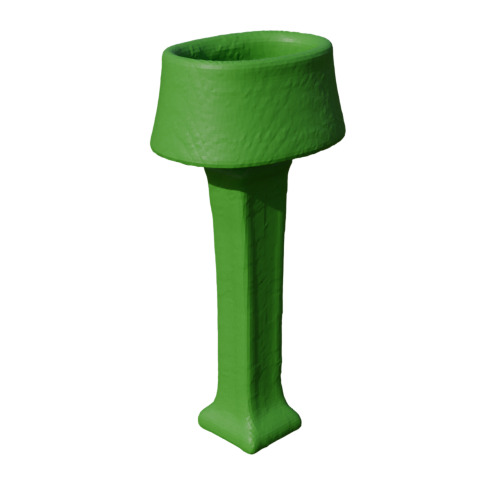} &
\includegraphics[width=.21\linewidth,valign=m]{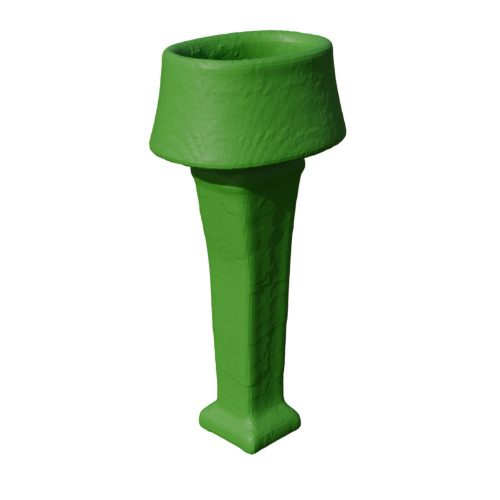} & 
\includegraphics[width=.21\linewidth,valign=m]{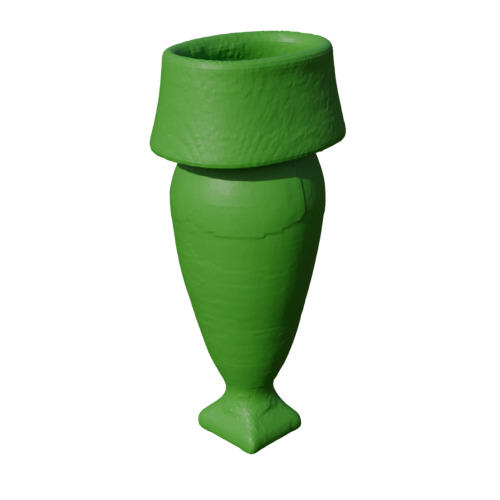} & 
\includegraphics[width=.21\linewidth,valign=m]{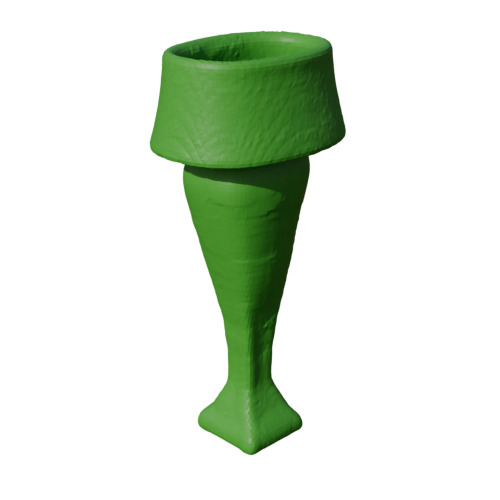} & 

\\
\addlinespace[-5pt]
\includegraphics[width=.21\linewidth,valign=m]{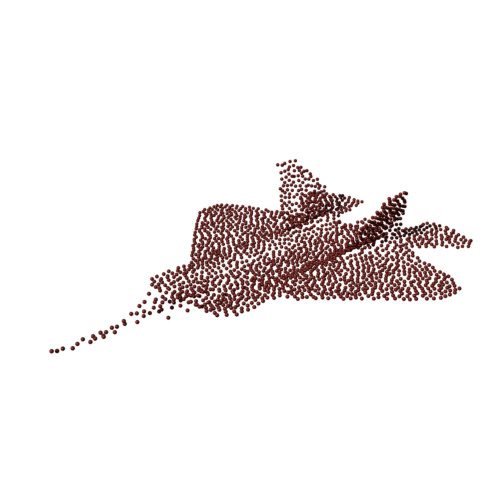} & 
\includegraphics[width=.21\linewidth,valign=m]{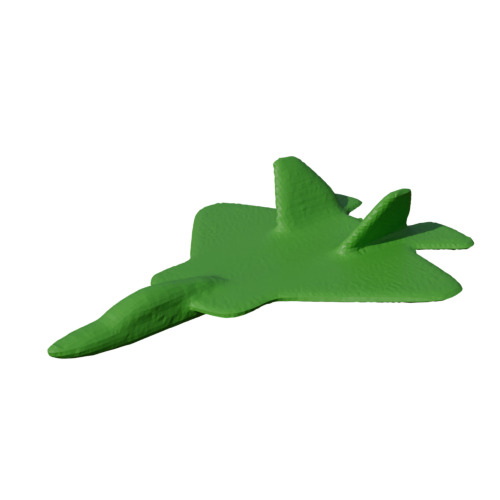} &
\includegraphics[width=.21\linewidth,valign=m]{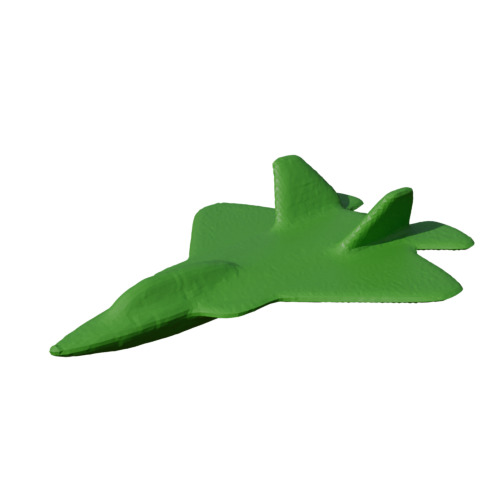} & 
\includegraphics[width=.21\linewidth,valign=m]{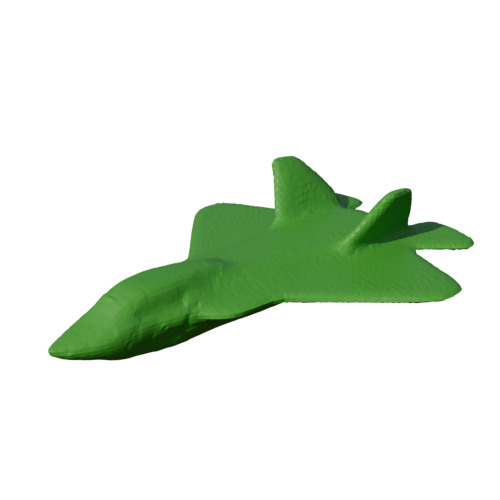} & 
\includegraphics[width=.21\linewidth,valign=m]{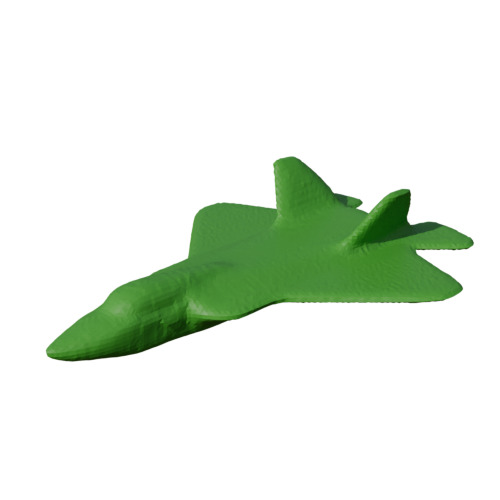} & 
\end{tabular}

\vspace{-10pt}
\caption{\textbf{Surface sampling conditioned on skeleton.} Our model can sample diverse and structurally consistent surfaces from generated skeletons. Surfaces were sampled using following category tokens (from top to bottom): chair, chair, lamp, lamp, airplane.}
\label{fig:surface_sampling_gen}
\vspace{-2mm}
\end{figure}

\begin{figure}[!b]
\begin{center}
\vspace{-5pt}
\setlength\tabcolsep{-2pt}
\begin{tabular}{ccccccc}
 Skeleton & Sample 1 & Sample 2 & Sample 3  \\

\addlinespace[5pt]
\includegraphics[width=.23\linewidth,valign=m]{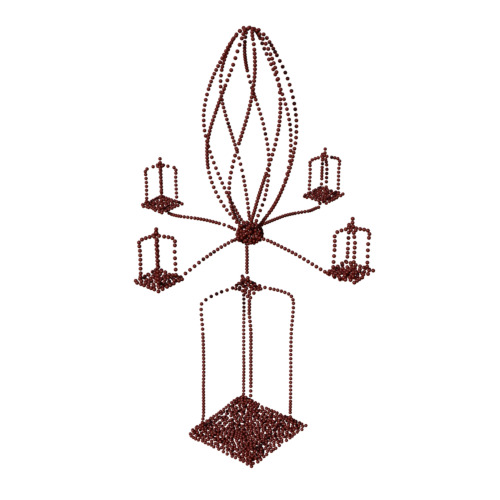} & 
\includegraphics[width=.23\linewidth,valign=m]{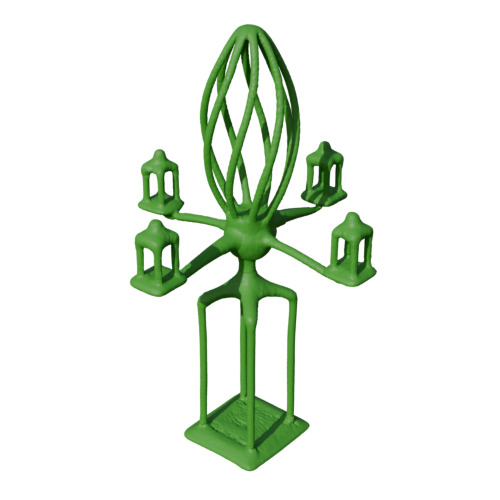} &
\includegraphics[width=.23\linewidth,valign=m]{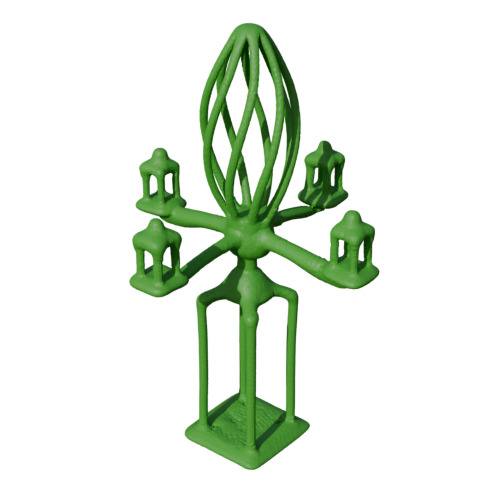} & 
\includegraphics[width=.23\linewidth,valign=m]{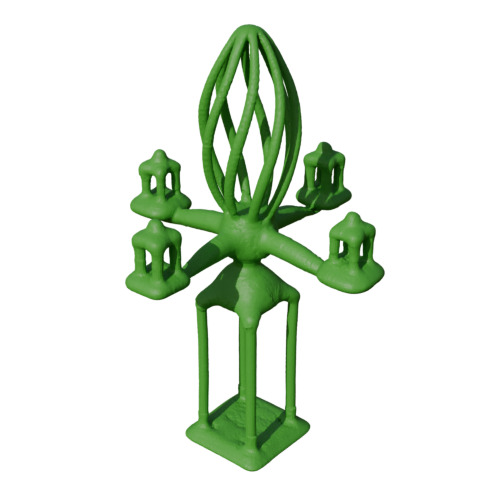} & 

\\
\addlinespace[-2pt]
\includegraphics[width=.23\linewidth,valign=m]{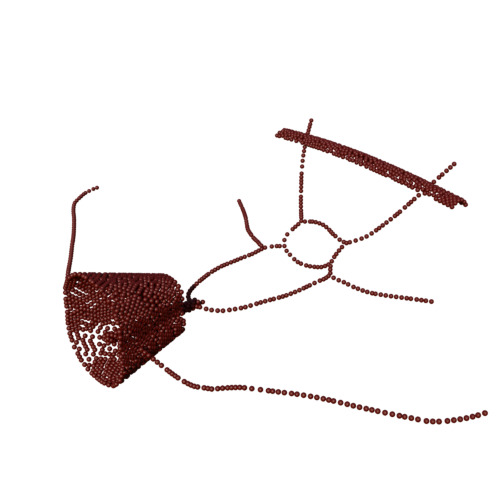} & 
\includegraphics[width=.23\linewidth,valign=m]{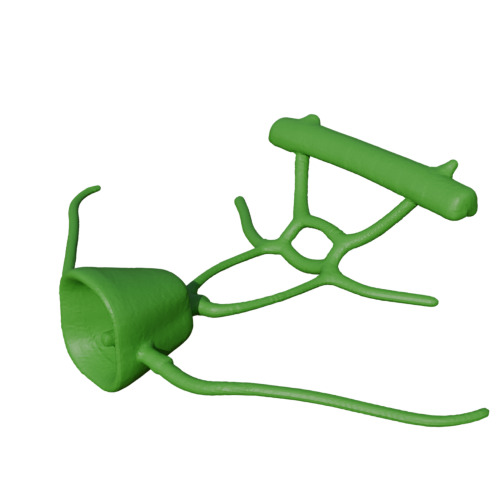} & 
\includegraphics[width=.23\linewidth,valign=m]{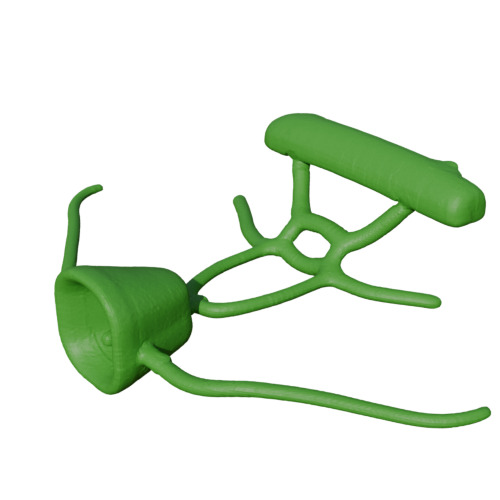} & 
\includegraphics[width=.23\linewidth,valign=m]{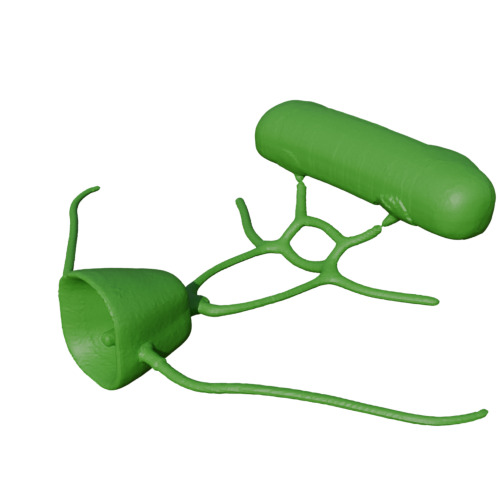} &

\\
\addlinespace[-5pt]
\includegraphics[width=.23\linewidth,valign=m]{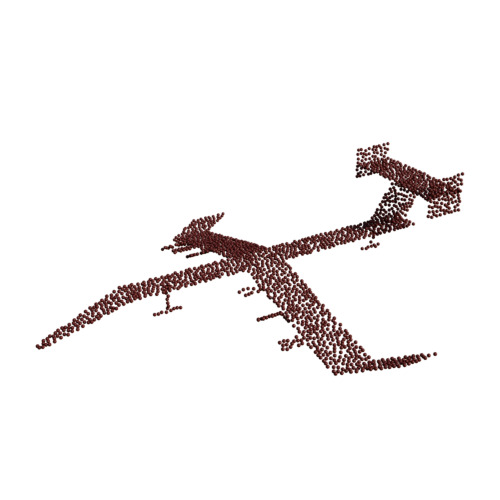} & 
\includegraphics[width=.23\linewidth,valign=m]{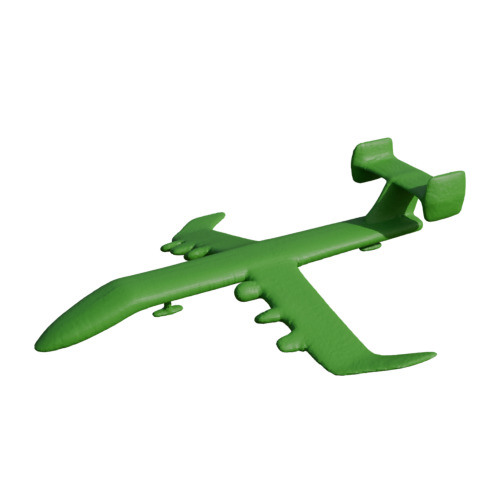} & 
\includegraphics[width=.23\linewidth,valign=m]{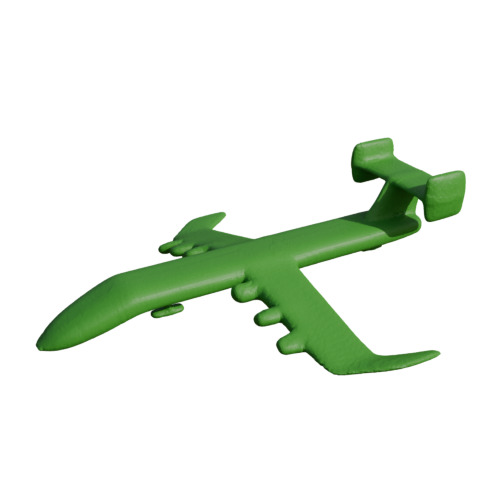} & 
\includegraphics[width=.23\linewidth,valign=m]{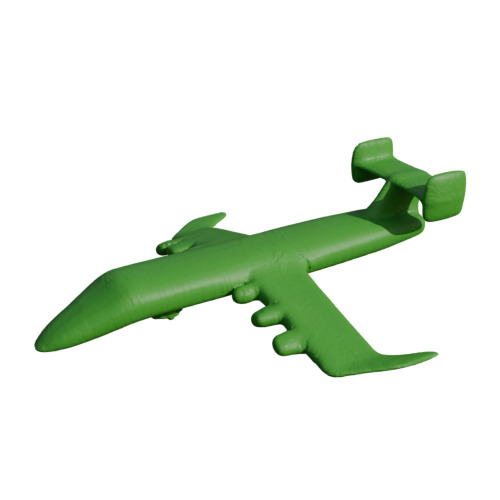} &

\\
\addlinespace[-5pt]
\includegraphics[width=.23\linewidth,valign=m]{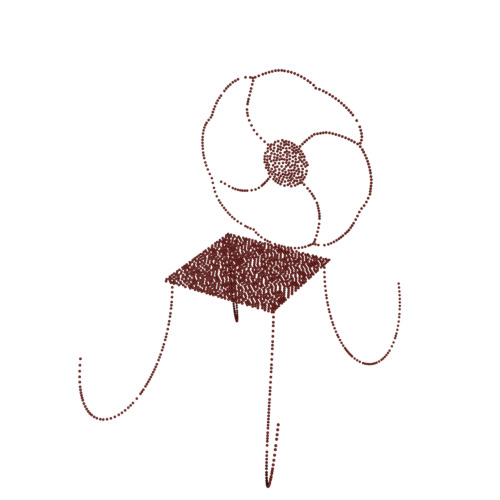} & 
\includegraphics[width=.23\linewidth,valign=m]{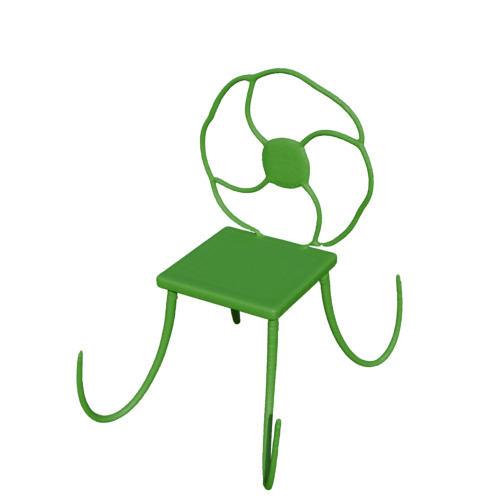} & 
\includegraphics[width=.23\linewidth,valign=m]{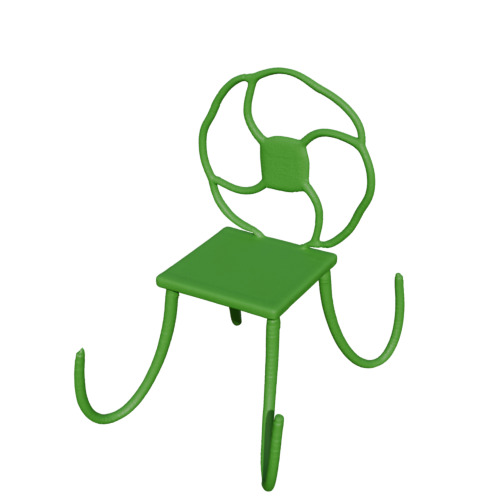} & 
\includegraphics[width=.23\linewidth,valign=m]{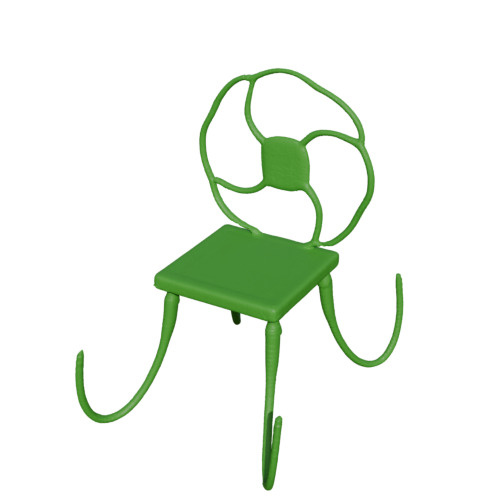} &

\end{tabular}

\end{center}
\vspace{-10pt}
\caption{\textbf{Surface generation based on user-modeled skeletons.} Our model is able to generalize to unseen structures encoded in user-provided skeletons, and produces plausible surface samples from them. Surfaces were sampled using following category tokens (from top to bottom): lamp, lamp, airplane, chair.}
\label{fig:zero_shot_gen}
\vspace{-10pt}
\end{figure}

\paragraph{Qualitative results}
Figures \ref{fig:comparison_pc_ae_shapenet} and \ref{fig:comparison_pc_thingi}
shows qualitative evaluation on ShapeNet-v2 and Thingi10K respectively. In particular, the Thingi10K results show that our method can better reconstruct surface topology e.g., the connectivity of different shape parts and their surface holes. We suspect that this is due to the ability of our method to capture structure and topology information through the neural medial representation,
which in turn guides the reconstruction. In contrast, 3DILG and 3D2VS tend to oversmooth surface details, miss connections, and close holes in the shapes. 

\paragraph{Ablation studies} In our supplementary material, we also show the impact of various choices in our method, including testing with different number of medial points and using closest medial points versus closest envelopes for reconstruction.

\subsection{Skeleton-driven shape synthesis}
An alternative application of our method is to generate surfaces driven by an input skeleton in the form of a point-based MAT. From a practical point of view, one possibility for users of our method is to execute our first diffusion stage, obtain various shape structures represented in their MATs, then select the one matching their target structure or topology goals more. Then the user can execute the second stage to obtain various shape alternatives conforming to the given structure.  Figure \ref{fig:surface_sampling_gen} highlights several examples of this application scenario. We show a generated MAT, then diverse surfaces conforming to it. For example, our method can generate diverse lamp stands from a table lamp structure or chairs of diverse part thicknesses conforming to a particular chair structure. 

Moreover, instead of obtaining a MAT from our method, an alternative scenario is that the user provides such as input. We asked an artist to draw the skeletons of completely fictional objects using 3D B-spline curves and patches. Our method was still able to generate diverse surfaces conforming to these out-of-distribution skeletons, as shown in Figure \ref{fig:zero_shot_gen}.

\section{Discussion}
\label{sec:conclusion}

We presented a skeleton-centered generative model of 3D shapes. Our method first captures structure and topology in the form of a generated skeleton, then synthesizes the surface guided from it. Users can also provide their own modeled skeletons in the process to control shape synthesis. 

\begin{wrapfigure}{R}{0.4\columnwidth}
\centering
 \vspace{-5mm} 
\includegraphics[width=.99\linewidth,valign=m]{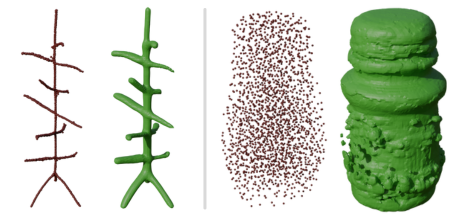} 



\vspace{-4mm}
  \caption{\textbf{Failure cases.} (Left) Generated MAT is unrealistic for lamp category. (Right) Generated surface is non-smooth.
  \label{fig:failure_cases} }
\vspace{-4mm}       
\end{wrapfigure}

\vspace{-2mm}
\paragraph{Limitations}
Failure cases of our method include unrealistic generated skeletons and noisy generated surfaces due to noise in MATs (Figure \ref{fig:failure_cases}). We hypothesize this is due to either noisy training MATs or the training data being too limited for the model to learn comprehensive structural priors for challenging categories.
Our method also has a number of other limitations. First, it relies on a simplified form of point-based skeletons. 
Second, the diversity of surfaces sampled from skeletons is sometimes limited. Investigation of less expressive skeletal representations or additional types of conditioning (e.g. images or text) can improve this aspect of our model. Other future directions would be to enable interactive skeleton editing, shape synthesis from edits, combinations of our model with sketch-based editing models \cite{luo20233d} and physics-based shape deformations \cite{lan2020medial}. 

\paragraph{Acknowledgements} We thank Joseph Black, Rik Sengupta and Alexey Goldis for providing useful feedback and input to our project. We acknowledge funding from (i) Adobe Research, (ii) the EU H2020 Research Innovation Programme, and (iii) the Republic of Cyprus through the Deputy Ministry of Research, Innovation and Digital Policy (GA 739578).



\bibliographystyle{ACM-Reference-Format}
\bibliography{main}

\clearpage

%



\clearpage
\appendix
\section*{Supplementary Material}


\section{Per-category evaluation}
In this appendix, we discuss  more detailed evaluation, including 
per-category evaluation for generation and reconstruction. 

\paragraph{Shape generation}  In terms of generative measures, we report precision and recall per category in Table \ref{tab:gen_quant_prd}, and PointBert-FID and PointBert-KID per category in Table \ref{tab:gen_quant_fid_kid}. We also report the per-category MMD-CD, MMD-COV, COV-CD, COV-MD in Table \ref{tab:gen_quant_mmd_cd_3x}. Overall, we observe that for the majority of categories, GEM3D outperforms the competing methods in terms of all metrics, especially in terms of generative recall. We hypothesize that is due to fact that our two step generative procedure allows model to learn better structural priors leading to generation of more structurally diverse shapes.

\begin{table}[b!]
\begin{adjustbox}{width=\columnwidth,center}
    \begin{tabular}{lccc|ccc}
\toprule
\midrule
 & \multicolumn{3}{c}{Precision ($\uparrow$)} & \multicolumn{3}{c}{Recall ($\uparrow$)} \\
 & 3DILG & 3DS2VS & Ours & 3DILG & 3DS2VS & Ours \\ 
\midrule
table & 76.0 & 74.0 & \textbf{91.3} & 74.0 & 77.4 & \textbf{95.6}  \\
car & 48.5 & \textbf{74.8} & 73.5 & 61.6 & 73.5 & \textbf{89.2} \\
chair &  \textbf{91.7} & 82.4 & 88.9 & 90.9 & 84.7 & \textbf{93.2} \\
plane & 82.2 & 74.4 & \textbf{88.8} & 78.8 & 86.4 & \textbf{95.3} \\
sofa & 64.8 &  \textbf{79.4} & 78.6 & 82.5 & 88.9 & \textbf{93.9} \\
rifle & 69.3 & 69.6 & \textbf{75.2} & 71.0 & 74.9 & \textbf{85.8} \\
lamp & \textbf{82.8} & 50.7 & 69.1 & \textbf{86.1} & 72.2 & 74.4 \\
w-craft & 72.3 & 77.6 & \textbf{84.4} & 86.1 & 83.8 & \textbf{87.1}
\\
bench & 71.6 & 70.4 & \textbf{81.3} & 84.5 & 79.2 & \textbf{88.4} \\
speaker & 71.7 & 71.0 & \textbf{74.4} & 78.9 & 81.5 & \textbf{89.5} \\
\bottomrule
\end{tabular}
\end{adjustbox}
\caption{Per-category evaluation of generated samples based on precision and recall using PointBert as a backbone to assess shape similarity. We note that for each method we generate a number of samples equal to $3$x the size of the ground truth test split per category.  \label{tab:gen_quant_prd}}
\end{table}

\begin{table}[b!]

\begin{adjustbox}{width=\columnwidth,center}
    \begin{tabular}{lccc|ccc}
\toprule
\midrule
 & \multicolumn{3}{c}{PointBert-FID ($\times 10^{-2}$) ($\downarrow$)} & \multicolumn{3}{c}{PointBert-KID ($\downarrow$)} \\
 & 3DILG & 3DS2VS & Ours & 3DILG & 3DS2VS & Ours \\ 
\midrule
table & 1.57 & 1.53 & \textbf{0.71} & 6.87 & 7.49 & \textbf{1.31 } \\
car & 1.39 & 0.98 & \textbf{0.82} & 7.68 & 3.56 & \textbf{3.13} \\
chair &  0.84 & 1.24 & \textbf{0.79} & 1.76 & 3.74 & \textbf{1.20 }\\
plane & 0.90 & 1.06 & \textbf{0.79} & 2.10 & 4.08 & \textbf{1.46} \\
sofa & 1.29 &  1.07 & \textbf{1.05} & 3.37 & \textbf{1.46} & 1.96 \\
rifle & 1.52 & 1.51 & \textbf{1.07} & 6.53 & 5.87 & \textbf{2.51} \\
lamp & \textbf{1.84} & 3.81 & 2.88 & \textbf{2.28} & 15.2 & 9.32 \\
w-craft & 1.70 & 1.49 & \textbf{1.34} & 4.18 & 1.70 & \textbf{0.86}
\\
bench & 1.70 & 1.75 & \textbf{1.47} & 2.84 & 2.71 & \textbf{1.16} \\
speaker & 1.85 & 1.79 & \textbf{1.62} & 2.95 & \textbf{1.85} & 2.50 \\
\midrule
\end{tabular}
\end{adjustbox}
\caption{Per-category evaluation of generated samples based on FID and KID scores using PointBert as a backbone for feature extraction. We note that for each method we generate a number of samples equal to $3$x the size of the ground truth test split per category. \label{tab:gen_quant_fid_kid}}

\end{table}

\paragraph{Surface reconstruction from point clouds} For reconstruction, we report CD, IoU and F1 scores for each of the largest $10$ categories of ShapeNet in Table \ref{tab:ShapeNet_detailed_ae}. We also report performance using $512$ latent codes vs $2048$ latent codes for all methods. 
For most categories, GEM3D outperforms the competing methods especially in terms of the surface-based metrics (CD and F1 scores), while it offers comparable performance in terms of IoU. As expected, the performance is improved with more latent codes for all methods.  As explained in our main text, this evaluation is done on a higher-resolution grid and with more uniform point-based surface sampling (furthest point sampling). In contrast, 3DShape2VecSet follows a different evaluation protocol based on a lower-resolution grid and random point sampling. We report the evaluation measures for reconstruction based on the original 3DShape2VecSet protocol on the categories reported in their paper in Table \ref{tab:ShapeNet_detailed_ae_orig}. We observe the same trends, yet, our gap with other methods slightly decreases given that this evaluation protocol is less sensitive to topological and surface details.

\begin{table*}[t!]
    \begin{tabular}{llll|lll|lll|lll}
\toprule
\midrule
 & \multicolumn{3}{c}{MMD-CD ($\times 10^{2}$) ($\downarrow$)} & \multicolumn{3}{c}{MMD-EMD ($\times 10^{2}$) ($\downarrow$)} & \multicolumn{3}{c}{COV-CD ($\times 10^{2}$) ($\uparrow$)} & \multicolumn{3}{c}{COV-EMD ($\times 10^{2}$) ($\uparrow$)} \\
 & 3DILG & 3DS2VS & Ours & 3DILG & 3DS2VS & Ours & 3DILG & 3DS2VS & Ours & 3DILG & 3DS2VS & Ours \\
\midrule
table & 11.00 & 9.87 & \textbf{9.01} & 12.63 & 11.74 & \textbf{10.67} & 44.6 & 59.4 & \textbf{65.0} & 46.7 & 53.0 & \textbf{62.9} \\
car & 5.42 & 5.38 & \textbf{5.07} & 6.37 & 6.16 & \textbf{5.92} & 41.9 & \textbf{44.0} & 42.8 & 34.2 & \textbf{46.6} & 46.4 \\
chair & 11.89 & 12.44 & \textbf{10.99} & 12.86 & 13.66 & \textbf{12.37} & \textbf{61.4} & 51.9 & 61.3 & \textbf{61.1} & 48.8 & 56.0 \\
airplane & 5.13 & 5.60 & \textbf{4.70 }& 6.67 & 7.23 & \textbf{6.21} & 56.2 & 46.5 & \textbf{64.3} & 61.3 & 47.0 & \textbf{63.3} \\
sofa & 12.2 & 9.7 & \textbf{9.4} & 12.51 & 10.63 & \textbf{9.82} & 43.2 & \textbf{54.8} & 53.1 & 40.3 & \textbf{54.0} & 53.8 \\
rifle & 5.27 & 5.35 & \textbf{5.05} & 6.83 & 6.86 & \textbf{6.79} & \textbf{59.3} & 40.4 & 55.9 & \textbf{60.4} & 44.3 & 53.9 \\
lamp & \textbf{15.06} & 15.93 & 15.32 & \textbf{19.24} & 20.88 & 19.69 & \textbf{64.6} & 41.7 & 51.0 & \textbf{61.7} & 43.4 & 49.2 \\
w-craft & 8.53 & 7.70 & \textbf{7.16} & 9.55 & 8.93 & \textbf{8.33} & 65.2 & 60.0 & \textbf{71.8} & 59.3 & 56.6 & \textbf{65.9} \\
bench & 9.68 & 9.31 & \textbf{9.05} & 10.72 & 10.39 & \textbf{10.03} & \textbf{54.8} & 51.8 & 54.0 & 53.3 & 52.5 & \textbf{55.1} \\
speaker & 12.27 & 12.05 & \textbf{10.56} & 12.57 & 12.66 & \textbf{11.45} & 57.3 & 55.7 & \textbf{65.4} & 52.3 & 53.1 & \textbf{64.9} \\
\bottomrule
\end{tabular}
\caption{Per-category evaluation of generated samples based on MMD-CD, MMD-COV, COV-CD, and COV-MD. We note that for each method we generate a number of samples equal to $3$x the size of the ground truth test split per category.
\label{tab:gen_quant_mmd_cd_3x}}
\end{table*}

\begin{table}[t!]
  \centering
\begin{tabular}{@{}ccccccccc@{}}
\toprule
\midrule
 &   &  \multicolumn{3}{c}{3DILG 3D2VS Ours}        &   \multicolumn{3}{c}{3DILG   3D2VS Ours}           \\ 
& \# latents  &   512          &  512      &     512     &     2048   & 2048 & 2048\\  \hline

\multirow{12}{*}{ CD $\downarrow$} 
&table       & 1.38 & 1.05 &\textbf{1.01} & 1.36 & 1.03 & \textbf{0.99} \\
&car         & 2.29 & 1.86 &\textbf{1.24} & 2.27 & 1.76 & \textbf{1.18} \\
&chair       & 1.42 & 1.09 &\textbf{1.05} & 1.39 & 1.06 & \textbf{1.03} \\
&airplane    & 0.98 & \textbf{0.59} &\textbf{0.59} & 0.97 & \textbf{0.58} & \textbf{0.58} \\
&sofa        & 1.37 & 1.06 &\textbf{1.02} & 1.35 & 1.04 & \textbf{1.00} \\
&rifle       & 0.94 & 0.47 &\textbf{0.45} & 0.92 & 0.46 & \textbf{0.45} \\
&lamp        & 1.82 & 1.02 &\textbf{0.72} & 1.8  & 1.02 & \textbf{0.72} \\
&w-craft  & 1.25 & 0.84 &\textbf{0.71} & 1.23 & 0.79 & \textbf{0.69} \\
&bench       & 1.28 & 0.94 &\textbf{0.84} & 1.27 & 0.93 & \textbf{0.82} \\
&speaker & 1.8  & 1.46 &\textbf{1.28} & 1.77 & 1.42 & \textbf{1.23} \\

\hline
 \multirow{12}{*}{ IOU $\uparrow$}
&table       & 88.9 & \textbf{95.9} & 93.5 & 88.9 & \textbf{96.2} & 94.3 \\
&car         & 92.9 & \textbf{95.8} & 94.8 & 93.1 & \textbf{96.2} & 95.7 \\
&chair       & 90.5 & \textbf{95.9} & 94.8 & 90.6 & \textbf{96.4} & 95.3 \\
&airplane    & 89.3 & \textbf{96.5} & \textbf{96.5} & 89.6 & \textbf{96.9} & 96.7 \\
&sofa        & 95.1 & \textbf{98.1} & 97.5 & 95.2 & \textbf{98.3} & 97.8 \\
&rifle       & 87.0 & 96.0 & \textbf{96.5} & 87.2 & 96.2 & \textbf{96.6} \\
&lamp        & 85.6 & 94.0 & \textbf{94.5} & 86.5 & \textbf{94.6} & 93.9 \\
&w-craft  & 90.8 & 96.4 & \textbf{96.5} & 91.0 & 96.7 & \textbf{96.8} \\
&bench       & 85.7 & \textbf{94.4} & 94.0 & 86.0 & 94.7 & \textbf{94.8} \\
&speaker & 92.4 & \textbf{95.9} & 94.7 & 92.6 & 96.3 & \textbf{97.7} \\
\hline
 \multirow{12}{*}{ F1 $\uparrow$}
&table       & 99.0 & 99.4 & \textbf{99.6} & 99.1 & 99.5 & \textbf{99.7} \\
&car         & 91.6 & 93.2 & \textbf{96.5} & 91.9 & 93.9 & \textbf{97.4} \\
&chair       & 98.5 & 99.2 & \textbf{99.4} & 98.7 & 99.3 & \textbf{99.4} \\
&airplane    & 99.6 & \textbf{99.9} & \textbf{99.9} & 99.6 & 99.9 & \textbf{99.9} \\
&sofa        & 98.7 & 99.2 & \textbf{99.5} & 98.9 & 99.4 & \textbf{99.6} \\
&rifle       & 99.7 & 99.9 & \textbf{100.0} & 99.7 & \textbf{100.0} & \textbf{100.0} \\
&lamp        & 97.4 & 98.6 & \textbf{99.5} & 97.4 & 98.7 & \textbf{99.4} \\
&w-craft  & 98.1 & 98.6 & \textbf{99.7} & 98.2 & 98.9 & \textbf{99.8} \\
&bench       & 98.7 & 99.3 & \textbf{99.7} & 98.9 & 99.3 & \textbf{99.7} \\
&speaker & 95.9 & 96.9 & \textbf{98.3} & 96.2 & 97.4 & \textbf{98.7} \\
\bottomrule
\end{tabular}
  \caption{ Evaluation of reconstructed surfaces on ShapeNet based on Chamfer Distance (CD), Intersection over Union (IoU), and F1 scores for the auto-encoding task. All numbers are scaled by 100. 
    \label{tab:ShapeNet_detailed_ae}
}
\vspace{-16pt}
\end{table}

\begin{table*}
  \centering
\begin{tabular}{ccccccccc}
\toprule
 &          & OccNet  & ConvOccNet & IF-Net          & 3DILG            & 3D2VS-LQ & 3D2VS-PQ & GEM3D \\
\midrule
 \multirow{9}{*}{CD $\downarrow$}
& table    & 0.041 & 0.036 & 0.029 & 0.026 & 0.026 & 0.026  & \textbf{0.014}\\
 & car      & 0.082 & 0.083 & 0.067 & 0.066          & 0.062 & 0.062   & \textbf{0.016}\\
  & chair    & 0.058 & 0.044 & 0.031 & 0.029          & 0.028          & 0.027 &  \textbf{0.015}\\
 & airplane & 0.037 & 0.028 & 0.020 & 0.019          & 0.018          & 0.017 &  \textbf{0.008}\\
 & sofa     & 0.051 & 0.042 & 0.032 & 0.030          & 0.030          & 0.029 & \textbf{0.015}\\
 & rifle    & 0.046 & 0.025 & 0.018 & 0.017          & 0.016          & 0.014 & \textbf{0.006}\\
  & lamp     & 0.090 & 0.050 & 0.038 & 0.036          & 0.035          & 0.032 & \textbf{0.010}\\ 
 \cline{2-9}
 & mean (selected)  & 0.058 & 0.040 & 0.034 & 0.032          & 0.031          & 0.030 & \textbf{0.012}\\
& mean (all)   & 0.072 & 0.052 & 0.041 & 0.040          & 0.039       &0.038 &  \textbf{0.015} \\
\hline
\multirow{9}{*}{ IoU $\uparrow$}  
 & table    & 0.823 & 0.847 & 0.901 & 0.963          & 0.965          & \textbf{0.971} & 0.960\\ 
 & car      & 0.911 & 0.921 & 0.952 & 0.961          & 0.966          & \textbf{0.969} & 0.936\\
 & chair    & 0.803 & 0.856 & 0.927 & 0.950          & 0.957          & \textbf{0.964} & 0.958 \\
 & airplane & 0.835 & 0.881 & 0.937 & 0.952          & 0.962          & \textbf{0.969} & 0.961\\
 & sofa     & 0.894 & 0.930 & 0.960 & 0.975          & 0.975          & \textbf{0.982} & 0.974\\
 & rifle    & 0.755 & 0.871 & 0.914 & 0.938          & 0.947          & \textbf{0.960} & 0.956 \\
 & lamp     & 0.735 & 0.859 & 0.914 & 0.926          & 0.931          & \textbf{0.956} & 0.934\\ \cmidrule(lr){2-9}
 &  mean (selected) & 0.822 & 0.881 & 0.929 & 0.952          & 0.957          & \textbf{0.967} & 0.954 \\
  & mean (all) & 0.825 & 0.888 & 0.934 & 0.953          & 0.955          & \textbf{0.965} & 0.944\\ \hline
 & table    & 0.961 & 0.982 & 0.998 & \textbf{0.999} & \textbf{0.999} & \textbf{0.999} & 0.991\\
 & car      & 0.830 & 0.852 & 0.888 & 0.892          & 0.898          & 0.899 & \textbf{0.965}\\
 & chair    & 0.890 & 0.943 & 0.990 & 0.992          & 0.994          & \textbf{0.997} & 0.984\\
 & airplane & 0.948 & 0.982 & 0.994 & 0.993          & 0.994          & 0.995 & \textbf{0.998}\\
 & sofa     & 0.918 & 0.967 & 0.988 & 0.986          & 0.986          & \textbf{0.990} & 0.987\\
 & rifle    & 0.922 & 0.987 & 0.998 & 0.997          & 0.998          & \textbf{0.999} & \textbf{0.999} \\
 & lamp     & 0.820 & 0.945 & 0.970 & 0.971          & 0.970          & 0.975 & \textbf{0.990} \\ 
 \cline{2-9}
 &  mean (selected)  & 0.898 & 0.951 & 0.975 & 0.976          & 0.977         & 0.979 &  \textbf{0.988}\\
\multirow{-9}{*}{ F1 $\uparrow$} &  mean (all)  & 0.858 & 0.933 & 0.967 & 0.966          & 0.966          & \textbf{0.970} & 0.967\\ 
\bottomrule
\end{tabular}
 
  \caption{Evaluation of reconstructed surfaces on ShapeNet based on Chamfer Distance (CD), Intersection over Union (IoU), and F1 scores for the auto-encoding task. Here we use the original 3DShape2VecSet protocol. We report the same seven categories and averages (numbers are not scaled as in their paper).}
  \label{tab:ShapeNet_detailed_ae_orig}
\end{table*}


\begin{table}
\centering
\begin{tabular}{@{}ccccccc@{}}
\toprule
Choice in the decoder & CD ($\downarrow$) & IOU ($\uparrow$) & F1($\uparrow$) \\
\midrule
Closest medial point  & 1.41 & 89.9 & 98.6 \\
\midrule
Closest medial envelope   &\textbf{ 1.38} & \textbf{95.5} & \textbf{98.7} \\
\bottomrule
\end{tabular}
\caption{Ablation study based on the ShapeNet point cloud reconstruction task (we report averages over all $55$ categories). \label{tab:ablation}}
\vspace{-8mm}
\end{table}

\section{Ablation}
We perform the following ablation studies:\\
(a) We tested using $2048$ vs $512$ different number of medial points and correspondingly, different number of latent codes. Results are evaluated in the task of surface reconstruction and are shown in Table \ref{tab:ShapeNet_detailed_ae}. With more latents, our performance is improved, as expected. \\

(b) We tested using closest medial points vs closest medial envolope in our implicit function. Using closest medial envelopes yielded the best performance, as shown in Table \ref{tab:ablation}.


\SetKwComment{Comment}{/* }{ */}
\SetKwInput{KwInput}{Input}
\SetKwInput{KwInit}{Initialize}
\SetKwInput{KwParams}{Parameters}

\begin{algorithm}
\KwInput{Surface point samples $P = \{\bp_i\}_{i=1}^N$ }
\KwParams{Learning rate $\lambda=0.1$, local shape diameter threshold $\tau_{\text{max}}=0.6$,  number of  neighbors $K=20$ for kernel SDF estimation, kernel bandwidth $\sigma^2=0.002$, number of iterations $M=50$}
\KwResult{Medial points $S=\{\bs_i\}_{i=1}^N$}

\KwInit{For each surface point $\bp_i \in P$ estimate local shape diameter function $\beta_i$ through ray casting and its normal $\bn_i$ from the original mesh; initialize medial points $\bq_i = \bp_i - \frac{1}{2}\beta_i \bn_i$}
\For{iterations $< M$}{
  For each $\bq_i$ find $K$ nearest neighbors $\bp_{ij}$;\\
  Estimate local SDF $f(q_i) = \frac{\sum_j \alpha_{ij}f(\bp_{ij})}{\sum_j \alpha_j}$, where:\\\
  $\alpha_{ij} = \exp-\frac{||\bq_{i} - \bp_{ij}||^2}{\sigma^2}$ and \\
  $f(\bp_{ij})$ is the signed distance of $\bq_{i}$ to $\bp_{ij}$'s tangent plane;
 \\
  Compute updated skeleton points $\hat{\bq}_{i} = \bq_{i} - \lambda \nabla_{\bq_{i}} f(\bq_{i}) $; \\
  \If{$||\hat{\bq}_{i} - \bp_i|| \le \tau_{\text{max}} \beta_i$}{$
    \bs_{i} \leftarrow \hat{\bq}_{i}$
  }
}
\caption{Medial extraction algorithm}
\label{alg:medial_extraction}
\end{algorithm}

\section{Skeletonization}

Skeletonization is a well-studied topic in the geometry processing literature \cite{TagliasacchiSOTAReport}. All skeletonization methods rely on different assumptions, approximation heuristics, and convergence criteria that might lead to  different MAT approximations. We tried  various skeletonizations methods, including mean curvature flow skeletons \cite{taglia_sgp12}, medial skeletal diagrams \cite{guo2023medial}, and the ``neural skeletons'' by Clemot et al. \shortcite{Clemot2023}. However, we found that all methods either were too slow to process large datasets or needed manual parameter runing.
For our purposes, we needed a skeletonization method that is scalable i.e., is able to handle large shape collections, such as ShapeNet (in other words, it is able to extract a skeleton from a mesh efficiently e.g.,  a few seconds for the largest mesh). The algorithm should also be robust to varying mesh tesselations, and most importantly should not not require manual parameter adjustment for different shape categories.

During our early experiments, we found out that none of existing methods satisfy all the above criteria. 
These issues led us to develop our own  skeletonization algorithm that balances computational efficiency with the need for accurate skeletons. 
A key component of this algorithm is a gradient descent procedure on the signed distance function (SDF) of the shape approximated through a local RBF kernel
to shrink the surface iteratively towards its interior. 
We note that our skeletonization is inspired by \cite{Clemot2023}, yet with several important differences, including in the surface sampling, initialization, gradient descent procedure, and stopping criteria.
The algorithm is summarized in Algorithm \ref{alg:medial_extraction}. 
It initiates medial points from surface sample points from a given mesh by shifting the surface points according to their negated surface normal and a multiplier of the local shape diameter estimated per point through ray casting. This initialization bootstraps the procedure, since the subsequent iterative phase of the method is slower.
In the iterative phase,  for each current position of 
 medial point, the algorithm computes a  Signed Distance Function approximation (SDF) by averaging its signed distances to the tangent planes of the nearest surface points. The averaging is performed with the help of RBF\ kernel. Following the gradient of the SDF gradually shrinks the shape. The update is constrained: it is accepted if the updated point remains within a distance less than a given threshold expressed as a multiplier (approximately half) of the local shape diameter. This ensures that the points will meet close to the middle of the shape and will not drift too further away. The method continues until the skeleton point positions converge up to a tolerance threshold. The parameters used in our method are listed in Algorithm \ref{alg:medial_extraction}. For all our shapes involved in our experiments, these parameters were fixed.

 \section{Model capacity}

Comparison between GEM3D and 3DS2VS in terms of number of parameters is shown in Table \ref{tab:param_comparison}. Despite having two diffusion stages, total number of parameters for GEM3D model is only 1.26x more to those of 3DS2VS. This is due to fact that our model has way simpler surface decoder. We didn't have to run hyperparameter optimization for our generative models but we hypothesize that it can be potentially substantially reduced, especially for surface decoder. 

\begin{table}[hb!]
\centering
\begin{tabular}{lcccc}
\hline
 & Decoder & Gen-MAT & Gen-Surf & Total \\
\hline
3DS2VS & 102M & - & 164.2M & 266.2M\\
GEM3D & 6.3M & 164.9M & 164.9M & 336.1M \\
\hline
\end{tabular}
\caption{Parameter comparison between 3D2SVS vs GEM3D. Despite having two diffusion stages, overall increase in parameters for our model is only 1.26x compared to 3DS2VS.}
\label{tab:param_comparison}
\vspace{-18pt}
\end{table}




\clearpage
\end{document}